\documentclass{article}
\usepackage[preprint]{neurips_2025}
\usepackage{graphicx} 
\usepackage{amsfonts}
\usepackage{amsmath}
\usepackage{amssymb}
\usepackage{amsthm}
\usepackage{multirow}
\usepackage{tabularx}
\usepackage{booktabs}
\usepackage{adjustbox}
\usepackage{caption}
\usepackage{array}
\usepackage{url}
\usepackage{natbib}
\usepackage[table,dvipsnames]{xcolor}
\usepackage{colortbl}  
\usepackage{tikz}
\usepackage[hidelinks]{hyperref}
\usepackage{parskip}
\usepackage{algorithm}
\usepackage{algorithmicx}
\usepackage{algpseudocode}
\usepackage{listings}
\usepackage{bm}
\usepackage{pifont}
\usepackage{enumitem}
\usepackage{verbatim}
\usepackage{longtable}
\floatname{algorithm}{Algorithm}

\usepackage{lineno}
\definecolor{warmbg}{RGB}{255, 249, 242}      
\definecolor{warmheader}{RGB}{240, 225, 205}  
\definecolor{warmrow}{RGB}{250, 242, 230}     
\newcommand{\cmark}{\textcolor{green!60!black}{\ding{51}}} 
\newcommand{\xmark}{\textcolor{red!70!black}{\ding{55}}}   

\usepackage{cleveref}
\usepackage[most]{tcolorbox}
\newtcbtheorem{Theorem}{Theorem}{
  enhanced,
  sharp corners,
  attach boxed title to top left={
    yshifttext=0mm
  },
  colback=white,
  colframe=teal,
  fonttitle=\bfseries,
  boxed title style={
    sharp corners,
    size=small,
    colback=teal,
    colframe=teal,
  } 
}{thm}
\newtcbtheorem[no counter]{Proof}{Proof}{
  enhanced,
  sharp corners,
  attach boxed title to top left={
    yshifttext=0mm
  },
  colback=white,
  colframe=teal!50,
  fonttitle=\bfseries,
  coltitle=black,
  boxed title style={
    sharp corners,
    size=small,
    colback=teal!50,
    colframe=teal!50,
  } 
}{prf}

\usepackage{mathtools}
\definecolor{Crimson}{RGB}{220,20,60} 

\newcommand{\fixedgenaligned}[1]{%
  \left\langle
  \begin{aligned}[t]
    #1
  \end{aligned}
  \right\rangle
}
\renewcommand{\fixedgenaligned}[1]{%
  \begin{aligned}[t]
    \mathopen{\langle}\; #1 \;\mathclose{\rangle}
  \end{aligned}%
}

\newcommand{\hlgen}[1]{%
  \begingroup\setlength{\fboxsep}{0.5pt}%
  \colorbox{BrickRed!10}{\ensuremath{\color{BrickRed!80!black}\mathbf{#1}}}%
  \endgroup
}

\newcommand{\gen}[1]{\langle #1 \rangle}

\newcolumntype{Y}{>{\raggedright\arraybackslash}X}

\newcommand{\sectionrow}[1]{%
  \rowcolor{gray!18}%
  {\bfseries\centering #1}\\[6pt]%
}
\renewcommand{\sectionrow}[1]{%
  \rowcolor{gray!18}%
  \makebox[\linewidth]{\bfseries #1}\\[2pt]%
}

\renewcommand{\sectionrow}[1]{%
  \rowcolor{gray!18}%
  \makebox[\linewidth]{\rule{0pt}{2.2ex}\bfseries #1}\\[2pt]%
}

\newcounter{variety}
\newcommand{\resetvarieties}{\setcounter{variety}{0}}

\newcommand{\vrowalt}[1]{%
  \ifodd\numexpr\value{variety}+1\relax
    \rowcolor{gray!6}%
  \else
    \rowcolor{gray!12}%
  \fi
  \stepcounter{variety}%
  \textbf{Variety \thevariety}\quad #1\\[6pt]%
}

\def\checkmark{\tikz\fill[scale=0.4](0,.35) -- (.25,0) -- (1,.7) -- (.25,.15) -- cycle;}

\newtheorem{example}{Example}
\DeclareMathOperator{\myht}{ht}

\title{Bridging the Gap Between Scientific Laws Derived by AI Systems and Canonical Knowledge via Abductive Inference with AI-Noether} 

\author{
  Karan Srivastava \\
  University of Wisconsin-Madison $\|$ IBM Research \\
  Madison, WI $\|$ Yorktown Heights, NY  \\
  \texttt{ksrivastava4@wisc.edu} \\
   \And
  Sanjeeb Dash\\
  IBM Research\\
  Yorktown Heights, NY \\
  \texttt{sanjeebd@us.ibm.com} 
  \And
  Ryan Cory-Wright\\
  Imperial Business School\\
  London,  UK\\
  \texttt{r.cory-wright@imperial.ac.uk}
  \And
  Barry Trager \\
  IBM Research\\
 Yorktown Heights, NY \\
  \texttt{bmt@us.ibm.com} 
    \And
\textbf{Cristina Cornelio} \\
  Samsung AI\\
  Cambridge, UK \\
  \texttt{c.cornelio@samsung.com} 
  \And
   \textbf{Lior Horesh} \\
  IBM Research\\
  Yorktown Heights, NY \\
  \texttt{lhoresh@us.ibm.com} 
}

\begin{document}

\maketitle

\begin{abstract}

Advances in AI have shown great potential in contributing to the acceleration of scientific discovery. Symbolic regression can fit interpretable models to data, but these models are not necessarily derivable from established theory. Recent systems (e.g., AI-Descartes, AI-Hilbert) enforce derivability from prior knowledge. However, when existing theories are incomplete or incorrect, these machine-generated hypotheses may fall outside the theoretical scope. Automatically finding corrections to axiom systems to close this gap remains a central challenge in scientific discovery. We propose a solution: an open-source algebraic geometry-based system that, given an incomplete axiom system expressible as polynomials and a hypothesis that the axioms cannot derive, generates a minimal set of candidate axioms that, when added to the theory, provably derive the (possibly noisy) hypothesis. We illustrate the efficacy of our approach by showing that it can reconstruct key axioms required to derive the carrier-resolved photo-Hall effect, Einstein's relativistic laws, and several other laws.

\end{abstract}

\section{Introduction: The AI Scientific Revolution and Abductive Inference}
The proliferation of artificial intelligence in accelerating scientific discovery has allowed us to vastly extend our exploration of science at a breakneck speed while also bringing a fundamental question to the forefront of research: How can we integrate AI-generated insights into canonical scientific knowledge? 
Since the release of ChatGPT in November 2022, the ubiquity of generative AI has made it easier and cheaper to make scientific discoveries \cite{baldi2014searching, guimera2020bayesian, davies2021advancing, AI-Feynman, brunton2016discovering, kubalik2020symbolic, kubalik2021multi, engle2022deterministic, bertsimas2023learning,science2, boiko2023autonomous, makke2024interpretable, reddy2025towards}. Indeed, machine learning and AI techniques have recently developed new antivenom \cite{vazquez2025novo}, new sorting algorithms \cite{mankowitz2023faster}, new drugs that have progressed to phase II clinical trials \cite{xu2025generative}, new mathematical discoveries \cite{romera2024mathematical, alphaproof2024ai}, and new equations in psychology \cite{peterson2021using}, to name a few applications. More recently, there has even been progress in integrating data and explicit background theory in the form of axioms within the process of discovery of mathematical models in science in symbolic form \cite{descartes, corywright2024evolving}.

However, the acceleration of scientific discovery poses both challenges and opportunities for the scientific community. First, with the reduced cost of making and disseminating incremental scientific discoveries, when combined with strong incentives for scientists to `publish-or-perish' \cite{parchomovsky2000publish}, the sheer volume of AI-generated and AI-assisted work threatens to overwhelm the peer-review system and obscure genuine advances amid incremental contributions or noise. For instance, the proportion of arXiv papers substantially modified by LLMs has steadily increased from around $2.5\%$ in $2022$ to around 10\% in $2024$ \cite{liang2024mapping}, with the authors who post the most preprints on arXiv being disproportionately likely to use LLMs. Second, from a scientific accuracy perspective, it is often unclear whether scientific discoveries made with generative AI or machine learning generalize well to unseen data and advance understanding \cite{krenn2022scientific, miret2024llms}, or merely fit known data well. This lack of interpretability and generalizability (see \citep{lipton2018mythos, rudin2019stop} for discussions) means that verifying scientific discoveries (e.g., via peer review) is becoming as important as the discovery process itself \cite{cornelio2025need}. Third, stacking AI-driven discoveries on top of each other poses a serious risk in domains where absolute truth matters more than predictive accuracy. This is especially true for discoveries generated by AI systems that prioritize minimizing in-sample training loss and may not generalize out-of-distribution. In such an environment, this can result in a 'scientific model collapse' in the sense of \cite{shumailov2024ai} where AI-systems stack plausible-seeming but incorrect hallucinations on top of one another, similar to the replication crisis that struck psychology in the early 2010s \cite{shrout2018psychology}. 

To address the challenges posed by and capitalize on the opportunities presented by AI-driven scientific discovery, we need a framework for \textit{systematic revision of scientific foundations} - what Kuhn called paradigm shifts \cite{kuhn1997structure}. In the life cycle of a scientific paradigm, a scientific community starts with a certain set of natural laws and beliefs (axioms). They then use these axioms and experimental data to make scientific discoveries, built on their axiomatic foundation, until the gap between the foundation and prominent discoveries becomes so apparent that a \textit{paradigm shift} to new axioms is needed. Kolmogorov's axioms in probability theory \cite{grimmett2020probability} are a good example of this process. Before Kolmogorov introduced his axioms in $1933$, discoveries were made in an ad hoc manner, leading to paradoxes such as Bertrand's paradox \cite{shackel2007bertrand}. However, following Kolmogorov's paradigm shift, it became possible to distinguish between correct and incorrect discoveries in probability theory. While the natural sciences involve more empirical complexities, Kuhnian paradigm shifts regularly occur, leading to significant breakthroughs when they do. For instance, in order to develop his theory of special relativity, Einstein necessarily had to modify fundamental postulates from prior Newtonian mechanics such as recognizing that the speed of light is constant \citep{chou2010optical}. Similarly, Maxwell's revision of Amp{\`e}re's circuital law to include a displacement term completed classical electrodynamics for time-varying fields and led to his prediction that light is an electromagnetic wave \cite{longair2015paper}.

These observations, coupled with the increasing availability of computational power driven by Moore's Law \citep{mack2011fifty}, suggest that AI and machine learning should be deployed to develop new scientific paradigms, much as AI has successfully been applied to scientific discovery. In particular, abductive inference should be employed within the scientific method to determine when a new scientific paradigm is necessary and when the set of scientific axioms used by a community needs revision. Accordingly, we propose an AI framework to solve the problem of abductive reasoning. 

\textbf{Main Contributions.} We present AI-Noether, a system that automates abductive inference to identify missing axioms in scientific theories. Given background axioms and a hypothesis that cannot be derived from them, AI-Noether systematically generates candidate axioms whose addition makes the hypothesis derivable, transforming paradigm shifts that were historically informal and human-driven into a scalable computational procedure. This enables us not only to identify discrepancies between theory and observation, but to resolve them by inferring the minimal axiomatic modifications required. We name our system \verb|AI-Noether| in honor of Emmy Noether, one of the leading mathematicians of her time who made significant contributions to both algebraic geometry and physics. We build on her work at the core of our system in order to make abductive inference possible, as we will see. This name also reflects Noether's legacy of uncovering the algebraic structures underlying physical laws, most famously via Noether's theorem on symmetries and conserved quantities \cite{noether1971invariant}, paralleling our goal of inferring the minimal additional axioms needed to render a hypothesis derivable. To support reproducibility and future research, we open-source AI-Noether's complete implementation, including all benchmark problems used in this study.

To the best of our knowledge, this is the first research work to study automated abductive reasoning in the context of scientific discovery, in which derivations may consist of fully general polynomial axioms over any set of variables; see Appendix \ref{append.litreview} for a review of existing automated approaches to abductive inference and a comparison of our system with existing AI-driven approaches to scientific discovery of mathematical models. Thus, our work constitutes a step toward the larger goal of augmenting and transforming the entire scientific method using AI by bridging the gap between machine-generated hypotheses and known theory.

\begin{figure}[t]
  \centering
  \includegraphics[width=0.8\textwidth]{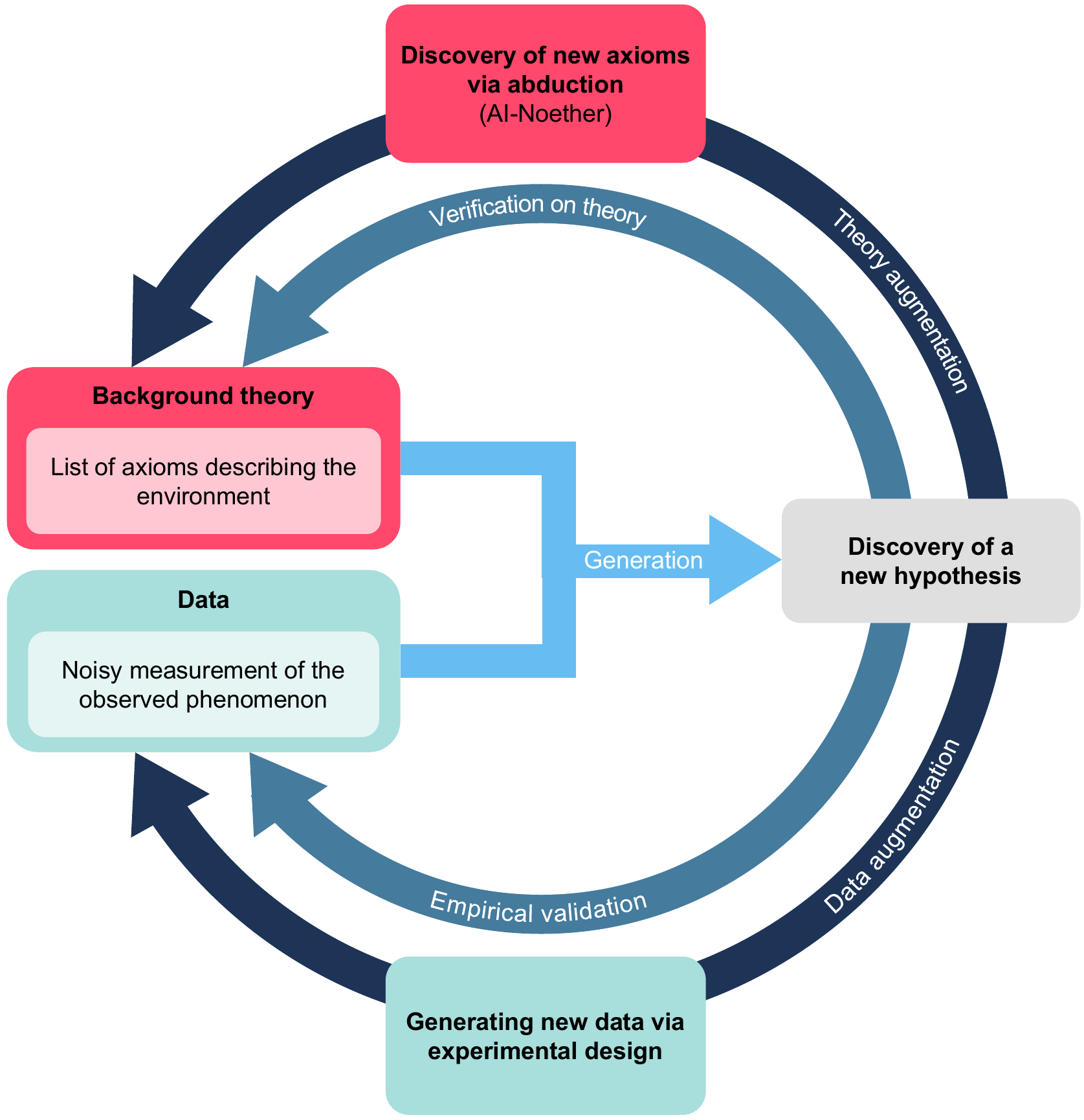} 
\caption{\textbf{AI-Noether--augmented scientific discovery loop.} Given background theory axioms and observational data, a new candidate hypothesis is generated that may or may not be compatible with the existing theory (under different notions of derivability and consistency) and may fit the data with varying approximation error. AI-Noether performs abductive inference to identify missing axioms that reconcile these hypotheses with canonical knowledge, while symmetrically, on the data-driven end, experimental design prescribes the generation of new data. Both procedures are instrumental to further refine both hypotheses and axioms, closing the discovery-verification loop.}

  \label{fig:circle}
\end{figure}

\section{Abductive Inference}\label{sec:abductive}
In the pursuit of scientific discovery, we often find ourselves confronted with a striking reality: established theoretical frameworks routinely fall short of fully explaining observable phenomena. For instance, standard models of the Big Bang theory predict that there should be about three times as much lithium in the universe as is observed \cite{fields2011primordial}. This disconnect between theory and observation presents a fundamental challenge: how do we systematically identify what is missing from our understanding?

Resolving discrepancies between theory and observation is increasingly relevant in the era of machine learning, because data-driven methods can reveal patterns that existing theory cannot explain. Thus, fully integrating insights from statistical AI into science is of considerable interest. However, discoveries from statistical AI are often difficult to verify due to their uninterpretable nature \cite{rudin2019stop, lipton2018mythos}, and integrating them into science remains an open problem. Bridging this disconnect can be expressed as the following abductive inference problem: given a set of known scientific principles (axioms) and a newly discovered relationship that cannot be derived from them, what additional principles must we postulate to bridge the gap? 

In this section, we formalize this problem for the case where both axioms and discoveries can be expressed as multivariate polynomial equations. We assume that all axioms, including any missing ones, and the hypotheses, are expressible as polynomial equations of the form $A(x)=0$. A collection of polynomial equations $A_1(x) = 0, \ldots, A_k(x) = 0$ implies the polynomial equation $Q(x) = 0$ if $Q(\bar x)=0$ for every solution $\bar x$ of the system $\{A_1, \ldots, A_k\}$.

Formally, we consider the problem:
\begin{center} 
    \textit{Given axioms $A_1,\ldots,A_k$ and a hypothesis $Q$ such that $A_1,\ldots,A_k$ do not imply $Q$, \\
    find additional $r$ axioms $A_{k+1},\ldots,A_{k+r}$ with the ``smallest residual'' relative to \\
    $A_1,\ldots,A_k$ such that the augmented system $A_1,\ldots,A_{k+r}$ implies $Q$.}
\end{center}

\textbf{Terminology:} Throughout the rest of this work, we use the terms $A_1, \ldots, A_k$ ``explain'', ``derive'' or ``imply'' $Q$ and $Q$ is a ``consequence of'' $A_1, \ldots, A_k$ synonymously. Intuitively, the notion of ``smallest residual'' corresponds to the idea that we wish to add as little new information as possible to $A_1, \ldots, A_k$ to derive $Q$; we will formalize this notion later on. We assume that no data are available for the missing axioms $A_{k+1},\ldots,A_{k+r}$; otherwise, abduction would be equivalent to hypothesis generation, a topic that has been frequently studied in scientific discovery. Under this terminology, our system generates a set of candidate axioms $\{\widehat{A}_{k+1},\ldots,\widehat{A}_l\}$ that, along with the known axioms, explain $Q$. Further, if no single candidate axiom can be added to explain $Q$, our system detects this. Current machine-assisted discovery tools either rely on partial or complete background theory, or none at all, to discover mathematical models that fit observational data, but few append or correct the theory in an automated way, and none do so at the same level of generality.

In the remainder of this section, we give an overview of our approach.

\subsection{Approach}

\begin{figure}[t]
  \centering
  \includegraphics[width=\textwidth]{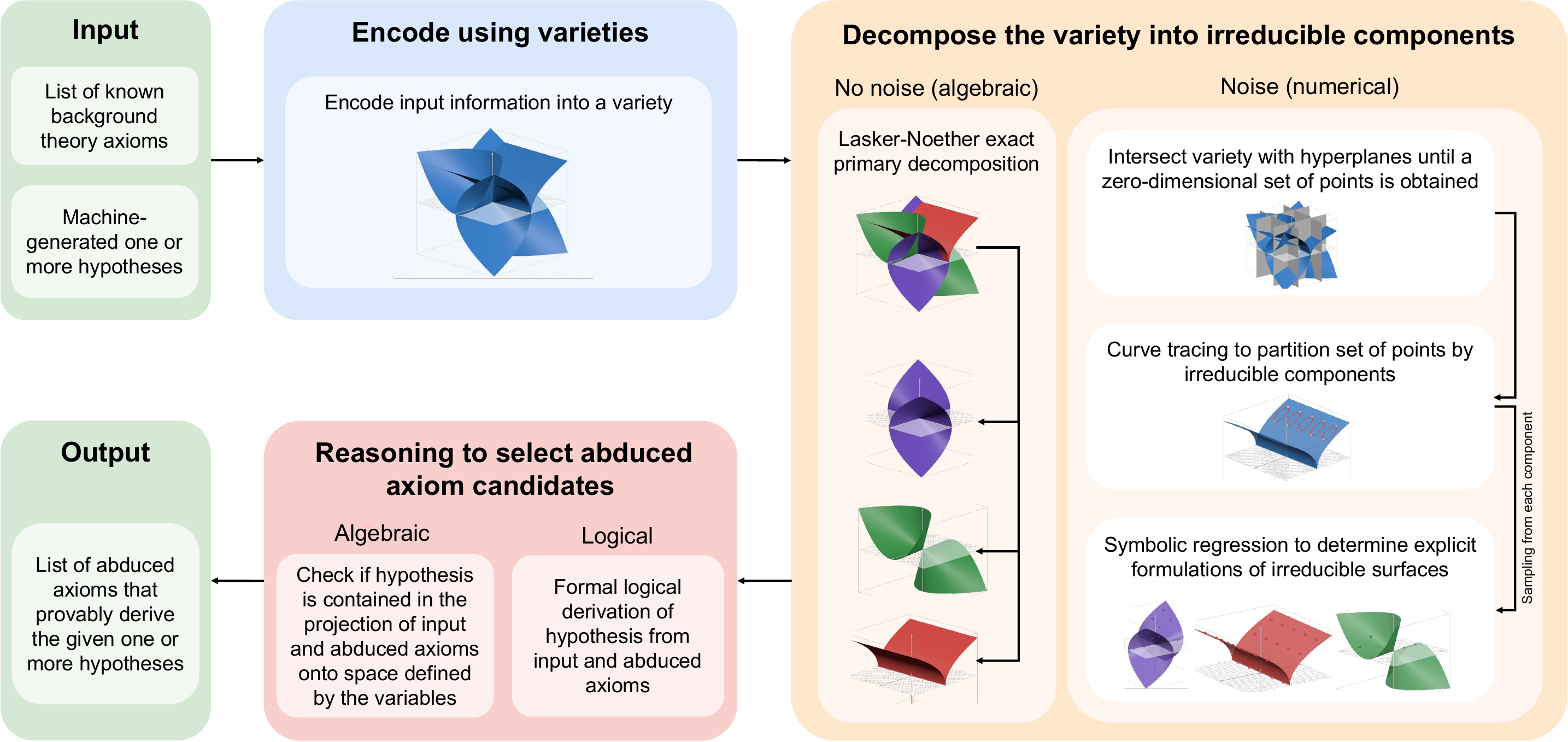} 
\caption{{\bf AI-Noether system overview.} 
Given background theory axioms and one or more candidate hypotheses that are not supported by the existing theory, AI-Noether identifies additional axioms that reconcile the hypotheses with the background knowledge. The system first encodes axioms and hypotheses into a geometric representation, decomposes this representation into fundamental components, and then reasons over these components to determine which candidate axioms, when added to the theory, are sufficient to explain the hypotheses. These axioms are returned as output after verification, which proves the hypotheses, using both algebraic and logic-based methods.
}
  \label{fig:method}
\end{figure}

Our goal is to identify candidate missing polynomial axioms $A_{k+1},\ldots,A_{k+r}$ that, along with a known set of axioms $A_1, \ldots, A_k$, can be used to derive a target hypothesis $Q$. In this study, we focus on systems defined over real polynomials with real solutions, though the approach extends naturally to complex equations (see Appendix~\ref{Ap:Complex_numbers}). The method consists of three main steps—\textbf{Encode}, \textbf{Decompose}, and \textbf{Reason}, illustrated in Figure~\ref{fig:method} and summarized in the ~\nameref{alg:high_level_inference}.

\begin{algorithm}[H]
\captionsetup{labelformat=empty} 
\caption{\textbf{Algorithm Summary}}\label{alg:high_level_inference}
\captionsetup{labelformat=default} 
\addtocounter{algorithm}{-1} 
\begin{algorithmic}[1]

\State \textbf{Input:} Axioms $A_1, \ldots, A_k$ and hypothesis $Q$
\State \textbf{Encode:} Construct the solution set (variety) of $\{A_1, \ldots, A_k, Q\}$
\State \textbf{Decompose:} Break the variety into irreducible components
\State \quad \textit{If noisy:} Use numerical methods (witness sets + symbolic regression)
\State \quad \textit{If exact:} Use symbolic primary decomposition
\For{each component}
    \State Extract candidate axioms (generators or fitted polynomials)
    \State \textbf{Reason:} Test whether candidates enable derivation of $Q$
    \State \quad \textit{If noisy:} Use theorem prover with existential quantifiers
    \State \quad \textit{If exact:} Use algebraic projection via Gr\"obner bases
\EndFor
\State \textbf{Output:} Candidate axiom sets $\{\widehat{A}_{k+1},\ldots,\widehat{A}_{l}\}$ sufficient to explain $Q$
\end{algorithmic}
\end{algorithm}

\textbf{Encode.} Given a collection of known axioms $A_1, \ldots, A_k$ and a hypothesis $Q$, we form the solution set of the system $\{A_1, \ldots, A_k, Q\}$. Each equation is defined over some subset of $n$ real variables, and so defines a hypersurface in $\mathbb{R}^n$. The intersection of these hypersurfaces defines a geometric object referred to as the \textit{variety} corresponding to the system, denoted $V(A_1,\ldots,A_k,Q)$. This geometric perspective is useful because algebraic derivability has a geometric interpretation: if $Q$ is algebraically derivable from $A_1,\ldots,A_k$, then $Q$ can be written as an algebraic combination $Q = \sum_{i=1}^k \alpha_i A_i$ where each $\alpha_i$ is a real polynomial. Then $Q$ vanishes whenever all the axioms vanish. So we can recast the idea of $Q$ being derivable from any set of axioms $A_1, \ldots, A_{k+r}$ as a geometric operation. Our first step in AI-Noether's process then will be to take the known axioms $A_1,\ldots,A_k$ and consequence $Q$ and \textit{encode} it into a variety. 

\begin{figure}[h]
  \centering
  \includegraphics[width=0.95\textwidth, trim={1cm 6cm 0cm 6cm}, clip]{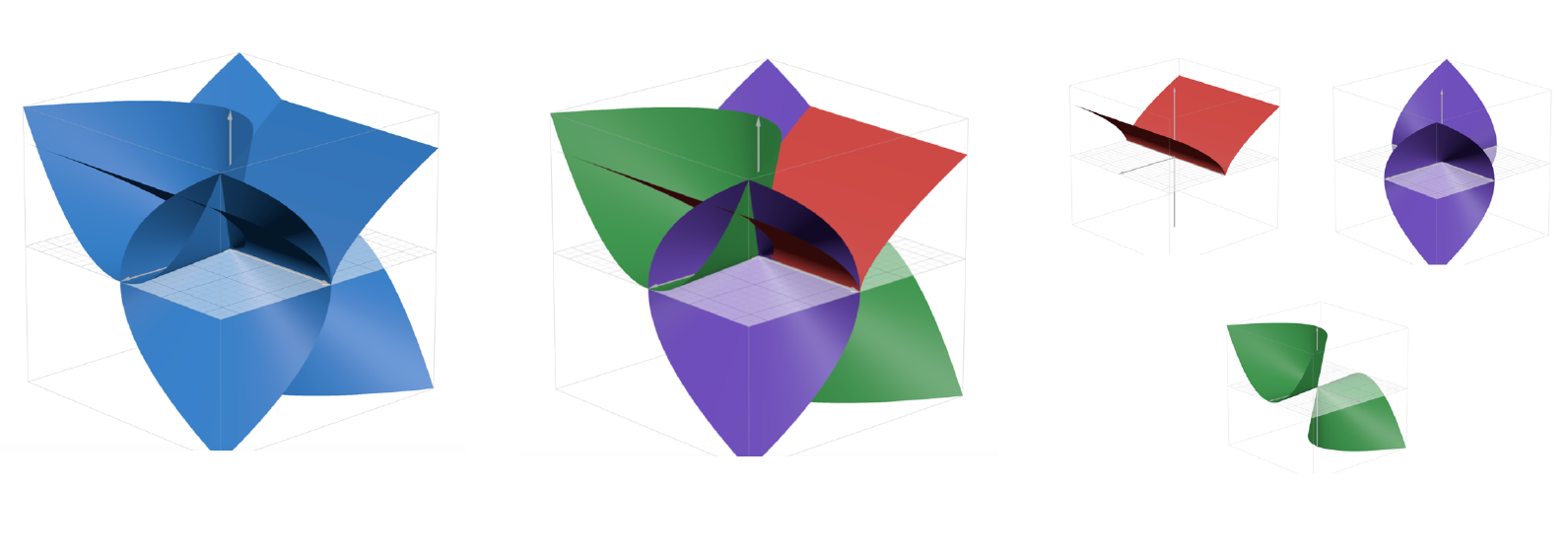} 
\caption{{\bf Geometric visualization of decomposition.} On the left, we have an example of a variety which is the solution to a system of polynomial equations. The middle and right panels show that this variety can be decomposed into irreducible components, which, geometrically, is the process of finding smaller varieties whose union is the original variety.}
  \label{fig:decompose}
\end{figure}

\textbf{Decompose.} The variety corresponding to the equation $xy = 0$ (where $x$ and $y$ are real variables) is \textit{reducible} as it is the union of two smaller varieties, i.e., solutions of $x = 0$ and of $y = 0$. We cannot further write either of these smaller varieties as the union of even smaller varieties, so they are \textit{irreducible}. In other words, $V(xy) = V(x) \cup V(y)$. In writing the variety $V(xy)$ as a union of two irreducible varieties, we have \textit{decomposed} it. We similarly decompose the variety $V(A_1, \ldots, A_k, Q)$ into its irreducible components representing minimal geometric subsets that cannot be decomposed further. This process is called \textit{Lasker-Noether Primary Decomposition} \cite{atiyahmacdonald1969}, and is geometrically illustrated in Figure \ref{fig:decompose}. 

If $Q$ is not a consequence of the axioms (i.e., is not derivable from the known axioms), then $V(A_1,\ldots,A_k,Q)$ contains additional irreducible components not present in $V(A_1,\ldots,A_k)$. To illustrate why this matters in the context of abductive inference, assume for example that we only need one more axiom, $A_{k+1}$ in order to derive $Q$. Then $Q = \alpha_1 A_1 + \cdots + \alpha_{k+1} A_{k+1}$ for unknown coefficients $\alpha_i$ and unknown $A_{k+1}$. We are of course interested then in the residual $\alpha_{k+1}A_{k+1}$. The key observation is that $V(A_1,\ldots,A_k,Q)$ is the same variety as $V(A_1,\ldots, A_k,\alpha_{k+1}A_{k+1})$. Similar to the example of $V(xy)$, $V(\alpha_{k+1}A_{k+1})$ is reducible. So intersecting $Q$ with the variety corresponding to the axioms $V(A_1,\ldots,A_k)$ introduces new irreducible components that are directly related to the residual $\alpha_{k+1}A_{k+1}$. Decomposition isolates these structures. So the next step in AI-Noether's system is to take the encoded variety and \textit{decompose} it into its irreducible components. The key difference from traditional symbolic regression-based discovery is that we do not assume any data for the unknown axiom or any prior knowledge of the variables being fitted. 

When consequences contain noise - for instance, when $Q$ has been inferred from experimental data - exact algebraic methods become numerically unstable. In such cases, we employ a numerical irreducible decomposition \cite{bates2024numericalnonlinearalgebra} via homotopy continuation, which intersects the variety with generic hyperplanes until a zero-dimensional witness set is obtained, then uses curve tracing to identify points lying on the same irreducible component. This produces a finite set of sample points on each component rather than symbolic generators. We then perform symbolic regression on these witness points to recover polynomial relations that approximately vanish on each component with an approach similar to vanishing component analysis \cite{pmlr-v28-livni13}. The smallest singular value of the resulting design matrix serves as a quality metric: low values indicate the recovered relation nearly vanishes on the component, suggesting it governs that portion of the solution space. This numerical workflow, with the numerical witness set computation and our symbolic regression step, enables decomposition even when consequences carry numerical error. This workflow is detailed in Appendix~\ref{sec:noise} 

\begin{figure}[h]
  \centering
  \includegraphics[width=\textwidth, trim={2cm 6cm 2cm 6cm}, clip]{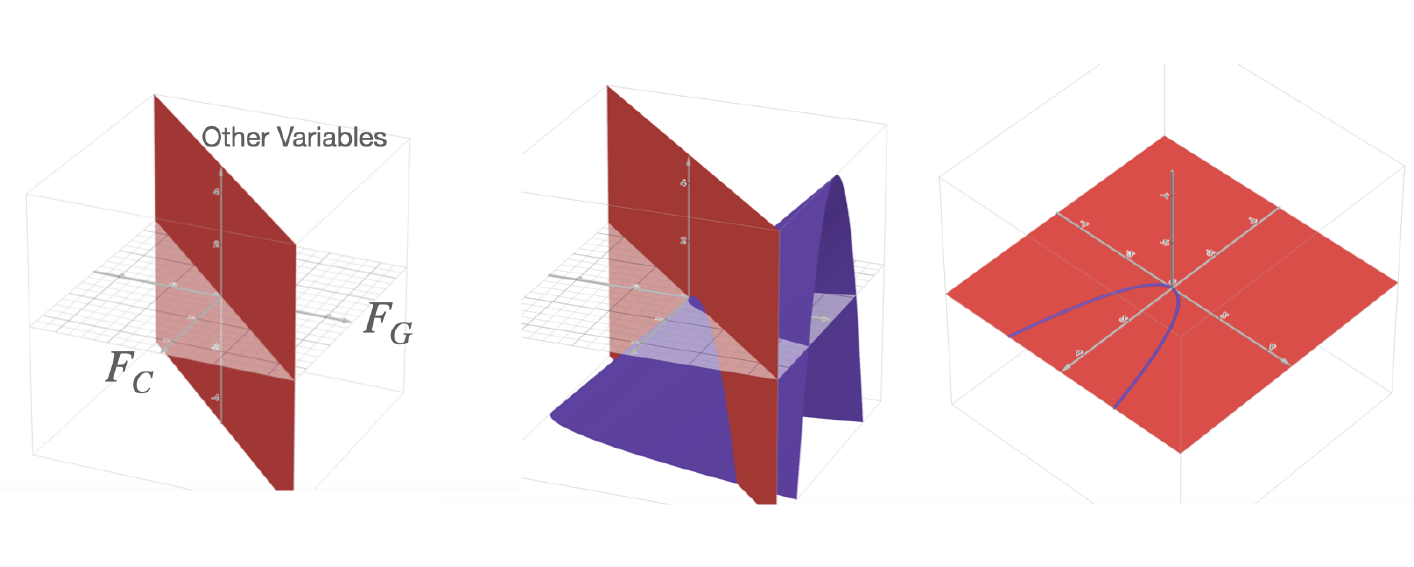} 
\caption{{\bf Geometric visualization of algebraic reasoning.} On the left, we have some surface defined by an axiom (here, $F_c=F_g$ as an example). In the middle, we have an additional equation that defines a new surface. In order to apply the axiom to the new surface, we project the surface onto the space defined by the known axiom(s).}
  \label{fig:reason}
\end{figure}

\textbf{Reason.} For each irreducible component, we extract its defining polynomials and test whether any subset $\{\widehat{A}_{k+1},\ldots,\widehat{A}_l\}$, when added to the known axioms, is sufficient to derive $Q$. This is done by forming the variety $V(A_1, \ldots, A_k, \widehat{A}_{k+1},\ldots,\widehat{A}_l)$ and projecting it onto the variables appearing in $Q$, illustrated in Figure \ref{fig:reason}. If this projection lies within the solution set of $Q$, then $Q$ is implied by the augmented system. Such sets are returned as candidate missing axioms. If no such set exists, we conclude that no single missing axiom explains $Q$. This is the final step of AI-Noether where we \textit{reason} with each decomposed component to discover which axiom candidates can derive $Q$. 

When axioms or consequences contain numerical constants from noisy data, algebraic projection may be overly sensitive to coefficient errors. As an alternative to algebraic derivability, for robustness to noise, we use a theorem prover to assess whether the conjecture follows logically from the augmented background theory (the original axioms together with the abduced ones). Building on AI-Descartes~\cite{descartes}, which introduced existential derivability to handle numerical constants in data-derived conjectures, we introduce the existential constants also in the abduced axioms. The conjecture is then deemed derivable when there exists an instantiation of these abstracted constants that makes it logically implied by the background theory. This extension allows the prover to handle noisy numerical parameters and provides a principled way to test whether the augmented axioms suffice to entail the conjecture.

Using the \textit{encode}, \textit{decompose}, and \textit{reason} steps, AI-Noether handles both exact symbolic computation and noisy numerical scenarios. For a detailed development of the algebraic and logical foundations underlying the three-step pipeline, see Appendix~\ref{sec:method_details}. For the details of our implementation, see Appendix~\ref{algorithm_details}. Having established the conceptual framework, we now demonstrate AI-Noether's capabilities across a range of scientific domains. Section~\ref{sec:results} presents experimental results organized by discovery type: recovering individual axioms (Section~\ref{sec:single_axiom}), discovering multiple axioms simultaneously (Section~\ref{sec:mult_missing_axioms} and \ref{sec:mult:consequences}), and handling noisy consequences (Section~\ref{sec:noise_experiments}). These experiments span classical mechanics, special relativity, orbital dynamics, and electromagnetic theory, illustrating the generality of the abductive inference approach.

\section{Results}\label{sec:results}

In this section, we present the results from testing our system on various real-world physical theories. We test AI-Noether on $9$ systems reported in AI-Hilbert \cite{corywright2024evolving} as well as 3 additional systems of varying complexity, the most complex being the results from a recent work on the Carrier-Resolved Photo-Hall Effect \cite{phothall}. The other two systems include a combined axiomatic system involving Einstein's relativistic mass, length contraction, and time dilation laws as consequences, and finally a baseline case of a simple harmonic oscillator. We illustrate how AI-Noether performs under different settings using different examples. 

\subsection{Single Missing Axiom}\label{sec:single_axiom}

\subsection*{Example: Carrier-Resolved Photo-Hall Effect}

\begin{figure}[t]
  \centering
  \includegraphics[
    width=1.5\textwidth,
    height=0.2\textheight,
    keepaspectratio
  ]{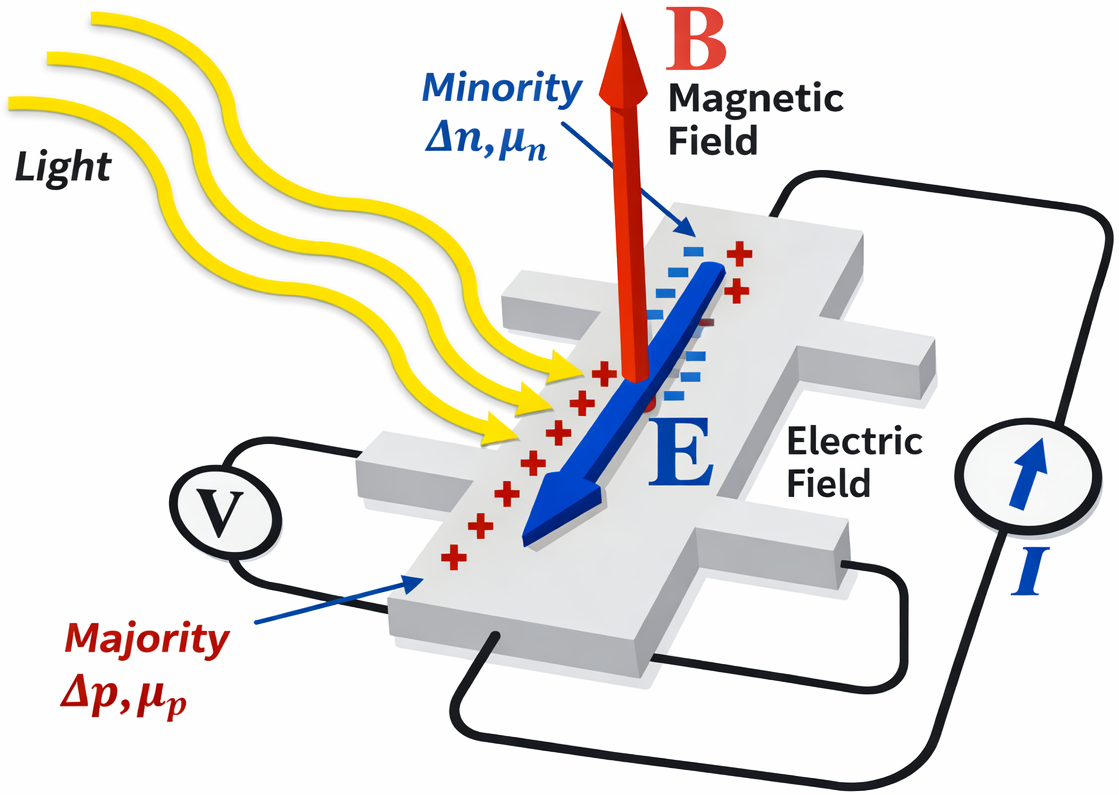}
  \captionsetup{font=small}
  \caption{\textbf{Illustration of the experimental setup for the Carrier-Resolved Photo-Hall Effect}. AI-Noether successfully recovers the necessary Hall coefficient relation required to prove the relationship between the various parameters of a semiconducting surface. }
  \label{fig:photohall}
\end{figure}

The Carrier-Resolved Photo-Hall effect \cite{phothall} equation describes the relationship between various parameters of a semiconducting surface, illustrated in Figure \ref{fig:photohall} and is given by:

\[
Q \;:=\; H-\frac{r\,e\,\mu_P^{\,2}\bigl[p_0+\Delta n(1-\beta^2)\bigr]}{\sigma^{2}}=0,
\]
where \(\sigma=e\mu_P\bigl[p_0+\Delta n(1+\beta)\bigr]\) is the conductivity. We assume the axioms
\begin{align*}
A_1 &: \beta \mu_P - \mu_N & \text{ Mobility ratio definition} \\
A_2 &: \mu_H - r \mu & \text{ Hall factor correction} \\
A_3 &: p_h - p_0 - \Delta p & \text{ Hole concentration balance} \\
A_4 &: n - \Delta n & \text{ Electron concentration tracking}\\
A_5 &: \Delta p - \Delta n & \text{ Charge neutrality condition}\\
A_6 &: \sigma - e p_h \mu_P - e n \mu_N & \text{ Total conductivity sum}\\
A_7 &: H (p_h + \beta n)^2 e - r p_h + r \beta^2 n & \text{ Hall coefficient relation.}
\end{align*}
Here \(\mu_N\) and \(\mu_P\) are electron and hole drift mobilities, \(\beta=\mu_N/\mu_P\) is the mobility ratio (encoded by \(A_1\)), \(\mu_H\) is the Hall mobility with Hall scattering factor $r$ and mobility $\mu$ related by \(r=\mu_H/\mu\) (encoded by \(A_2\)), \(\Delta n\) and \(\Delta p\) are the electron and hole photo-carrier densities (equal in steady state by \(A_5\)), \(H\) is the Hall coefficient, \(\sigma\) is the conductivity, \(n\) is the electron density, \(p_0\) is the background hole density, and \(p_h\) is the hole density (for a p-type material).

The photo-Hall effect measurement begins with the semiconductor in its dark equilibrium state, where only majority carriers (e.g., holes with concentration $p_0$ in p-type material) contribute to transport, establishing baseline conductivity $\sigma_0$ and Hall coefficient $H_0$. Upon illumination with above-bandgap photons, electron-hole pairs are generated satisfying charge neutrality $\Delta n = \Delta p$, increasing the total hole concentration to $p_h = p_0 + \Delta p$ and electron concentration to $n \approx \Delta n$. 

An alternating current (AC) magnetic field from the rotating parallel dipole line (PDL) system deflects both carrier types via the Lorentz force, generating a Hall voltage across the sample. As light intensity $I$ varies, both conductivity $\sigma(I) = e(p_h \mu_P + n \mu_N)$ and Hall coefficient $H(I)$ evolve non-linearly due to competing contributions from holes and electrons with different mobilities. The key innovation is the relation $\Delta\mu_H = \frac{d(\sigma^2 H)}{d\sigma}$, which extracts the hole-electron mobility difference from the measured transport data.

An exact derivation of $Q$ from the axioms, both directly and as an algebraic combination, can be found in Appendix \ref{Ap:photoHall}. However, it suffices to recognize that, because $Q$ is algebraically derivable from the known axioms, we can write

\[
Q = \sum_{i=1}^7 \alpha_i A_{i}
\]

where the $\alpha_i$ are polynomials (Appendix \ref{Ap:photoHall}) and $i\in \{1,\ldots,7\}$. 

One of the key axioms in the derivation is axiom $A_7$, which relates the Hall coefficient to all the essential quantities. In our setting, we assume that we neither have $A_7$ nor any of the multipliers $\alpha_1,\ldots,\alpha_7$. Using methods described in \cite{phothall, AI-Feynman, descartes, corywright2024evolving}, we can recover $Q$ from data. Encoding the known information $A_1,\ldots,A_6,Q$ into a variety, decomposing each generator, and passing it through the reasoning step yields the following polynomials.

\resetvarieties
\begingroup
\begin{tabularx}{\linewidth}{@{}Y@{}}
\sectionrow{Decomposition}
\vrowalt{\(\gen{\sigma,\; \mu_P,\; n-\Delta n,\; \mu_N,\; \Delta p-\Delta n,\; ph-p0-\Delta n,\; \mu r-\mu_H}\)}
\vrowalt{\(\gen{\sigma,\; e,\; n-\Delta n,\; \Delta p-\Delta n,\; ph-p0-\Delta n,\; \mu_N-\mu_P\beta,\; \mu r-\mu_H}\)}
\vrowalt{%
\(\fixedgenaligned{
  &n-\Delta n,\; \Delta p-\Delta n,\; ph-p0-\Delta n,\; \mu_N-\mu_P\beta,\; \mu r-\mu_H,\\
  &\mu_N e \Delta n + e\mu_P p_0 + e\mu_P \Delta n - \sigma,\;
    e\mu_P p_0 + e\mu_P \Delta n\beta + e\mu_P \Delta n - \sigma,\\
  &\hlgen{\sigma H\bigl(p_0+\Delta n(1+\beta)\bigr)\;+\;r\bigl(\mu_N\beta\Delta n-\mu_P(p_0+\Delta n)\bigr)},\\
  &\mu_N^2 \Delta n r+\mu_N \Delta n\sigma H-\mu_P^2 p_0 r-\mu_P^2 \Delta n r+\mu_P p_0\sigma H+\mu_P \Delta n\sigma H,\\
  &\hlgen{r\mu_P\bigl(p_0+\Delta n(1-\beta^2)\bigr)\;-\;\sigma H\bigl(p_0+\Delta n(1+\beta)\bigr)},\\
  &\hlgen{-\,e\sigma H\bigl(p_0+\Delta n(1+\beta)\bigr)\;+\;r\bigl(\sigma(1-\beta)+e\mu_P\beta p_0\bigr)},\\
  &\mu_N e\mu_P p_0 r-\mu_P\beta r\sigma+\mu_P r\sigma-\sigma^2 H,\;
    \mu_N^2 e p_0 r-\mu_P\beta^2 r\sigma+\mu_P\beta r\sigma-\beta\sigma^2 H,\\
  &\mu\sigma^2 H-\mu_N\mu_H e\mu_P p_0+\mu_H\mu_P\beta\sigma-\mu_H\mu_P\sigma,\;
    \mu\beta\sigma^2 H-\mu_N^2\mu_H e p_0+\mu_H\mu_P\beta^2\sigma-\mu_H\mu_P\beta\sigma,\\
  &\mu p_0\sigma H+\mu \Delta n\beta\sigma H+\mu \Delta n\sigma H-\mu_H\mu_P p_0+\mu_H\mu_P \Delta n\beta^2-\mu_H\mu_P \Delta n,\\
  &\mu\mu_N \Delta n\sigma H+\mu\mu_P p_0\sigma H+\mu\mu_P \Delta n\sigma H+\mu_N\mu_H\mu_P \Delta n\beta-\mu_H\mu_P^2 p_0-\mu_H\mu_P^2 \Delta n,\\
  &\hlgen{eH\bigl(p_0+\Delta n(1+\beta)\bigr)^2\;-\;r\bigl(p_0+\Delta n(1-\beta^2)\bigr)}
}\)%
}
\sectionrow{Generators selected by Reasoning module}
\rowcolor{gray!6}
\makebox[\linewidth]{\(\color{BrickRed!80!black}\mathbf{\sigma H\bigl(p_0+\Delta n(1+\beta)\bigr)\;+\;r\bigl(\mu_N\beta\Delta n-\mu_P(p_0+\Delta n)\bigr)}\)}\\[4pt]
\rowcolor{gray!12}
\makebox[\linewidth]{\(\color{BrickRed!80!black}\mathbf{r\mu_P\bigl(p_0+\Delta n(1-\beta^2)\bigr)\;-\;\sigma H\bigl(p_0+\Delta n(1+\beta)\bigr)}\)}\\[4pt]
\rowcolor{gray!6}
\makebox[\linewidth]{\(\color{BrickRed!80!black}\mathbf{-\,e\sigma H\bigl(p_0+\Delta n(1+\beta)\bigr)\;+\;r\bigl(\sigma(1-\beta)+e\mu_P\beta p_0\bigr)}\)}\\[4pt]
\rowcolor{gray!12}
\makebox[\linewidth]{\(\color{BrickRed!80!black}\mathbf{eH\bigl(p_0+\Delta n(1+\beta)\bigr)^2\;-\;r\bigl(p_0+\Delta n(1-\beta^2)\bigr)}\)}\\[4pt]
\end{tabularx}
\endgroup

Notice that each of the obtained relationships is the exact Hall coefficient relation with other axiom substitutions applied. The first equation is $A_7$ with $A_3,A_4,A_5$ substituted and multiplied by $\mu_P$. The second is the same as the first, with $A_1$ substituted in as well. The third is the same as the second, except that $A_6$ is substituted and multiplied by $e \mu_P$. The last is the same as the first, with $A_1$ and $A_2$ substituted. 

Each of these axioms contains the same information as $A_7$. The alternate expressions reflect the fact that the expression of $A_7$ is not unique. We could have defined the same relation in multiple forms. 

Additionally, if we start with an overly strong assumption, such as $\sigma=0$ or $\mu_P=0$ from Variety 1, we could also derive $Q = 0$. We provide an additional layer of filtering for such cases when using projections to ensure that we do not eliminate more variables than necessary. A discussion of this can be found in Appendix~\ref{Ap:filtering}.

\subsection*{Example: Simple Harmonic Oscillator}

\begin{figure}[t]
  \centering
  \includegraphics[
    width=\textwidth,
    height=0.2\textheight,
    keepaspectratio
  ]{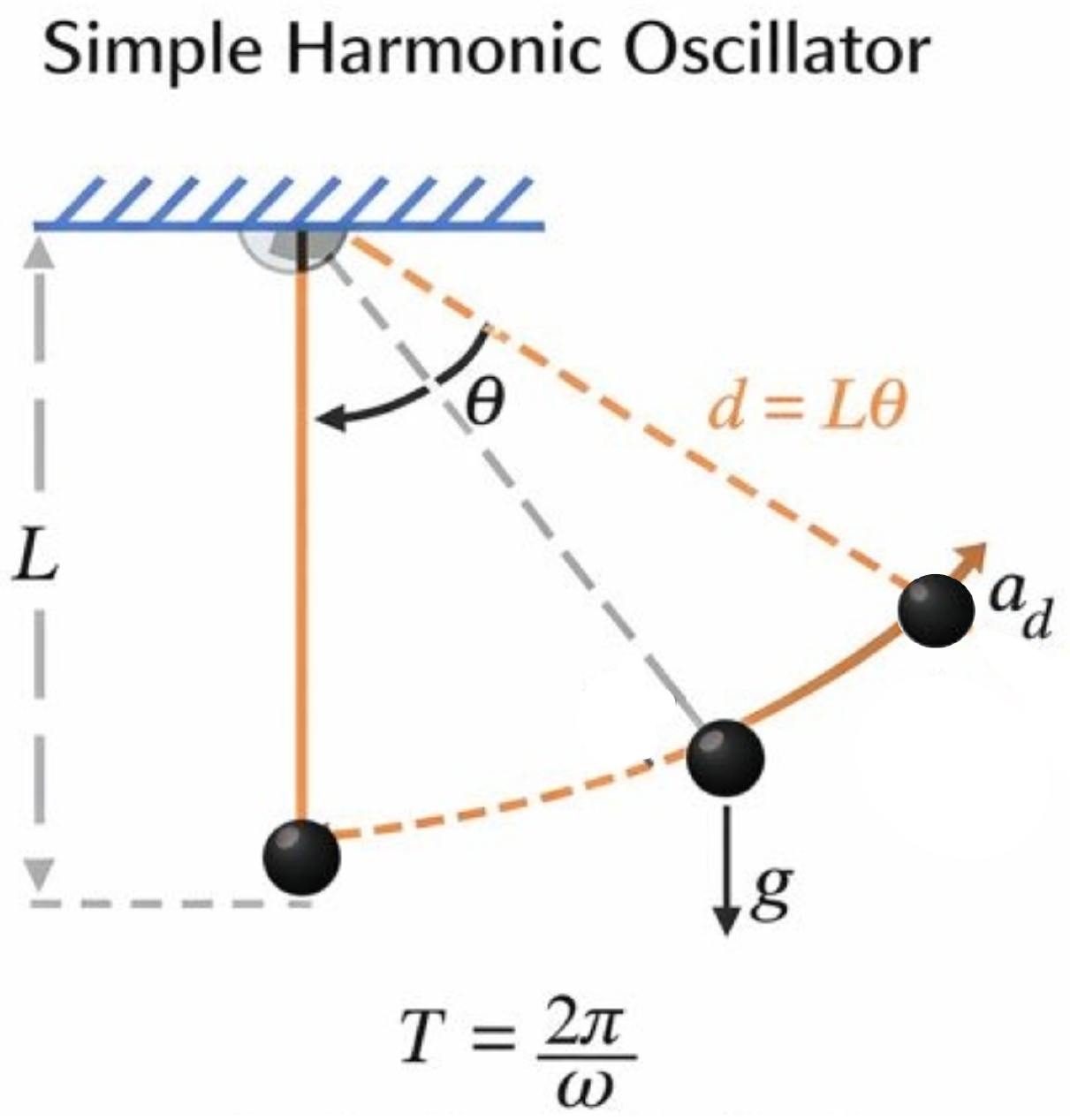}
  \captionsetup{font=small}
  \caption{\textbf{Illustration of the experimental setup for a simple harmonic oscillator.} AI-Noether can infer the requirement of the small angle assumption, gravitational component of acceleration, and any of the other essential axioms required to derive the period of a simple harmonic oscillator. This simpler example illustrates the output of AI-Noether's modules.}
  \label{fig:harmonic_oscillator}
\end{figure}

For a simpler and more illustrative example, consider the case of a simple harmonic oscillator with a point mass $m$ and length $L$, with a force $F$ acting on it creating a small angle $\theta$, illustrated in Figure \ref{fig:harmonic_oscillator}. Taking $g$ as the acceleration constant, at $j$ iterations, we have the equation that describes the time period $T_j$ given by:
\[
Q:= T_j- 2\pi j \sqrt{\frac{L}{g}}
\]
whose polynomial form, $gT_j^2-4\pi^2j^2L = 0$, can be derived (either traditionally or as an algebraic combination) from the following axioms:

\begin{align*}
A_1 &: a_d - g\sin{\theta} & \text{Gravitational component of acceleration} \\
A_2 &: d-L\theta & \text{Height of oscillator} \\
A_3 &: T\omega-2\pi & \text{Frequency-period relation} \\
A_4 &: d\omega^2-a_d & \text{Acceleration equation}\\
A_5 &: T_j-jT & \text{Time period over $j$ iterations}\\
A_6 &: \sin{\theta} - \theta & \text{Small angle assumption}\\
\end{align*}
where $a_d$ is the acceleration of the object, $\omega$ is the frequency, and $T$ is the time period for one oscillation. A small detail here is that in the algebraic context, like in \cite{corywright2024evolving}, we treat $\sin{\theta}$ and $\pi$ symbolically here. One of the key axioms here is $A_1$, the gravitational component of acceleration. If we remove this assumption from our system, then we can still recover $Q$ from data, as seen in \cite{AI-Feynman}. 
Applying AI-Noether to $Q$ and the known axioms results in the following output: 

\resetvarieties
\begingroup
\begin{tabularx}{\linewidth}{@{}Y@{}}
\sectionrow{Decomposition}
\vrowalt{%
\(\fixedgenaligned{
  &\theta-\mathrm{\sin{\theta}},\; 2j\pi-\omega T_j,\; Tj-T_j,\; \mathrm{\sin{\theta}}L-d,\; dg-ad\,L,\; \hlgen{\mathrm{\sin{\theta}}\,g-ad},\\
  &T\omega-2\pi,\; 2\omega L\pi-Tg,\; 2\omega d\pi-ad\,T,\; \omega^2L-g,\; T^2g-4L\pi^2,\; \omega^2d-ad,\; ad\,T^2-4d\pi^2
}\)%
}
\sectionrow{Generators selected by Reasoning module}
\rowcolor{gray!6}
\makebox[\linewidth]{\(\color{BrickRed!80!black}\mathbf{\mathrm{\sin{\theta}}\,g-ad}\)}\\[4pt]
\end{tabularx}
\endgroup

\textbf{Results summary on other problems}

\begingroup
\arrayrulecolor{black}
\begin{table}[t]
\centering
\footnotesize
\renewcommand{\arraystretch}{1.3}
\begin{tabular}{l@{\hspace{12pt}}c@{\hspace{12pt}}c@{\hspace{12pt}}c@{\hspace{12pt}}c}
\toprule
\rowcolor{blue!20}
\textbf{Problem} & \textbf{Axioms} & \textbf{Success Rate} & \textbf{Variables} & \textbf{Max Degree} \\
\midrule
Hagen--Poiseuille Equation     & 4  & 100\% & 9  & 4 \\
\rowcolor{blue!5}
Kepler                         & 5  & 100\% & 8  & 4 \\
Time Dilation                  & 5  & 100\% & 8  & 4 \\
\rowcolor{blue!5}
Escape Velocity                & 5  & 100\% & 9  & 4 \\
Light Damping                  & 5  & 100\% & 10 & 8 \\
\rowcolor{blue!5}
Neutrino Decay                 & 5  & 100\% & 8  & 2 \\
Simple Harmonic Oscillator     & 6  & 100\% & 11 & 5 \\
\rowcolor{blue!5}
Carrier-Resolved Photo-Hall    & 7  & 100\% & 15 & 6 \\
Relativistic Laws              & 7  & 71\%* & 12 & 5 \\
\rowcolor{blue!5}
Hall Effect                    & 7  & 100\% & 16 & 4 \\
Inelastic Relativistic Collision            & 9  & 100\% & 11 & 6 \\
\rowcolor{blue!5}
Compton Scattering             & 10 & 100\% & 15 & 6 \\
\bottomrule
\end{tabular}
\caption{Summary of Axiom Recovery Performance by Problem. In each case, we iterate through every axiom, remove it from the theory, and attempt to recover the axiom with the remaining information. In almost every case, we are able to recover every axiom. *Indicates partial recovery due to recovering alternate axioms. We discuss this particular case further in Section \ref{sec:mult:consequences}.}
\label{tab:single_ax_summary}
\end{table}
\endgroup

We report the performance of AI-Noether on each test system in Table \ref{tab:single_ax_summary}. In each case, we iterate through the axioms, omit an axiom from the system, and test if we can recover the missing axiom while not making any assumption about it. In almost every case, we were able to recover the missing axiom. Appendix \ref{Ap:single_axiom_table} provides a further discussion of some of these systems. We will discuss the problem of the relativistic laws in Section \ref{sec:mult:consequences}, where we only partially recover missing axioms correctly while recovering alternate axioms for some cases.

\subsection{Multiple Missing Axioms}\label{sec:mult_missing_axioms}

We now examine the problem of recovering more than one axiom to explain a hypothesis $Q$. In this case, we have 

$$Q = \sum_{i=1}^{k+r} \alpha_i A_{i}$$ 

However, we only have access to $A_1,\ldots,A_{k}$. We seek a candidate set of axioms that, together with $A_1,\ldots,A_{k}$, suffices to derive $Q$.

\subsection*{Example: Kepler's third law}

\begin{figure}[t]
  \centering
  \includegraphics[width=1\textwidth, trim={0cm 6cm 3cm 5cm}, clip]{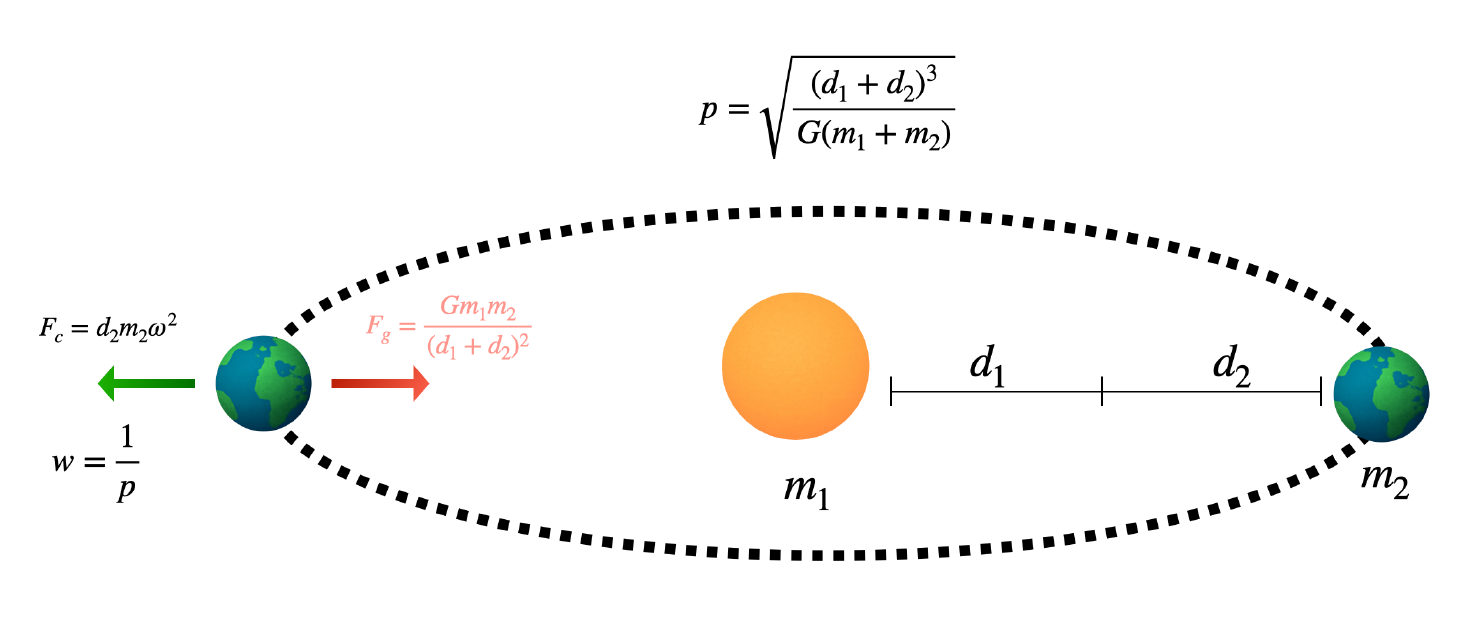} 
  \captionsetup{font=small} 
  \caption{\textbf{Illustration of the axioms required to derive Kepler's third law of planetary motion.} AI-Noether detects the existence of gravitational force, an essential axiom required to derive Kepler's third law. When multiple axioms are missing, AI-Noether can also recover coupled versions of the missing axioms.}
  \label{fig:kepler_2}
\end{figure}

Consider Kepler's third law, which relates the orbital period of two celestial bodies to their distances and masses. It can be written as
\[
  Q := p - \sqrt{\frac{(d_1+d_2)^3}{G(m_1+m_2)}} = 0,
\]
where $p$ is the orbital period (scaled by a factor of $2\pi$), $m_1,m_2$ the masses, and $d_1,d_2$ their distances from the center of mass. As in \cite{corywright2024evolving}, we will scale units to have $G=1$ for consistency with their framework, though this is not necessary. For circular orbits, we use the following axioms (with $p$ non-dimensionalized so that $\omega p=1$), illustrated in Figure \ref{fig:kepler_2}:
\begin{align*}
    A_1 &: (d_1+d_2)^2F_g - m_1m_2 & \text{Gravitational Force}\\
    A_2 &: F_c - m_2d_2\omega^2 & \text{Centrifugal Force}\\
    A_3 &: F_g - F_c &\text{Newton's third law}\\
    A_4 &: \omega p - 1 &\text{Frequency-period relation}\\
    A_5 &: m_1d_1 - m_2d_2  &\text{Center of mass equation}
\end{align*}
to derive
\[
  (m_1+m_2)p^2 - (d_1+d_2)^3 = 0,
\]
which is Kepler's law in polynomial form. 

Assume that we are missing axioms $A_3$ and $A_4$ - the relationship between centrifugal and gravitational force and the relationship between frequency and period. Then, encoding the known information - equations $A_1$, $A_2$, $A_5$, and $Q$ - our system returns the following relevant component of the primary decomposition and selected generator from the reasoning module:

\resetvarieties
\begingroup
\setlength{\tabcolsep}{8pt}%
\renewcommand{\arraystretch}{1.28}%
\arrayrulecolor{white}%

\begin{tabularx}{\linewidth}{@{}Y@{}}
\sectionrow{Selected varieties from Decomposition module}

\vrowalt{%
\(\fixedgenaligned{
  m_1 d_1 - m_2 d_2,\; \hlgen{F_g p^2 - m_2 d_2},\;
  F_g(d_1^2 + 2 d_1 d_2 + d_2^2) - m_1 m_2, \\
  \qquad d_1^3 + 2 d_1^2 d_2 + d_1 d_2^2 - m_2 p^2,\;
  m_1 p^2 - d_1^2 d_2 - 2 d_1 d_2^2 - d_2^3, \\
  \qquad F_c - \omega^2 m_2 d_2,\;
  F_c p^2 - \omega^2 d_1^3 d_2 - 2 \omega^2 d_1^2 d_2^2 - \omega^2 d_1 d_2^3
}\)%
}

\sectionrow{Generators selected by Reasoning module}
\rowcolor{gray!6}
\makebox[\linewidth]{\(\color{BrickRed!80!black}\mathbf{F_g p^2 - m_2 d_2}\)}\\[4pt]

\end{tabularx}
\endgroup

We return {\(\color{BrickRed!80!black}\mathbf{F_g p^2 - m_2 d_2}\)}. This equation is the same as the centrifugal force equation $A_2$ but with the missing relationships $F_c=F_g$ and $\omega = \frac{1}{p}$ substituted in. Therefore, the missing information is discovered, since equation $A_2$ is assumed to be known, although it is not decoupled. To go one step further, we can remove three out of the five axioms: equations $A_2$, $A_3$, and $A_4$, leaving us with only the gravitational force equation $A_1$ and the center of mass equation $A_5$. Similarly, running our system with the Kepler hypothesis gives us a similar component:

\resetvarieties
\begingroup
\setlength{\tabcolsep}{8pt}%
\renewcommand{\arraystretch}{1.28}%
\arrayrulecolor{white}%

\begin{tabularx}{\linewidth}{@{}Y@{}}
\sectionrow{Selected varieties from Decomposition module}

\vrowalt{%
\(\fixedgenaligned{
  m_1 d_1 - m_2 d_2,\; \hlgen{F_g p^2 - m_2 d_2},\;
  F_g(d_1^2 + 2 d_1 d_2 + d_2^2) - m_1 m_2, \\
  \qquad d_1^3 + 2 d_1^2 d_2 + d_1 d_2^2 - m_2 p^2,\;
  m_1 p^2 - d_1^2 d_2 - 2 d_1 d_2^2 - d_2^3
}\)%
}

\sectionrow{Generators selected by Reasoning module}
\rowcolor{gray!6}
\makebox[\linewidth]{\(\color{BrickRed!80!black}\mathbf{F_g p^2 - m_2 d_2}\)}\\[4pt]

\end{tabularx}
\endgroup

We get the same axiom candidate. However, in the previous case, where we knew the centrifugal force equation, we now assume that we do not. So we are not only recovering the relationships $F_c=F_g$ and $\omega p=1$, but also the centrifugal force axiom itself $F_c=m_2d_2\omega^2$ with the former relations coupled with it. A criterion for exactly recovering axioms is given in section \ref{recoverability}. 

Table  \ref{tab:Kepler_multiple_missing_axiom_results} shows the results of attempting to recover various combinations of pairs and triples of axioms for Kepler's third law. 

\begingroup
\arrayrulecolor{black}
\begin{table}[t]
    \centering
    \renewcommand{\arraystretch}{1.2}
    \begin{tabular}{l|c|l|c}
        \toprule
        \multicolumn{2}{c|}{\textbf{Pairs of missing axioms}} & \multicolumn{2}{c}{\textbf{Triples of missing axioms}} \\
        \textbf{Missing Axioms} & \textbf{Recovered} & \textbf{Missing Axioms} & \textbf{Recovered} \\
        \midrule
        $\{\texttt{G}, \texttt{cf}\}$ & X & $\{\texttt{G}, \texttt{cf}, \texttt{C}\}$ & X \\
        $\{\texttt{G}, \texttt{C}\}$ & \checkmark & $\{\texttt{G}, \texttt{cf}, \texttt{wp}\}$ & X \\
        $\{\texttt{G}, \texttt{wp}\}$ & X & $\{\texttt{G}, \texttt{cf}, \texttt{cm}\}$ & X \\
        $\{\texttt{G}, \texttt{cm}\}$ & \checkmark & $\{\texttt{G}, \texttt{C}, \texttt{wp}\}$ & X \\
        $\{\texttt{cf}, \texttt{C}\}$ & \checkmark & $\{\texttt{G}, \texttt{C}, \texttt{cm}\}$ & \checkmark \\
        $\{\texttt{cf}, \texttt{wp}\}$ & \checkmark & $\{\texttt{G}, \texttt{wp}, \texttt{cm}\}$ & X \\
        $\{\texttt{cf}, \texttt{cm}\}$ & X & $\{\texttt{cf}, \texttt{C}, \texttt{wp}\}$ & \checkmark \\
        $\{\texttt{C}, \texttt{wp}\}$ & \checkmark & $\{\texttt{cf}, \texttt{C}, \texttt{cm}\}$ & X \\
        $\{\texttt{C}, \texttt{cm}\}$ & \checkmark & $\{\texttt{cf}, \texttt{wp}, \texttt{cm}\}$ & X \\
        $\{\texttt{wp}, \texttt{cm}\}$ & \checkmark & $\{\texttt{C}, \texttt{wp}, \texttt{cm}\}$ & X \\
        \bottomrule
    \end{tabular}
    \caption{Recovery results on the Kepler problem for various subsets of missing axioms. 
    Shorthand: \texttt{G} = gravitational force $(d_1+d_2)^2F_g = m_1m_2$, 
    \texttt{cf} = centrifugal force $F_c = m_2d_2\omega^2$, 
    \texttt{C} = force balance $F_g = F_c$, 
    \texttt{wp} = frequency-period relation $\omega p = 1$, 
    \texttt{cm} = center of mass $m_1d_1 = m_2d_2$.
    }
    \label{tab:Kepler_multiple_missing_axiom_results}
\end{table}
\endgroup

\textbf{Results summary on other problems.}
Across twelve benchmark systems, we systematically removed axiom pairs and triples and attempted to recover explanations for $Q$ and have summarized our results in Table \ref{tab:tuple_ax_summary}. Success rates vary considerably by problem complexity and number of missing axioms. For 2-axiom recovery, success rates range from 10\% (Time Dilation, Light Damping) to 67\% (Kepler, Simple Harmonic Oscillator), with a median of 47\%. For 3-axiom recovery, performance degrades substantially: success rates drop to 0–40\%, with most systems achieving 10–25\% recovery. The best-performing systems are Kepler (70\% pairs, 20\% triples), Simple Harmonic Oscillator (67\% pairs, 20\% triples), and Relativistic Laws (81\% pairs, 69\% triples). The Hall Effect and Compton Scattering systems, despite their larger axiom sets (9–10 axioms), achieve moderate success (53\% and 38\% for pairs; 26\% and 21\% for triples, respectively). The Carrier-Resolved Photo-Hall system (62\% pairs, 42\% triples) demonstrates the method's scalability to complex systems with 7 axioms and 15 variables. Overall, recovery difficulty increases with the number of missing axioms and, to a lesser extent, with system size, though the relationship is non-monotonic.

Clearly having more information is helpful, and so the success rate is higher in cases where we were only missing two axioms rather than three. The highest success rate was for the Relativistic Laws (81\% pairs, 69\% triples), likely because this is the only system with not one but three consequences that we are drawing information from, as we will see in Section \ref{sec:mult:consequences}.

\begingroup
\arrayrulecolor{black}
\begin{table}[h]
\centering
\footnotesize
\renewcommand{\arraystretch}{1.3}
\begin{tabular}{l@{\hspace{8pt}}c@{\hspace{8pt}}c@{\hspace{8pt}}c@{\hspace{8pt}}c@{\hspace{8pt}}c@{\hspace{8pt}}c}
\toprule
\rowcolor{blue!20}
\textbf{Problem} & \textbf{Vars} & \textbf{Axioms} & \textbf{Pairs} & \textbf{Triples} & \textbf{2-Ax Success} & \textbf{3-Ax Success} \\
\midrule
Hagen--Poiseuille      & 9  & 4  & 6  & 4  & 3/6  & 1/4  \\
\rowcolor{blue!5}
Kepler                 & 8  & 5  & 10 & 10 & 7/10 & 2/10 \\
Time Dilation          & 8  & 5  & 10 & 10 & 1/10 & 1/10 \\
\rowcolor{blue!5}
Escape Velocity        & 9  & 5  & 10 & 10 & 8/10 & 4/10 \\
Light Damping          & 10 & 5  & 10 & 10 & 1/10 & 0/10 \\
\rowcolor{blue!5}
Neutrino Decay         & 8  & 5  & 10 & 10 & 3/10 & 1/10 \\
Simple Harmonic Osc.   & 10 & 6  & 15 & 20 & 10/15 & 4/20 \\
\rowcolor{blue!5}
Carrier-Resolved PH    & 15 & 7  & 21 & 35 & 13/21 & 15/35 \\
Relativistic Laws      & 12 & 7  & 21 & 35 & 17/21 & 24/35 \\
\rowcolor{blue!5}
Hall Effect            & 16 & 7  & 21 & 35 & 13/21 & 17/35 \\
Inelastic Relativistic Collision    & 11  & 9 & 36 & 84 & 12/36 & 15/84 \\
\rowcolor{blue!5}
Compton Scattering     & 15 & 10 & 45 & 120 & 17/45 & 25/120 \\
\bottomrule
\end{tabular}
\caption{Summary of 2- and 3-Axiom Tuple Recovery Performance by Problem. 
Pairs = $\binom{n}{2}$, Triples = $\binom{n}{3}$ where $n$ is the number of axioms.}
\label{tab:tuple_ax_summary}
\end{table}
\endgroup

\subsection{Multiple Consequences}\label{sec:mult:consequences}

As we will see in Section~\ref{recoverability}, having multiple consequences can help us recover decoupled axioms in the case where multiple axioms are missing. We demonstrate AI-Noether's ability to handle multiple related consequences simultaneously, exploring relativistic phenomena. Here, each consequence $Q_i$ is the sum 
\[
Q_i = \sum_{j=1}^{k+r} \alpha_{i,j}A_j
\]
where each $\alpha_{i,j}$ is unknown and $A_{k+1},\ldots,A_{k+r}$ are unknown as well. 

\subsection*{Example: Relativistic Laws}

\begin{figure}[t]
  \centering
  \includegraphics[
    width=\textwidth,
    height=0.4\textheight,
    keepaspectratio,
    trim={0cm 7cm 0cm 7.5cm}, clip
  ]{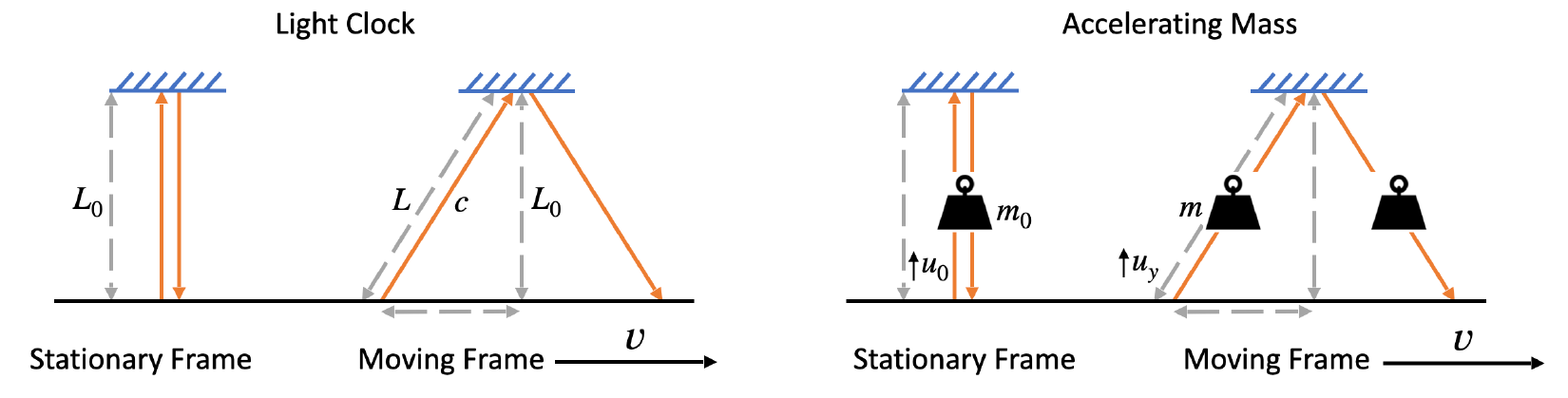}
  \captionsetup{font=small}
  \caption{\textbf{Illustration of the multiple experimental setups required to establish each of the relativistic laws - time dilation, length contraction, and relativistic mass.} The left two images illustrate a light clock experiment in a stationary and moving frame of reference required for time dilation and length contraction. The right two images illustrate an accelerated mass experiment in a stationary and moving frame of reference required for deriving relativistic mass. AI-Noether can extrapolate axioms from multiple consequences with more complex derivability dependency structures.}
  \label{fig:relativistic_laws}
\end{figure}

Consider Einstein's relativistic mass, length contraction, and time dilation formulae, which describe how mass increases and lengths contract for objects moving at relativistic speeds, illustrated in Figure \ref{fig:relativistic_laws}. These can be expressed in polynomial form as:
\begin{align*}
Q_1 &:= -c^2 f^2 + f_0^2(c^2 - v^2) = 0 & \text{Time dilation}\\
Q_2 &:= -m_0^2 c^2 + m^2 (c^2 - v^2) = 0 & \text{Relativistic mass}\\
Q_3 &:= L^2 c^2 - L_0^2(c^2 - v^2) = 0 & \text{Length contraction}
\end{align*}

\begin{figure}[t]
  \centering
  \includegraphics[width=\textwidth]{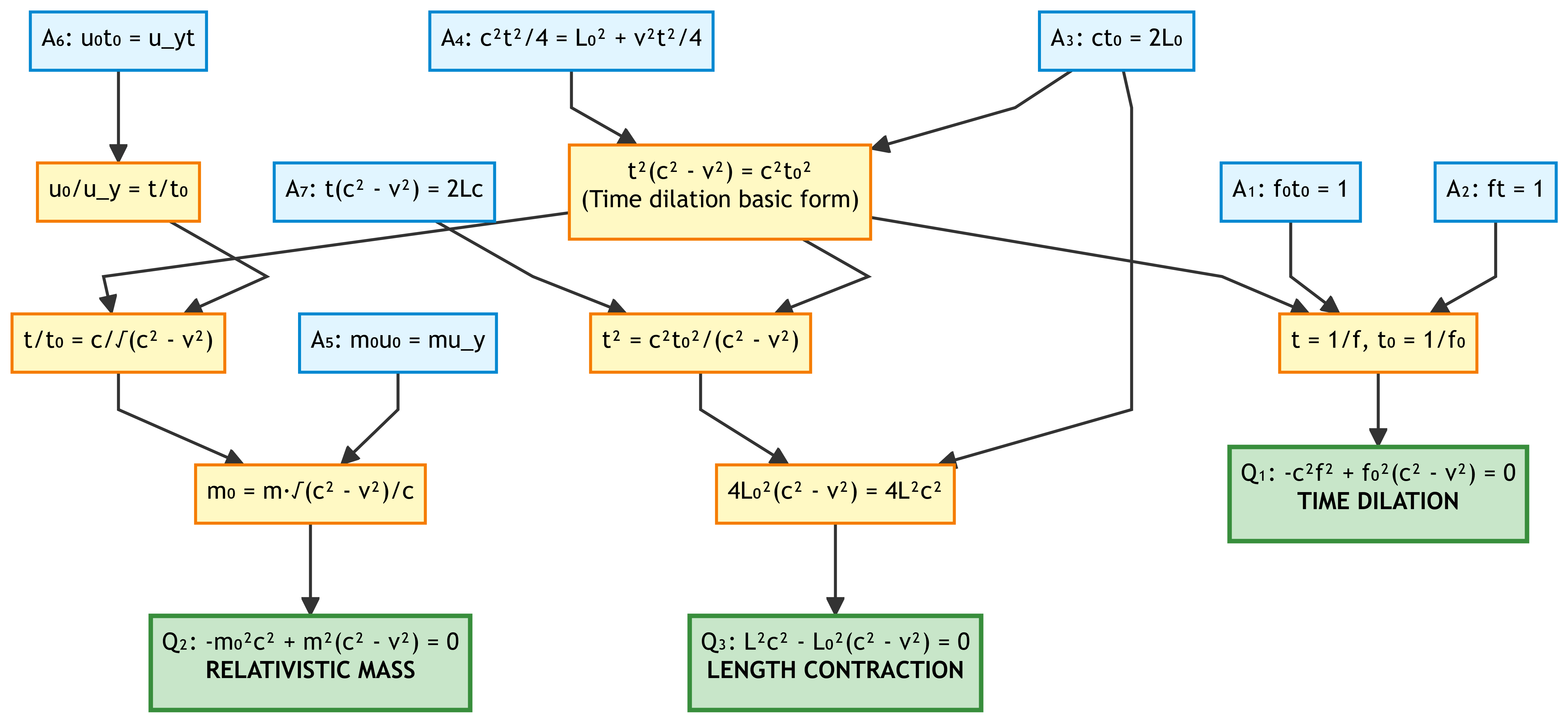}
  \caption{\textbf{Derivation structure for relativistic consequences.} The graph shows how axioms $A_1$--$A_7$ combine through intermediate results to derive the three consequences. Solid blue boxes represent axioms, yellow boxes show intermediate algebraic steps, and green boxes are the final consequences. The derivation of $Q_1$ (time dilation) uses axioms $A_1$--$A_4$; $Q_2$ (relativistic mass) additionally requires $A_5$--$A_6$; and $Q_3$ (length contraction) further needs $A_7$.}
  \label{fig:relativity_derivation}
\end{figure}

These consequences arise from a unified set of axioms describing light clock geometry in different reference frames. The axiom structure, illustrated in Figure~\ref{fig:relativity_derivation}, reveals how different subsets of axioms are required to derive each consequence: axioms $A_1$--$A_4$ form the foundation for time dilation ($Q_1$), with additional axioms $A_5$--$A_6$ needed for relativistic mass ($Q_2$), and additional axiom $A_7$ required for length contraction ($Q_3$).

\begin{align*}
A_1 &: f_0 t_0 - 1 & \text{Frequency-period (rest)}\\
A_2 &: f t - 1 & \text{Frequency-period (moving)}\\
A_3 &: ct_0 - 2L_0 & \text{Perpendicular light travel (rest)}\\
A_4 &: \frac{c^2t^2}{4} - L_0^2 - \frac{v^2t^2}{4} & \text{Light path geometry (moving)}\\
A_5 &: m_0 u_0 - m u_y & \text{Transverse momentum conservation}\\
A_6 &: u_0 t_0 - u_y t & \text{Transverse distance invariance}\\
A_7 &: t(c^2-v^2) - 2Lc & \text{Parallel light travel time}
\end{align*}

\textbf{Physical Interpretation.} 

Axioms $A_1$ and $A_2$ define the period-frequency relationship in the rest and moving frames, respectively. Axiom $A_3$ encodes that light travels distance $2L_0$ in time $t_0$ for a perpendicular round trip in the rest frame. Axiom $A_4$ captures the geometric constraint from a perpendicular light clock: a moving observer sees light travel along the hypotenuse of a right triangle, relating the rest-frame quantities to the moving-frame time and velocity. Axioms $A_5$ and $A_6$ ensure conservation of momentum and distance in the transverse (perpendicular) direction. Finally, axiom $A_7$ describes light traveling parallel to the direction of motion, accounting for the asymmetric forward and backward travel times.

The derivation structure proceeds as follows. From $A_3$ and $A_4$, we obtain $t^2(c^2 - v^2) = c^2t_0^2$, the basic time dilation relation. Substituting the frequency-period relations from $A_1$ and $A_2$ yields $Q_1$. For relativistic mass ($Q_2$), we additionally use $A_6$ to relate transverse velocities through the time dilation factor, then apply momentum conservation from $A_5$ to obtain the mass transformation. For length contraction ($Q_3$), axiom $A_7$ provides the parallel light travel constraint, which when combined with the time dilation relation and $A_3$ gives the length contraction formula.

\textbf{Abductive inference with multiple consequences.}
We test AI-Noether's ability to recover missing axioms in two scenarios: single-axiom removal and multi-axiom removal with multiple consequences.

\emph{Single-axiom case.} We saw in Table \ref{tab:single_ax_summary} that the combined relativistic laws were the only case where there were two axioms which we do not recover exactly. When removing individual axioms and attempting recovery using a single consequence, AI-Noether exhibits interesting behavior for axioms $A_4$ and $A_7$. Instead of recovering the original quadratic forms, the system returns factorized alternatives. For instance, when $A_4$ is removed, AI-Noether recovers the pair:
\[
c - 2f_0^2 L_0 t \quad \text{and} \quad c + 2f_0^2 L_0 t
\]
whose product $(c - 2f_0^2 L_0 t)(c + 2f_0^2 L_0 t) = c^2 - 4f_0^4 L_0^2 t^2$ is algebraically equivalent to the original constraint after substituting $f_0 t_0 = 1$ and $ct_0 = 2L_0$. A similar factorization occurs for $A_7$. This factorization phenomenon reflects the fundamental algebraic structure: the solution variety of the quadratic constraint decomposes into a union of two linear varieties (hyperplanes), each corresponding to one factor. 

This factorization phenomenon reflects the structure of the primary decomposition: when $A_4$ is removed, the variety $V(\{A_1, A_3, A_5, A_6, A_7, Q\})$ decomposes into components whose generators include these linear factors. These factors do not individually reproduce the physics of $A_4$, but they appear in the decomposition because they define subvarieties that, when intersected with the other axioms, collectively constrain the solution space in a way that allows $Q$ to be derived. This illustrates a subtle but important point: abductive inference may return algebraic expressions that are structurally different from the original axioms yet sufficient to derive the target consequence when combined with background knowledge. The system identifies what is algebraically necessary to close the gap between axioms and consequence, which may differ from the canonical physical formulation of the missing law.

\emph{Multi-axiom case with multiple consequences.} 

We now remove three axioms simultaneously: $A_2$ (required for all consequences), $A_3$ (required for $Q_1$ and $Q_3$), and $A_5$ (required for $Q_2$). Running AI-Noether with all three consequences $Q_1$, $Q_2$, $Q_3$ and the remaining axioms yields:

\resetvarieties
\begingroup
\setlength{\tabcolsep}{8pt}%
\renewcommand{\arraystretch}{1.28}%
\arrayrulecolor{white}%

\begin{tabularx}{\linewidth}{@{}Y@{}}
\sectionrow{Generators selected by reasoning module}
\rowcolor{gray!6}
\makebox[\linewidth]{\(\color{BrickRed!80!black}\mathbf{f t - 1}\)}\\[4pt]
\rowcolor{gray!12}
\makebox[\linewidth]{\(\color{BrickRed!80!black}\mathbf{ct_0 - 2L_0}\)}\\[4pt]
\rowcolor{gray!6}
\makebox[\linewidth]{\(\color{BrickRed!80!black}\mathbf{m_0 u_0 - m u_y}\)}\\[4pt]
\end{tabularx}
\endgroup

Notably, the system recovers \emph{all three} missing axioms exactly---$A_2$, $A_3$, and $A_5$---without the coupling artifacts observed in Section~\ref{sec:mult_missing_axioms}. This demonstrates a key advantage of having multiple consequences: the additional constraints allow the decomposition to isolate the individual missing axioms rather than returning their coupled combinations. As we formalize in Section~\ref{recoverability}, multiple consequences increase the algebraic constraints on the variety, forcing the primary decomposition to separate axioms that would otherwise appear only in entangled form. This multi-consequence strategy is essential for recovering decoupled axioms and represents a significant advantage over single-consequence abductive inference.

\subsection{Noisy Consequences}\label{sec:noise_experiments}

We now consider the effect of noise in the consequences. For each of the problems in this section, as well as with the remaining test problems, we ran AI-Hilbert \cite{corywright2024evolving} on synthetically generated datasets for $Q$ (respectively, $Q_i$ for relativistic laws) with standard Gaussian noise added to the data proportional to the means of each column with a noise level $\epsilon$. We therefore obtain perturbations of each consequence rather than the true consequences themselves. We illustrate this with the following two examples of Kepler's third law and relativistic laws continued from earlier.

\subsection*{Example with single axiom correction: Kepler, continued}

Taking the previous example of Kepler's third law, assume that the gravitational force axiom $F_g(d_1+d_2)^2-m_1m_2$ is missing. With a noise level of $\epsilon = 1\%$, the resulting consequence $Q$ that was obtained is given by:
\[
-1.01387623 m_2 d_2^3 - 2.02927363 m_2 d_1 d_2^2 - 1.01837465 m_2 d_1^2 d_2 + 1.0142876 m_1 m_2p^2=0
\]
This, however, does not come with a certificate of derivability, since the remaining axioms cannot derive this consequence. We assume as before that we have no data for the missing axiom, so we cannot use a traditional symbolic regression approach to determine the missing axiom. Algebraically computing a primary decomposition results in obtaining $Q$ as the only axiom candidate. Since algebraic decomposition uses exact computation, we can only handle exact derivability via algebraic projection. Therefore, we turn to a numerical irreducible decomposition that attempts to generate approximate points on the irreducible components, and then perform symbolic regression on these points. The data generated is defined over all the variables of the system, not just those in the consequence or any single axiom; therefore, there is a hyperparameter of which variables to fit to that is to be tuned. Of course, one could choose various hyperparameters to get a family of axiom candidates. Fitting the data generated in our system yields the following axiom candidate, which we then verify using our reasoning module.

\resetvarieties
\begingroup
\setlength{\tabcolsep}{8pt}%
\renewcommand{\arraystretch}{1.28}%
\arrayrulecolor{white}%

\begin{tabularx}{\linewidth}{@{}Y@{}}
\sectionrow{Axiom generated by Decomposition and Reasoning modules}
\rowcolor{gray!6}
\makebox[\linewidth]{\(\color{BrickRed!80!black}\mathbf{-0.502463F_g d_1^2 -1.012319F_g d_1d_2 -0.497537F_gd_2^2 + 0.497536 m_1 m_2}\)}\\[4pt]

\end{tabularx}
\endgroup

Notice that this indeed is a noisy perturbation of the missing axiom: $F_g(d_1^2 + 2 d_1 d_2 + d_2^2) - m_1 m_2$.

\subsection*{Example: Relativistic laws with noisy consequences}

We test the robustness of our approach by combining multiple challenges: noisy consequences, multiple missing axioms, and multiple consequences. Running AI-Hilbert~\cite{corywright2024evolving} to generate noisy versions of the three relativistic consequences from synthetically generated data with 1\% Gaussian noise (standard deviation proportional to the mean of each column) gives the following perturbations of the relativistic laws (rounded to four decimal places for brevity):

\begin{align*}
\tilde{Q}_1 &:= -1.0089c^2f^2 + 0.9986f_0^2c^2 - 0.9956f_0^2v^2 = 0 & \text{(Time dilation)}\\
\tilde{Q}_2 &:= -0.9876m_0^2c^2 + 1.0134m^2c^2 - 1.0089m^2v^2 = 0 & \text{(Relativistic mass)}\\
\tilde{Q}_3 &:= 0.9945L^2c^2 - 1.0123L_0^2c^2 + 1.0067L_0^2v^2 = 0 & \text{(Length contraction)}
\end{align*}

We removed three axioms that span different roles in the derivation structure (recall Figure~\ref{fig:relativity_derivation}): $A_2$ ($ft - 1$), required for time dilation; $A_5$ ($m_0u_0 - mu_y$), required for relativistic mass; and $A_4$ ($\frac{c^2t^2}{4} - L_0^2 - \frac{v^2t^2}{4}$), the geometric constraint from the perpendicular light clock. Without these, we again do not get any certificate of derivability of the laws.

This selection is also particularly interesting because, as we saw in Section~\ref{sec:mult:consequences}, axiom $A_4$ exhibits factorization behavior in the noiseless single-axiom case---the algebraic decomposition returns linear factors rather than the original quadratic form. This raises the question: can numerical methods recover what algebraic methods transform?

Using numerical irreducible decomposition (Section~\ref{sec:noise}) followed by symbolic regression on witness points sampled from each component, our system recovers noisy approximations of all three missing axioms:

\resetvarieties
\begingroup
\setlength{\tabcolsep}{8pt}%
\renewcommand{\arraystretch}{1.28}%
\arrayrulecolor{white}%

\begin{tabularx}{\linewidth}{@{}Y@{}}
\sectionrow{Axioms recovered via numerical decomposition and automated reasoning}
\rowcolor{gray!6}
\makebox[\linewidth]{\(\color{BrickRed!80!black}\mathbf{1.0000\, ft - 0.6599}\)}\\[4pt]
\rowcolor{gray!12}
\makebox[\linewidth]{\(\color{BrickRed!80!black}\mathbf{-0.9899\, m_0u_0 + 1.0000\, mu_y}\)}\\[4pt]
\rowcolor{gray!6}
\makebox[\linewidth]{\(\color{BrickRed!80!black}\mathbf{-0.2525\, c^2t^2 + 0.2451\, v^2t^2 + 1.0000\, L_0^2}\)}\\[4pt]
\end{tabularx}
\endgroup

The recovered axioms closely approximate the true missing relationships despite 1\% noise in the consequences. Most notably, axiom $A_4$ is recovered in its original quadratic form with coefficient errors of approximately 1--2\%, rather than the factorized linear expressions produced by algebraic decomposition in the noiseless case. This highlights a fundamental difference between algebraic and numerical abductive inference:

\begin{itemize}
\item \textbf{Algebraic decomposition} operates purely symbolically, decomposing varieties into unions of irreducible components. When $A_4$ is missing, the variety structure naturally factors into linear subvarieties, yielding generators like $c \pm 2f_0^2 L_0 t$. While algebraically sufficient for derivability, these do not match the canonical physical form.

\item \textbf{Numerical decomposition} samples witness points on each irreducible component, then performs symbolic regression to fit polynomials that approximately vanish on those points. Because we regress using the monomial support of the original missing axiom as a basis (here, $\{c^2t^2, v^2t^2, L_0^2, 1\}$), the method directly recovers a noisy version of $A_4$ rather than its factorization. The witness points constrain all monomials simultaneously, bypassing the algebraic factorization that occurs in the symbolic setting.
\end{itemize}

This demonstrates a practical advantage of the numerical approach. By operating on sampled data rather than symbolic manipulation, it can recover axioms in forms that more closely match their canonical physical formulations. The choice of monomial basis serves as an inductive bias that guides the regression toward physically meaningful parameterizations, while preserving the principled algebraic foundation of the decomposition step. The complete numerical workflow---witness set computation via homotopy continuation, robust symbolic regression with outlier removal, and coefficient normalization---is detailed in Section~\ref{sec:noise}.

\subsection*{Results summary on other problems}

\begin{figure}[t]
  \centering
  \includegraphics[width=1\textwidth, clip]{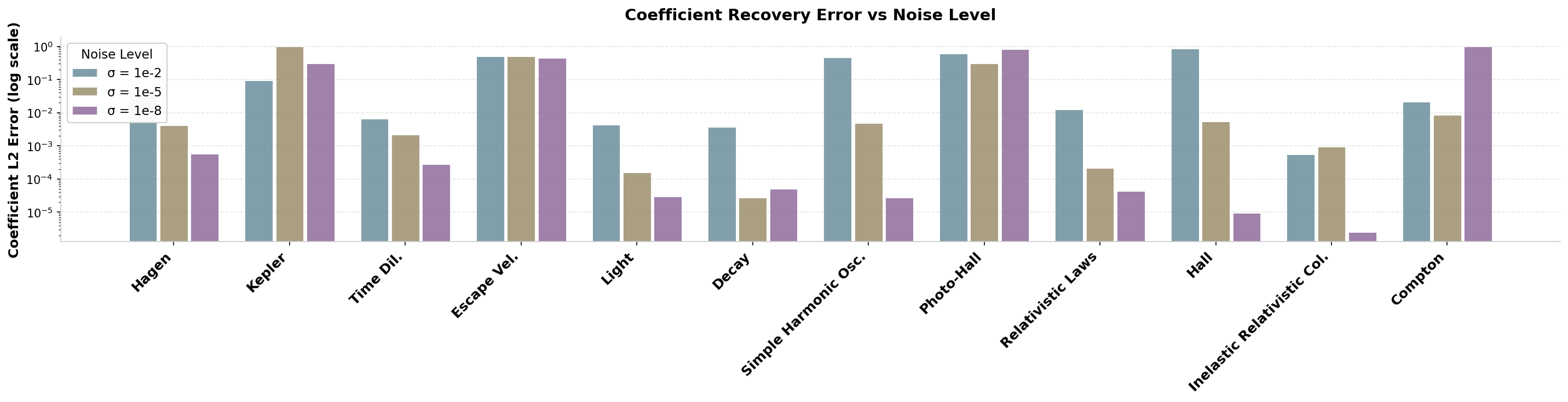} 
  \captionsetup{font=small} 
  \caption{\textbf{Results on running AI-Noether on various problems} with 1,2, and 3 axioms missing with noisy consequences obtained from running AI-Hilbert \cite{corywright2024evolving} with varying levels of noise in the data. For each problem, we plot the normalized coefficient distance \cite{corywright2024evolving, Cox+Others/1991/Ideals} and the noise level for successful runs.}
  \label{fig:noisy_results}
\end{figure}

We summarize our results in Figure \ref{fig:noisy_results}. For each problem, we apply the numerical irreducible decomposition (details in Section \ref{sec:noise}) to obtain witness points on each component of the variety defined by the remaining axioms and target. We then perform symbolic regression to recover the missing axiom by fitting a polynomial in the null space of the Vandermonde-like matrix constructed from these points. To improve robustness against numerical artifacts and outliers introduced by the homotopy continuation, we employ iterative outlier removal: points with the largest residuals (top 10\% by default) are discarded, and the regression is repeated until convergence. The recovered polynomial is normalized so that its largest coefficient has magnitude one, and we apply the same normalization to the ground-truth axiom. The coefficient distance reported is the $\ell_2$ norm between these normalized coefficient vectors, following \cite{corywright2024evolving}; since $p = 0$ and $-p = 0$ define the same constraint, we report the minimum of $\|c_{\text{fit}} - c_{\text{true}}\|$ and $\|c_{\text{fit}} + c_{\text{true}}\|$.

Failure cases in Figure \ref{fig:noisy_results} arise from two sources. (i) Timeouts during witness set computation or symbolic regression, which we cap at two hours by default; this time constraint primarily affected the Compton scattering results and is detailed in Appendix \ref{algorithm_details}. (ii) Cases where the reasoning module (Section \ref{sec:noise}) cannot verify that the recovered axiom, combined with the remaining axioms, implies the target consequence. We give a discussion on the criteria for when an axiom is recoverable in the noiseless and noisy case in Section \ref{recoverability}.

\subsection{Criteria for Recoverability of Axioms}\label{recoverability}

\subsection*{Recovering axioms without noise in the consequences}

The theorem below provides a sufficient criterion for recovering missing axioms. In order to state the technical details, we will use terminology from algebraic geometry. A detailed overview of all terminology relevant to the core method can be found in Appendix~\ref{sec:method_details}.

First, we require the notion of the height of an ideal. For a prime ideal $P$, we define its height, $\myht(P)$, as the length $n$ of the longest chain $P_0 \subsetneq P_1 \subsetneq \cdots \subsetneq P_n = P$ of prime ideals properly contained in $P$. For a general ideal, its height is the minimum of the heights of its associated primes, which is the same as the minimum of the heights of all primes that contain the ideal.

\begin{Theorem}{Common minimal associated prime}{thm:recoverability}
Given ideals $I \subset J \subsetneq \mathbb{R}[x_1,\ldots,x_n]$ such that $ \myht(I) = \myht(J)$, then $I$ and $J$ have a common minimal associated prime.
\end{Theorem}

\begin{Proof}{Theorem \ref{thm:thm:recoverability}}{}
Let $P$ be an associated prime for $J$ such that $\myht(P) = \myht(J)$. Then we have that $I \subset J \subset P$.
Since $\myht(I) = \myht(P)$, there is no prime $Q$ such that $I \subset Q \subsetneq P$.
Otherwise $\myht(I) \le \myht(Q) < \myht(P)$.
But then $P$ is a minimal prime containing $I$ and thus a minimal associated prime for $I$. 
\end{Proof}

Example: Let $J$ be the ideal of the full set of axioms $\langle A_1,\ldots,A_{k+1}\rangle$ (so assume only one axiom $A_{k+1}$ is missing) and let 
$Q$ be a consequent of $J$. Now treat $A_{k+1}$ as a missing axiom and let $I$ be the ideal generated by the known axioms and the consequence $\langle A_1,\ldots,A_k, Q\rangle$. If $I$ and $J$ have the same height, then we know that they have a common associated prime which must contain the missing axiom $A_{k+1}$. Under the usual situation where 
$Q$ and $A_{k+1}$ are not contained in any of the associated primes for $\langle A_1,\ldots,A_{k}\rangle$, then we have that $\myht \langle A_1,\ldots,A_k, Q\rangle =
\myht \langle A_1,\ldots,A_{k+1}\rangle
$, the theorem applies, and we can examine the associated primes for $I$ to recover the missing axiom $A_{k+1}$.

\textbf{Intuitive interpretation:} Geometrically, each axiom defines a hypersurface that reduces the dimension of the solution space by at most one. Removing an axiom removes a "cut," increasing the variety's dimension. Adding a hypothesis 
$Q$ introduces a new cut. The theorem states that if the dimension lost by removing axioms is exactly recovered by adding hypotheses, then the decomposition of $I$ will reveal a component containing the missing axiom. In other words, the hypotheses carry enough information to "reconstruct" the missing constraints.

Typically, if multiple axioms are missing, adding back a single phenomenon $Q$ will not be sufficient, so in general we will need to add back several consequent phenomena $\{ Q_1, \ldots Q_h \} $ in order to generate an ideal which is guaranteed to have a component that contains all the missing axioms.

Our missing axiom construction attempts to find axioms that are sufficient to derive $Q$, i.e., whether or not $Q$ is contained in the ideal generated by the axioms. If we want to decide whether or not a set of axioms implies $Q$, given the ideal $I$ generated by the axioms, we are asking whether $V(I) \subset V(Q)$.
To make this test effective, we need to construct the biggest ideal with the same set of zeros, this is $I(V(I))$. So the test whether a set of axioms $A$ implies $Q$ is whether $Q \in I(V\langle A\rangle)$. If our varieties are defined over the reals, then $I(V \langle A\rangle)$ can be computed as the real radical of the ideal $\langle A\rangle$.

\subsection*{Recovering axioms with noise in the consequences}

When the consequences are noisy, then it is no longer true that removing an axiom and adding a noisy consequent produces an ideal which is exactly contained in the original axiom ideal. It is in some sense ``nearby'' to an ideal which is so contained. To handle this, we replace exact ideal decomposition with numerical ideal decomposition \cite{bates2024numericalnonlinearalgebra}. This produces sample points on each component of the perturbed ideal. Assuming the perturbation is small and generic, we expect sample points for each numerical component to be approximate zeros of one of the components of the true ideal. \cite{cobian2024robustnumericalalgebraicgeometry}. When this is the case we can use symbolic regression on the sample points generated by numerical decomposition to generate approximations to the missing axiom.

\subsection{Comparison with State-of-the-Art}

We evaluate our abductive reasoning system against two representative baselines: 1) solver-based abduction using cvc5 with SyGuS grammars, and 2) LLM-based abduction using GPT-5 Pro constrained to output polynomial hypotheses. 
All methods are tested on a common benchmark of algebraic physics systems, where one or more axioms are removed, and the goal is to infer a missing axiom that will provide derivability of a target conjecture.
Each proposed axiom set is checked by the same symbolic verification pipeline in Macaulay2. Our experiments are both on clean and noisy data for each problem in our benchmark set, and use multiple levels of axiom removal. 

Overall, our system demonstrates substantially stronger performance than both baselines (see Table~\ref{tab:method_comparison}). It succeeds in 97\% of single-axiom cases where both baselines achieved 0\% (LLM) or timed out (cvc5), including cases where the missing axiom involves numerical constants or requires longer multi-step symbolic reasoning. Moreover, cvc5 generally fails to generate a polynomial within the allowed grammar, as it exceeds the runtime we set. GPT-5 Pro can generate syntactically reasonable expressions but doesn't yield axioms that satisfy the strict algebraic membership or literal-appearance criteria. Instead, our approach, in most cases, recovers minimal, correct, and theory-compatible axioms and remains robust to noise. 

More details about our experiments comparing AI-Noether with the state of the art, along with ablations and runtime statistics, can be found in Appendix~\ref{app:comparison2sota}.

\begin{table}[h!]
\arrayrulecolor{black}
\centering
\renewcommand{\arraystretch}{1.25}
\rowcolors{2}{warmrow}{warmbg}
\begin{tabularx}{\textwidth}{>{\bfseries}l*{6}{>{\centering\arraybackslash}X}}
\rowcolor{warmheader}
\toprule
Method & Single Axiom & Multi Axiom & Data-Free & Noise Robust & Var. Agnostic & Explainable \\ 
\midrule
Traditional SR & \xmark & \xmark & \xmark & \xmark & \xmark & \xmark \\
Integrated SR & \xmark & \xmark & \xmark & \cmark & \cmark & \cmark \\
LLM (GPT-5 Pro) & 0\% & 0\% & \cmark & \xmark & \cmark & \cmark \\
cvc5 & t-out & t-out & \cmark & \xmark & \xmark & \cmark \\
\midrule
AI-Noether & \textbf{97\%}~\cmark & \textbf{49\%}~\cmark & \cmark & \cmark & \cmark & \cmark \\
\bottomrule
\end{tabularx}

\vspace{4pt}
\caption{\textbf{Comparison of abductive and symbolic regression methods.}
Green ticks (\cmark) indicate capability support; red crosses (\xmark) indicate absence.  
“Data-Free” means the method operates without numeric data or prior parameter information,  
and “Var. Agnostic” indicates invariance to variable naming. ``t-out'' means that the method ran out of time without producing a solution.}
\label{tab:method_comparison}
\end{table}

\section{Final Remarks}

\subsection*{Code Availability}
An open-source implementation of AI-Noether as well as all the benchmark problems we tested on are available at \url{https://github.com/IBM/AI-Noether}. Algorithm implementation details are provided in Appendix~\ref{algorithm_details}. The repository includes documentation for reproducing all experiments presented in this paper.

\subsection*{Limitations of AI-Noether}

AI-Noether is a purely algebraic system for generating a minimal set of missing axioms from a polynomial hypothesis and a set of background theory expressible as polynomial equations. We employ the same techniques as in AI-Hilbert to deal with non-algebraic functions, derivatives and integrals. The first method is that we simplify integrals and derivatives as much as possible and then assign new variables to any remaining integrals, derivatives and non-algebraic functions. The second is that we treat derivatives symbolically, treating them as indeterminates in a polynomial. For trigonometric functions (as in the case of the simple harmonic oscillator, light damping, and compton scattering), we can consider trigonometric polynomials by looking at the corresponding axioms as polynomials in the respective trig functions. This becomes more challenging, however, when dealing with compositions of trig functions with other polynomial functions. It will be interesting therefore to extend AI-Noether to handle cases when these techniques are not sufficient to derive the consequences from the full set of axioms.

Moreover, AI-Noether can recover a missing axiom only if each variable appearing in that axiom also occurs in at least one of the available axioms or in the target hypothesis. If a missing axiom involves latent variables that are entirely absent from the observed theory and consequences, the current framework cannot identify or introduce them automatically. However, as seen in Section \ref{sec:mult_missing_axioms} with Kepler's third law and multiple missing axioms, AI-Noether in cases can recover information about variables involved in multiple missing axioms by discovering coupled versions of the missing axioms and variables. 

Finally, although effective on a wide range of benchmark systems, AI-Noether relies on algebraic decomposition, and symbolic verification whose computational cost can grow rapidly with the number of variables, axioms, and polynomial degrees, depending on the system.

\subsection*{Future Directions}
In this study, we have focused on developing what we believe is a first-of-its-kind abductive inference system and testing it on existing laws of nature. A natural next step is therefore to employ AI-Noether in scientific domains where existing theories are incomplete or internally inconsistent, with the aim of supporting genuinely novel discoveries and of discovering new axioms as substantive scientific contributions that extend or revise existing theories. We also aim to extend AI-Noether beyond equations over a polynomial basis to handle differential operators and inequality constraints, enabling abductive inference across a broader range of scientific domains. 

\subsection*{Conclusion}

This work introduces AI-Noether, an automated abductive reasoning system that connects AI-generated scientific hypotheses with established theoretical knowledge. Given a set of background axioms and a hypothesis not supported by existing theory, AI-Noether discovers a minimal set of additional axioms that reconcile the hypothesis with the theory, provided both are expressible as polynomials. By combining algebraic-geometry tools with symbolic verification, the system provides candidate explanations and formal guarantees for deriving target hypotheses from the augmented axioms.

AI-Noether not only demonstrates effectiveness in establishing a breadth of known scientific laws but also positions axiomatic discovery as a fundamental form of scientific inquiry. AI-Noether supports the discovery of novel hypotheses (e.g., those generated by data-driven or theory-based methods) by determining whether and how to integrate them into coherent theoretical frameworks. In this sense, the system advances the scientific method by explicitly addressing the problem of identifying and explaining the underlying principles governing the environment in which a phenomenon occurs.

We demonstrated the effectiveness of AI-Noether across a broad set of physical systems, including classical mechanics, electromagnetism, and semiconductor physics. We showed that it robustly recovers missing axioms under both noiseless and noisy conditions. Comparisons against solver-based methods and language model baselines highlight our approach's advantages in correctness and formal verifiability. More broadly, applying AI-Noether to domains where current theories are incomplete or internally inconsistent points toward a shift in the scientific method itself: one in which abductive inference not only explains existing discoveries better, but also uncovers new principles that reveal the fundamental structure of the world around us.

\bibliographystyle{plain} 
\bibliography{refs} 

\newpage
\appendix

\section{Extended Literature Review}\label{append.litreview}

Abductive reasoning is a field of inference that generates plausible explanations from incomplete observations \citep{peirce1934collected, paul1993approaches}. Unlike deductive reasoning, which obtains a conclusion from a set of premises, abductive reasoning starts with an outcome and ends with a set of plausible explanations, e.g.,
\begin{itemize}
    \item The fact $C$ is observed.
    \item If $A$ were true then $C$ would be true.
    \item There is reason to believe that $A$ is true.
\end{itemize}
As observed by Peirce \cite{peirce1934collected}, abductive reasoning is arguably ``the only logical operation which introduces any new ideas''. Moreover, it is often applied by humans in everyday situations, such as reading between the lines \citep{norvig1987inference} and counterfactual reasoning \citep{pearl2002reasoning}. Doctors use abductive reasoning to propose possible diseases that cause observed symptoms. It has been argued that automating inference may be necessary for understanding and automating consciousness \cite{bengio2017consciousness}.
However, abductive reasoning has not yet been incorporated into state-of-the-art AI systems, such as AI-Feynman, AI-Descartes, or AI-Hilbert \cite{AI-Feynman, descartes, corywright2024evolving}. Thus, automating abductive inference is an important component of automating the scientific method.

\paragraph{Automated Abductive Inference in Logic and Classical AI.} 
Abduction, firstly introduced by Peirce as the inferential step that proposes an explanatory hypothesis for a surprising observation in the logic of scientific inquiry~\cite{abduction-stanford}, was then later on, in AI, sharpened into a logic-based hypothesis synthesis problem~\cite{marquis}. 
Given an incomplete background theory $T$ and observations (target claim) $O$, find hypotheses $H$ (additional assumptions from a designated space of abducibles, e.g., defined by a grammar) that render the claim derivable, meaning that $T \cup H \models O$ and $T \cup H$ is consistent (typically subject to minimality criteria) \cite{eiter1995complexity}.
This was first introduced for propositional logic and then extended to more expressive forms such as first-order~\cite{marquis}. 
Abductive logic programming (ALP) then further operationalized these ideas by combining abducibles with integrity constraints and proof procedures~\cite{kakas1992alp}, and more recent work extended it to even richer constraint families (e.g., including probabilistic integrity constraints~\cite{bellodi2021nonground}).

Most relevant to our work is the case in which hypotheses are \emph{equations} (or more generally, algebraic constraints): in this setting, abduction can be realized via consequence generation in equational logic, where explanations are returned as equalities/disequalities that complete the theory enough to derive the target~\cite{echenim2017prime}. More broadly, implicate-generation modulo background theories can be used to generate missing conditions (including equalities and arithmetic constraints) whenever a suitable decision procedure is available~\cite{echenim2018generic,echenim2013approach}.
However, these methods either support only limited arithmetic expressiveness (usually including only basic arithmetic operators) or are computationally challenging as the hypothesis space grows.

More recently, abduction has become part of automated reasoners/solvers (see paragraph below for more details): SMT/SyGuS-based methods synthesize hypotheses directly in background theories with equality or linear arithmetic \cite{reynolds2020scalable}. Automated solvers, like cvc5 expose abduction as a reusable primitive \cite{barbosa2022cvc5}. These can be also incorporated in verification-oriented systems where abduction fills in preconditions and specifications so that program proofs can find a proof compositionally \cite{spies2024quiver,barbosa2023interactive}.

Finally, beyond logic, several papers have investigated abductive inference in the context of scientific discovery: Swanson and Smalheiser’s interactive system ARROWSMITH~\cite{swanson} for identifying complementary literatures in other branches of science exploits abductive hypothesis formation as constructing bridging concepts across disjoint research areas; 
and more recently Langley~\cite{langley24} argued for integrated discovery systems that treat abduction as one component within an end-to-end pipeline that combines multiple reasoning and learning mechanisms.

We follow this line of work in interpreting abduction as the process of selecting or synthesizing hypotheses that complete an incomplete theory in order to explain a target hypothesis.

\paragraph{Automated Abductive Inference for Machine Learning.} The recent surge in popularity of black-box models, including large language models, has led to an interest in explaining the predictions of these models. LIME \cite{ribeiro2016} and similar methods such as SHAP \cite{lundberg2017unified} provide (local) interpretable explanations that describe the local behavior of any ML model using a weighted linear combination (often sparse) of input features; see \cite{rudin2019stop, lipton2018mythos} for reviews of interpretable and explainable machine learning. However, existing works on interpretability and explainability generally focus on computing explanations for black-box models, rather than generating theories that could be used for scientific discovery. In particular, techniques like LIME or SHAP scores cannot, to our knowledge, be leveraged to establish missing axioms in a scientific discovery context.

More recently, a new line of work has explored abductive inference for machine learning \cite{dai2019bridging, ignatiev2019abduction, zhou2019abductive, huang2021fast, yang2024analysis}; see also \citep{wan2024towards} for a recent review of this line of work. Initiated by \cite{dai2019bridging, ignatiev2019abduction}, the key idea is to embed purely logical reasoning models within the machine learning process to take advantage of the benefits of both machine learning and logical reasoning. For instance, in \cite{ignatiev2019abduction}, the authors use abductive inference to generate explanations of ML models given input background theories and a set of hypothetical explanations. Moreover, \cite{dai2019bridging} combines machine learning techniques with a first-order logic model to obtain an output that depends on both models. However, these works do not apply abductive inference to scientific discovery. 

\paragraph{SMT Solvers and Abductive Reasoning in Automated Theorem Proving.}
Recent work has explored the use of Satisfiability Modulo Theories (SMT) solvers for abductive reasoning, particularly in the context of automated theorem proving and program verification. SMT solvers generalize Boolean SAT by incorporating theories over specific domains (e.g., arithmetic, arrays, bit-vectors), enabling reasoning about richer constraints \citep{barrett2009smt, deharbe2009smt}. 

Barbosa et al.~\citep{barbosa2023interactive} developed an SMT-based tactic in the Coq proof assistant that uses abductive reasoning to identify missing axioms needed to complete a proof. When the cvc5 SMT solver fails to prove a goal, their \texttt{abduce} tactic generates facts that, when added to the background theory, allow the proof to succeed. This work shares conceptual similarities with AI-Noether, as both identify missing axioms to bridge a derivability gap, though their approach operates within formal proof systems rather than scientific discovery contexts. Reynolds et al.~\citep{reynolds2020scalable} introduced scalable algorithms for abduction via syntax-guided synthesis (SyGuS), demonstrating that SMT solvers can efficiently generate formulas consistent with axioms that entail a given goal. Their work establishes theoretical foundations for abductive synthesis that inform our comparison with cvc5 in Section~\ref{app:comparison2sota}.

Beyond theorem proving, abductive reasoning has been applied to scientific hypothesis generation through Abductive Logic Programming (ALP). For example, Ray~\citep{ray2007automated} demonstrated how ALP methods outperform theorem-proving approaches. While ALP focuses on logic-based hypothesis synthesis, AI-Noether operates in the polynomial algebraic setting, leveraging computational algebraic geometry rather than first-order logic, and works in the robust cases of noisy machine-generated hypotheses. 

\paragraph{Large Language Models for Scientific Discovery and Equation Discovery.}
The emergence of large language models (LLMs) has opened new avenues for automating scientific discovery. Recent work has explored using LLMs' embedded scientific knowledge and code generation capabilities to discover governing equations from data, a task traditionally addressed by symbolic regression.

Shojaee et al.~\citep{shojaee2024llmsr} introduced LLM-SR, which combines LLMs with evolutionary search to discover scientific equations by treating them as program skeletons with placeholder parameters. Their approach leverages domain-specific priors from LLM training to navigate the equation search space more efficiently than classical symbolic regression methods. Follow-up work by DrSR~\citep{drsr2025} enhances this framework with dual reasoning that combines data-driven insights with reflective learning, analyzing structural relationships like monotonicity and correlation to guide equation generation. These LLM-based symbolic regression approaches are complementary to AI-Noether: while they discover equations \emph{from data}, AI-Noether discovers equations needed to \emph{explain} already-formulated hypotheses without requiring data for the missing axioms.

Beyond symbolic discovery, LLMs have enabled autonomous experimental systems. Boiko et al.~\citep{boiko2023autonomous} demonstrated fully autonomous chemical research where LLMs plan multi-step syntheses, control laboratory robotics, and analyze results—closing the loop between computational hypothesis generation and physical experimentation. This represents a step toward Kitano's "Nobel Turing Challenge"~\citep{kitano2021nobel} of building AI systems capable of Nobel-quality autonomous discovery.

More broadly, LLMs have been applied to automated theorem proving and mathematical discovery. Romera-Paredes et al.~\citep{romera2024mathematical} introduced FunSearch, an evolutionary procedure pairing pretrained LLMs with systematic evaluation to solve open problems in extremal combinatorics, demonstrating that LLMs can push beyond existing human knowledge when combined with search algorithms. Lu et al.~\citep{lu2024aiscientist} proposed The AI Scientist, a system for fully automated open-ended scientific discovery that generates research ideas, runs experiments, and writes papers. In formal mathematics, recent work on automated theorem proving using LLMs~\citep{polu2020generative, yang2023leandojo} has shown promise in completing proof steps, though performance still lags human experts on research-level problems. These systems focus on \emph{generating} new knowledge or proofs, whereas AI-Noether performs \emph{abductive inference} to identify which axioms are missing from an incomplete theory. The axioms we are inferring are, by assumption, not derivable from the other known axioms and consequences, so traditional forward inference approaches would not work.

\paragraph{Neuro-Symbolic Methods and Physics-Informed Discovery.}
Recent work has explored hybrid approaches that combine neural networks with symbolic reasoning for scientific discovery. Petersen et al.~\citep{petersen2021deep} introduced deep symbolic regression using reinforcement learning with risk-seeking policies to discover mathematical expressions, demonstrating improved exploration of the symbolic search space. Valipour et al.~\citep{valipour2021symbolicgpt} developed SymbolicGPT, a transformer-based model for symbolic regression that learns from large corpora of synthetic equations. These neural-symbolic approaches excel at pattern recognition and generalization across equation families, but typically require training data for the target expressions.

Physics-informed approaches incorporate domain knowledge as inductive biases. Gao et al.~\citep{gao2023physicsinformed} showed that embedding physical constraints (e.g., conservation laws, symmetries) into neural network architectures improves equation discovery accuracy and generalization. Similarly, SINDy \citep{brunton2016discovering} leverages sparsity constraints to identify governing equations of dynamical systems and Hamzi and Owhadi \cite{HAMZI2021132817} similarly demonstrate the use of kernel flows in learning dynamical systems. While these methods successfully integrate prior knowledge, they focus on \emph{parameter fitting} within known functional forms rather than \emph{structural discovery} of missing axioms. AI-Noether addresses the complementary problem where the functional form itself is unknown and must be inferred from algebraic constraints.

\begin{figure}[h]
  \centering
  \includegraphics[width=1.0\textwidth]{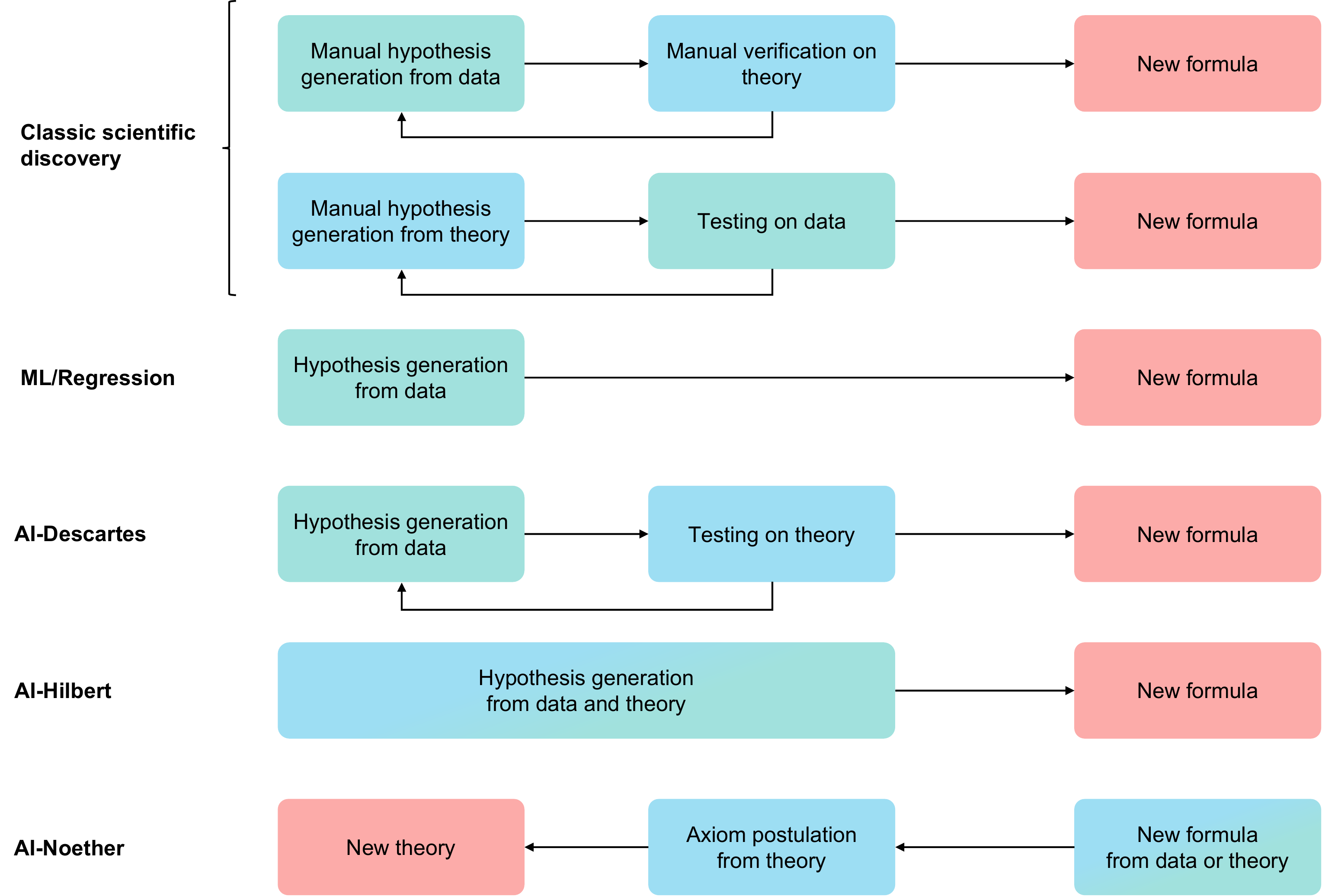} 
  \captionsetup{font=small} 
\caption{{\bf Comparison of AI-Noether with other scientific discovery models.} 
We compare classical scientific discovery, machine-learning–based regression, and recent integrated systems (AI-Descartes and AI-Hilbert) with AI-Noether. While most approaches focus on generating new hypotheses or formulas from data and/or existing theory, AI-Noether targets hypotheses that are not supported by the current theory and performs abductive inference to identify and justify the missing axioms required to reconcile them with known knowledge.}
  \label{fig:comparison}
\end{figure}

\paragraph{AI-Noether vs SOTA scientific discovery models.}
Figure~\ref{fig:comparison} compares \verb|AI-Noether| with four other scientific discovery models in terms of their functionality in discovering theory from data and existing theory. The traditional discovery process starts with known theory and data and seeks to discover a new formula that explains the data in a way that is consistent with the theory. The traditional, first-principles-based scientific discovery, broadly speaking, operates on two paradigms: either generate a hypothesis by studying patterns in known data and verify the hypothesis by checking derivability or consistency with theory or derive possible hypotheses from first-principles and test the derived hypotheses on data. This approach is manual and time-consuming. ML/Regression approaches help automate the hypothesis generation step by allowing for automated hypothesis generation using data. Verification here, however, is still done manually, and is often the bottleneck. AI-Descartes and AI-Hilbert solve this problem by automating the verification step by either automating the testing on theory using theorem provers (AI-Descartes) or combining fit-to-theory in the search process (AI-Hilbert). AI-Noether targets a new problem of automation in the scientific process: in the event that theory is insufficient for deriving a machine-generated hypothesis, instead of changing the hypothesis-generation mechanism, AI-Noether proposes corrections / completions to the theory in order to discover a new, minimal theory required to validate the hypothesis.

\section{A Walk-through  Example}

\begin{example} Consider an axiom system over five real quantities, indicated by $x,y,z, a, b$:
\[ x^2 - ay^2 = 0,\]
\[ by - z = 0.\]
Further, suppose that $a,b$ are physical constants in the system, while $x,y,z$ are quantities that may vary. Let $p_1$ be the polynomial $x^2 - ay^2$ and let $p_2 = by - z$, and the axioms be $p_1 = 0$ and $p_2 = 0$.
Further, let
\[ q = b^2x^2 - az^2.\]
The equation $q = 0$ then follows from the axioms as
\begin{equation}\label{eq-ex} q = b^2x^2 - az^2 = b^2(x^2 - ay^2) + a(by + z) (by-z) = b^2 p_1 + a(by + z) p_2.
\end{equation}
\end{example}

In the example above, we have expressed $q$ as $\alpha p_1 + \beta p_2$ where $\alpha = b^2$ and $\beta = a(by+z)$ are polynomials defined over $x,y,z,a,b$. We say that $q$ is an {\em algebraic combination} of the polynomials $p_1$ and $p_2$ and say that $q = 0$ is {\em derivable} from $p_1 = 0$ and $p_2 = 0$. As $q = \alpha p_1 + \beta p_2$, it follows that $q = 0$ for all  values of $x,y,z,a,b$ such that $p_1 = 0$ and $p_2 = 0$. We therefore say that the axioms in the example imply or explain $q = 0$.

\begin{table}[h]\caption{Six minimal axiom systems; each column corresponds to one such system, consisting of two equations that explain $q = 0$.}\label{tab1}
\centering
\begin{tabular}{l|l|l|l|l|l}
\midrule
$b = 0$ & $b = 0$ & $b = 0$ & $x^2 - ay^2 = 0$ & $x^2 - ay^2 = 0$ & $x^2 - ay^2 = 0$ \\
$a = 0$ & $by + z = 0$ & $by - z = 0$ & $a = 0$ & $by + z = 0$ & $by - z = 0$ \\
\midrule
\end{tabular}
\end{table}

Notice that there are many alternative theories from which the equation $q = 0$ follows as a consequence. For example, looking at the equation (\ref{eq-ex}) and letting $p_3 = by+z$, it is clear that $q = 0$ follows from the axioms $p_1 = 0$ and $p_3 = 0$ as $q = \alpha p_1 + ap_2p_3$. In a similar manner, we can trivially create the six {\em minimal} theories in Table~\ref{tab1} where the first axiom is formed by taking a polynomial that is a factor of $\alpha p_1$ and setting it to 0, and the second axiom is similarly composed from a factor of $\beta p_2$; $q = 0$ is a consequence of each theory.

Each of the 6 axiom systems above is minimal in the sense that we cannot remove an axiom from it and still have $q = 0$ as a consequence. Depending on the context, some axiom systems may be more interesting than others. For example, if $a$ represents the mass of a pendulum, $a = 0$ is likely not an interesting condition. On the other hand, if $a$ represents temperature in Celsius, $a = 0$ may represent a meaningful condition. 

The axiom system $p_1 = 0, p_2p_3 = 0$ is also a valid theory explaining $q = 0$, but we call it {\em reducible} in the sense that we can trivially get a simpler axiom system if we replace $p_2p_3$ by either of its factors. Thus, if our only available axioms were $p_1 = 0$, the ``residual'' of the set of axioms $\{p_2p_3 = 0\}$ is not as small as possible.
There are other ways of building an infinitude of axiom systems such that $q = 0$ is a consequence. Starting from the axiom system $p_1 = 0, p_2 = 0$, it is easy to see that $p_1 = 0,  rp_1 + sp_2 = 0$, where $r,s$ are nonzero real numbers, are valid theories that explain $q = 0$. Again, starting with $p_1 = 0$, the axiom set $\{rp_1 + sp_2 = 0\}$ does not have the least residual.

 Now, assume we have experimental data for a phenomenon consisting of a number of data points specifying values of $x, z, a,b$ (with $a$ and $b$ unchanging in the data), that the data satisfy the equation $b^2x^2 - az^2 = 0$, and that we are able to numerically obtain this equation from the data. Further, suppose that we have a partial theoretical explanation of the phenomenon, where the partial theory assumes the presence of another variable $y$ and the equation $x^2 - ay^2 = 0$, i.e., $p_1 = 0$. In this situation, we consider $y$ as an unmeasured quantity, and $x$, $z$, $a$, and $b$ as measured quantities. Given that the data points at hand do not have any values for $y$, it is not possible to infer the relationship $by - z$ directly from the data.
 
 Consider the equation $0 = b^2p_1 = b^2x^2 - b^2ay^2$ that follows from $p_1 = 0$. Subtracting the derived equation $b^2 x^2 - az^2 = 0$ from it, we obtain the relationship $a(z^2 - b^2y^2) = 0$ as a consequence of $p_1 = 0$ and $q = 0$. This implies that either $a = 0$ or $by - z = 0$ or $by+z = 0$. 
Each of the equations (or products of pairs of equations) in the previous line can be appended to $p_1 = 0$ to get a valid theory (in the absence of other information) explaining $q = 0$. 

Using the notation above, ``derivable from'' implies ``can be explained by''. We will later discuss conditions under which the converse is true (see Section \ref{recoverability}).

\section{Methodology details}\label{sec:method_details}

In this section, we develop the details of \textit{AI-Noether} and its \textbf{Encode-Decompose-Reason} modules. We first develop the algebraic theory of the method and then extend it to numerical algorithms that enable us to handle noise. For further reading on these topics, one can refer to the following textbooks and resources on computational algebraic geometry \cite{Cox+Others/1991/Ideals, smith2013invitation, atiyahmacdonald1969, eisenbud1995commutative, GIANNI1988, Eisenbud1992, hartshorne}. This section will include all the algebraic geometry framework and, as such, will be denser than the rest of the paper.

\subsection{Preliminaries - Algebraic Background Without Noise}\label{sec:prelim}

We define notions specific to real algebraic geometry; see \cite{Cox+Others/1991/Ideals} for further details. Let $\mathbb{R}$ denote the set of real numbers and let $R = \mathbb{R}[x_1,\ldots,x_n]$ be the ring of polynomials defined over the $n$ variables $x_1, \ldots, x_n$ with real coefficients. Axioms and hypotheses in our model will be equations of the form $p(x) = 0$ where $p \in R$. We abuse notation and identify an axiom by its defining polynomial. We will focus on real-valued solutions of axiom systems; see the appendix for more details. Finally, we will focus the exposition on the case with one axiom correction and one consequence for ease of explanation. This will generalize well to the multiple missing axiom and consequence case and we will make a note of these generalizations for each module. 

\textbf{Encode.} An {\em ideal} over $R$ is a subset $I$ of $R$ that is closed under addition and also multiplication by elements of $R$: if $f, g \in I$, then $f + g \in I$ and $\alpha f \in I$ for any $\alpha \in R$. Given polynomials $A_1,\ldots,A_k \in R$, we denote the set of all algebraic combinations of $A_1, \ldots, A_k$ by
$$\langle A_1, \ldots, A_k \rangle = \{f \in R: f = \sum_{i=1}^k \alpha_i A_i \text{ for some }\alpha_i \in R\}.$$ 
 This set is called an ideal over $R$ with \textit{generators} $A_1, \ldots, A_k$. 
Let $\mathcal{A}$ stand for $\{A_1, \ldots, A_k\}$, and let $I(\mathcal{A})$ stand for the ideal $\langle A_1, \ldots, A_k\rangle$.
 Our earlier notion of ``derivability'' corresponds to ideal membership:  $Q = 0$ is derivable from $\mathcal{A}$ if $Q$ is an algebraic combination of $A_1, \ldots, A_k$, i.e., if $Q \in I(\mathcal{A})$. 
Given an ideal $I$, we define the \textit{variety} of $I$, denoted $V(I)$, as the set
$$V(I) = \{\mathbf{x}\in \mathbb{R}^n : f(\mathbf{x}) = 0 \text{ for each } f \in I\}.$$
A standard fact from algebraic geometry is that the set above is the same as the solution set to any finite collection of polynomials that generate the same ideal.
Encoding equations with ideals and varieties has a key advantage in that it allows us to study logical implication and consistency questions geometrically: if $Q = 0$ can be derived from $A_1 = 0,\ldots,A_k = 0$, then $V(A_1,\ldots,A_k,Q) = V(A_1,\ldots,A_k)$. If adding $Q$ to $\mathcal{A}$ introduces extra structure into the resulting variety,
then $Q$ is not derivable from $\mathcal{A}$. As such, this holds for any $Q$, and so in the case where we have multiple consequences $Q_i$, this holds for each $Q_i$. In that case, we define $V(A_1,\ldots,A_k,Q_1,\ldots,Q_s)$ as our variety.

\textbf{Decompose.} 
Assume $Q$ is not implied by $\mathcal{A}$. Then $V(\mathcal{A},Q) \neq V(\mathcal{A})$. We will detail this in the case where a single axiom correction is needed. Suppose $Q$ can be derived {\em nontrivially} from an axiom $A_{k+1}$ along with $\mathcal{A}$, i.e., $Q = \alpha_1 A_1 + \cdots \alpha_{k+1} A_{k+1}$ for some polynomials $\alpha_i$ without $\alpha_{k+1}=1, A_{k+1}=Q$. In this case $V(A_1,\ldots,A_k,Q) = V(A_1,\ldots,A_k, \alpha_{k+1} A_{k+1})$ and since $\alpha_{k+1} \neq 1$, we know that intersecting $V(Q)$ with $V(\mathcal{A})$ is the same as intersecting $V(\alpha_{k+1}A_{k+1})$ with $V(\mathcal{A})$. Since  $V(\alpha_{k+1}A_{k+1})$ is reducible, we are introducing new reducibility into $V(\mathcal{A})$ that is a factor of the residual $\alpha_{k+1}A_{k+1} = Q-(\sum_{i=1}^k \alpha_i A_i)$ (and if we had multiple consequences, then we would have $V(A_1,\ldots,A_k,Q_1,\ldots,Q_s) = V(A_1,\ldots,A_k,\alpha_{k+1}A_{k+1},\ldots,\alpha_{l-k+1}A_{l-k+1})$). We therefore can take our intersected variety $V(\mathcal{A},Q)$ and study its irreducible components. We next define the mechanisms required for this process. 

A \textit{prime ideal} is an ideal $P \subset R$ such that if $fg \in P$ with $f,g \in R$, then $f\in P$ or $g\in P$. 

For a prime ideal $P$, $V(P)$ is irreducible, i.e., it cannot be represented as the union of two proper subsets that are varieties. The {\em radical} of an ideal $I$ is denoted by $\sqrt{I}$ and is defined as:
\[ \sqrt{I} = \{ f \in R : f^m \in I \text{ for some positive integer } m\}. \]

The following key theorem allows us to break up a variety into irreducible components:

\begin{Theorem}{Primary Decomposition}{thm:Primary_decomposition}
Any ideal $I$ can be decomposed into an intersection of ideals
$$I = I_1\cap \ldots \cap I_r$$
such that 
\begin{enumerate}
    \item There are no redundancies among the $I_i$, i.e., for each $j$ we have $\cap_{i\neq j}I_i \not\subset I_j$.
    \item For each $i$, $\sqrt{I_i}$ is a prime ideal (called an \textit{associated prime} of $I$).
\end{enumerate}
\end{Theorem}
Let $P_i = \sqrt{I_i}$.
Using the fact that for any ideals $I,J$, we have $V(I\cap J) = V(I)\cup V(J)$ and $V(I) = V(\sqrt{I})$ \cite{Cox+Others/1991/Ideals}, we get from the primary decomposition that a variety can be decomposed into a union of varieties corresponding to prime ideals: $V(I) = \bigcup_{i=1}^r V(P_i)$.

Decomposition in this setting serves a role analogous in spirit to Principal Component Analysis (PCA) in linear algebra, identifying latent, independent substructures that explain variation in the system. Whereas PCA decomposes a dataset into orthogonal directions of variance, primary decomposition identifies algebraically independent pieces of a solution space introduced by inconsistency or incompleteness.

\textbf{Reason.} To test derivability, we rely on an algebraic technique for eliminating variables from polynomial systems. A \emph{reduced Gr\"obner basis} is a canonical generating set for an ideal that can be computed algorithmically. Given a variable ordering $x_1 < \cdots < x_n$ (e.g., lexicographic), the reduced Gr\"obner basis $\{G_1, \ldots, G_m\}$ of $I = \langle A_1, \ldots, A_k \rangle$ satisfies a key elimination property:

\begin{Theorem}{Elimination Theorem}{Elimination_Theorem}\label{Elimination Theorem} 
For any $1 \leq d \leq n$, the polynomials in the Gr\"obner basis $\{G_1,\ldots,G_m\}$ involving only $x_1,\ldots,x_d$ generate the same ideal as all polynomials in $I$ involving only those variables:
\[
\langle \{ G_1,\ldots,G_m\} \cap \mathbb{R}[x_1,\ldots,x_d]\rangle = \langle A_1,\ldots,A_k\rangle \cap \mathbb{R}[x_1,\ldots,x_d].
\]
\end{Theorem}

So if we want to eliminate the variables $x_{d+1},\ldots,x_n$ from the system, we compute a set of generators for the elimination ideal $I \cap \mathbb{R}[x_1,\ldots,x_d]$ by finding the reduced Gr\"obner basis and extracting polynomials involving only $x_1, \ldots, x_d$. Geometrically, this corresponds to projecting the variety $V(I) \subset \mathbb{R}^n$ onto the subspace of the first $d$ coordinates. The elimination ideal defines the Zariski closure of this projection: $V(I\cap\mathbb{R}[x_1,\ldots,x_d]) = \overline{\pi_d(V(I))}$, where the closure is taken over $\mathbb{C}$ and we then restrict to real points.

This provides a way to test algebraic consequences: if $Q$ depends only on $x_1,\ldots,x_d$, we eliminate all other variables and check whether $Q$ belongs to the elimination ideal. We use this to test whether $Q$ can be derived from the axioms together with a candidate axiom extracted from decomposition. Elimination also filters out trivial candidates whose projections represent undesired boundary conditions under which $Q$ holds vacuously.

We present sufficient conditions for axiom recovery in Section \ref{recoverability}. The generalized algorithm is detailed in Algorithm \ref{alg:explanatory_axioms} in Appendix \ref{algorithm_details}.

\subsection{Handling Noisy Consequences - Numerical Extensions} \label{sec:noise}

The purpose of this study is to bridge the gap between machine-generated hypotheses and theory. Therefore, we must address the case where consequences (and possibly axioms) contain noise due to measurement error or numerical approximation in data-driven discovery. Accordingly, we make two key changes to our algorithm to account for noise.

\textbf{Numerical Irreducible Decomposition.} The algebraic decomposition method described in Section~\ref{sec:prelim} requires exact symbolic computations and is therefore not stable under coefficient perturbations of the consequences. We address this by replacing symbolic primary decomposition with \emph{numerical irreducible decomposition}, a well-established technique in numerical algebraic geometry \citep{bates2024numericalnonlinearalgebra}.

\begin{figure}[h]
  \centering
  \includegraphics[width=0.9\textwidth , trim={0cm 6cm 0cm 6cm}, clip]{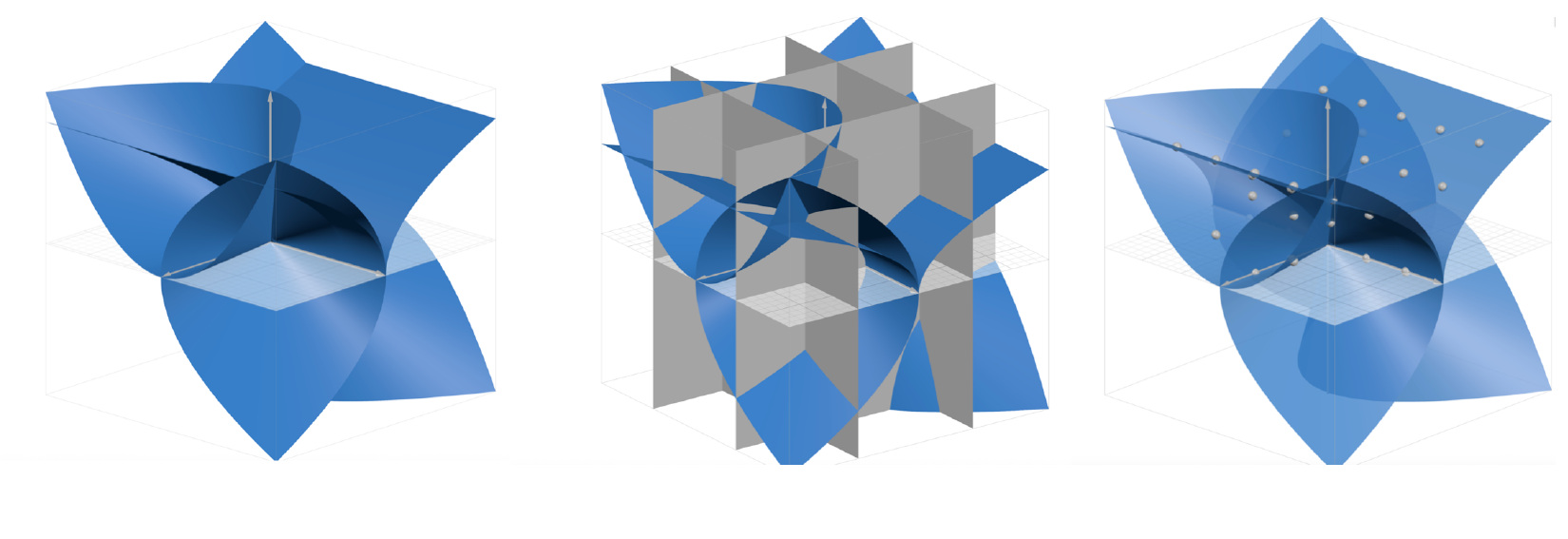}
\caption{Geometric visualization of slicing: intersecting generic hyperplanes with a variety until a zero-dimensional set is obtained. Left: the variety. Middle: the slicing process. Right: the zero-dimensional witness set.}
  \label{fig:slice}
\end{figure}

\begin{figure}[h]
  \centering
  \includegraphics[width=0.9\textwidth , trim={0cm 6cm 0cm 6cm}, clip]{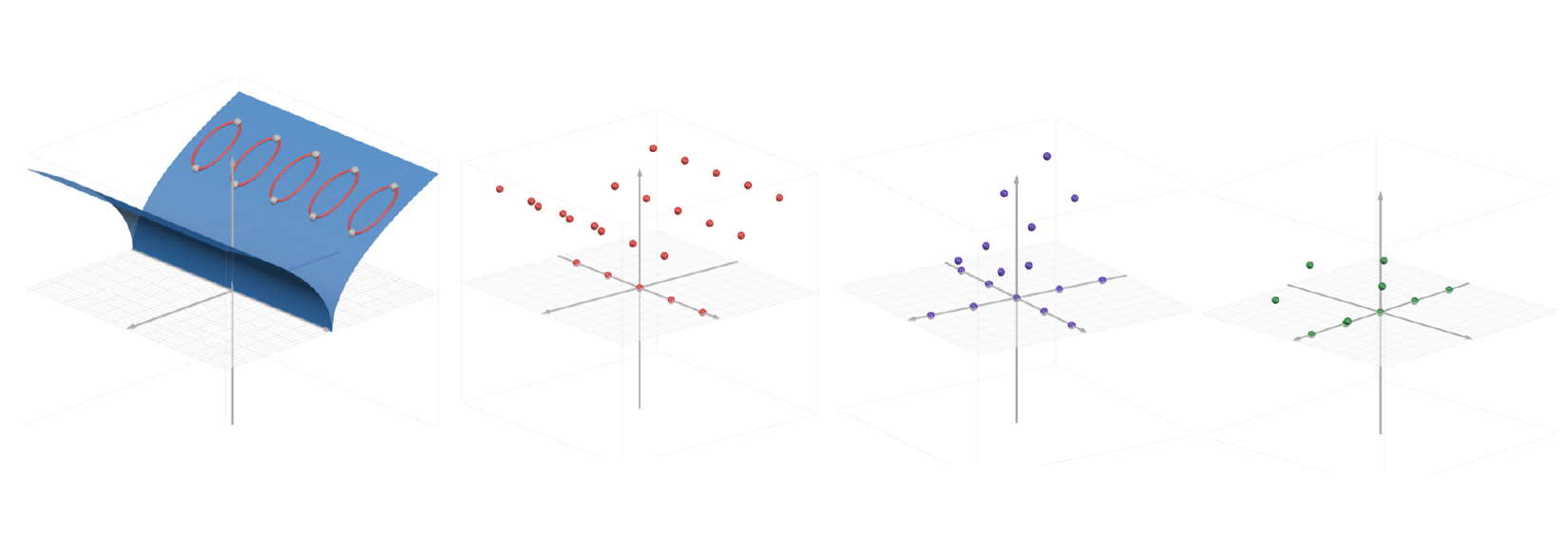}
\caption{Geometric visualization of curve tracing: defining homotopies between points to test whether sets of points lie on the same irreducible surface. Left: illustration of homotopy paths along the surface. Right: sets of points partitioned by irreducible component. Details in \cite{bates2024numericalnonlinearalgebra}.}
  \label{fig:curve_trace}
\end{figure}

The numerical approach proceeds in two stages. First, the variety is intersected with generic hyperplanes until a zero-dimensional set of witness points is obtained, as illustrated in Figure~\ref{fig:slice}. These witness points sample the variety at discrete locations. Second, homotopy continuation-based curve tracing determines which points lie on the same irreducible components (Figure~\ref{fig:curve_trace}). The method leverages fundamental constructs from numerical algebraic geometry, including witness sets \citep{sommese2002witness, hauenstein2020general} and monodromy-based component identification \citep{sommese2001monodromy}.

The output of numerical irreducible decomposition is a collection of witness sets---finite point samples on each irreducible component of the variety. While this provides geometric information about the decomposition, it does not directly yield the algebraic structure we seek (polynomial generators of each component). We therefore perform symbolic regression on the witness points to recover candidate polynomial relations.

\textbf{Symbolic Regression on Witness Sets.} Given witness points on an irreducible component, we fit polynomial relations that approximately vanish on those points. This problem---learning polynomials from data---has been studied extensively in the context of vanishing ideals \citep{pmlr-v28-livni13, heldt2009avi}.

For each removed axiom $A_i$ with symbolic form
\[
A_i(x_1,\dots,x_m) = \sum_{k} c_{ik} \, M_{ik}(x_1,\dots,x_m) = 0,
\]
we use its \emph{monomial support} $\{M_{ik}\}$ as a fixed regression basis. Given sampled points $\{p_j\}_{j=1}^N$ on a single irreducible component, we form the evaluation (Vandermonde-like) matrix $\Phi \in \mathbb{C}^{N\times K}$ with entries $\Phi_{jk} = M_{ik}(p_j)$. The coefficients $\mathbf{c}_i = (c_{i1},\dots,c_{iK})$ are estimated by solving
\[
\min_{\bm{c}_i} \Vert 
\Phi \, \mathbf{c}_i \Vert_2^2 \quad \text{subject to} \quad \|\mathbf{c}_i\|_2 = 1,
\]
which is equivalent to computing the right singular vector corresponding to the smallest singular value of $\Phi$ \citep{pmlr-v28-livni13}. This formulation identifies the (approximate) kernel of the evaluation map, recovering a polynomial that nearly vanishes on the sample set. Alternative approaches based on border basis methods \citep{heldt2009avi} or least-squares sensitivity analysis could be employed, though we adopt the SVD-based approach for its numerical stability and direct connection to nullspace identification.

The resulting polynomial $\hat{A}_i(x) = \sum_k \hat{c}_{ik} M_{ik}(x)$ is the best-fitting algebraic relation in the least-squares sense that holds on the numerically computed points. Each component yields its own fitted relation, enabling us to capture branch-specific structures when the noisy variety decomposes into multiple sheets.

\textbf{Choice of Variables and Basis Functions.}
The variable set and monomial basis used in regression serve as tunable hyperparameters. For each axiom $A_i$, we restrict regression to the variables actually appearing in $A_i$, denoted $\mathrm{vars}(A_i)$, and use its observed monomial support as basis functions. This prevents overfitting and ensures fitted coefficients are directly comparable to the symbolic axiom. Alternatively, one could learn variable subsets or expand the monomial basis to include cross-variable interactions or higher-degree terms, though we adopt the fixed-support approach for clarity and stability.

\begin{figure}[h]
  \centering
  \includegraphics[width=0.9\textwidth , trim={0cm 6cm 0cm 6cm}, clip]{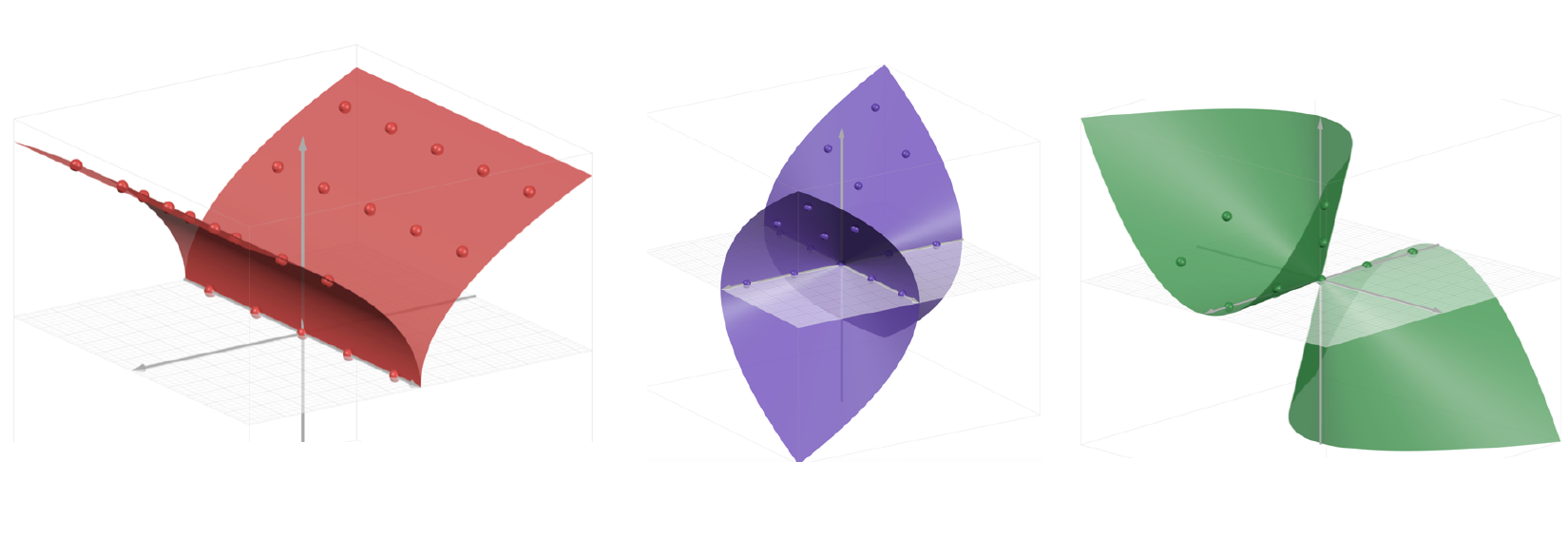}
\caption{Geometric visualization of symbolic regression output: recovered polynomial relations governing the irreducible surfaces identified by curve tracing (Figure~\ref{fig:curve_trace}).}
  \label{fig:sr}
\end{figure}

The fitted residual $\|\Phi \mathbf{c}_i\|_2 / \|\mathbf{c}_i\|_2$ serves as a numerical quality metric. Low residuals indicate that the recovered relation nearly vanishes on the component, suggesting the axiom (or its noisy perturbation) governs that portion of the variety. The resulting polynomials form our set of candidate axioms to test, as illustrated in Figure~\ref{fig:sr}.

\textbf{Robust Fitting.} In practice, witness sets may contain outliers due to numerical error accumulation in homotopy continuation. To improve robustness, we implement an iterative outlier removal procedure: after initial coefficient estimation, we remove points with the largest residuals (top 10\% by default) and refit. This process repeats until convergence or a maximum iteration count, yielding coefficient estimates less sensitive to numerical artifacts. Implementation details, including coefficient normalization and error metrics for comparing recovered and true axioms, are provided in the code repository.

\textbf{Reason with Axiom and Consequences containing Numerical Constants.} 

An alternative to algebraic derivability is to employ a theorem prover to determine whether the conjecture is logically derivable from the expanded set of axioms (i.e., the original axioms together with the abduced ones). Our approach follows the strategy proposed in AI-Descartes~\cite{descartes}, which introduced two complementary notions of derivability for scientific laws: a direct (symbolic) derivability test and an \emph{existential} one. The latter is designed for situations in which expressions inferred from data contain numerical constants (e.g., $Q = x/(0.709\,x + 0.157)$), since such expressions are unlikely to be matched syntactically (often the background theory contains only symbolic constants/parameters).
To do this, AI-Descartes abstracts each numeric constant into a new logical variable $c_i$, for example,
\[
Q = \frac{x}{0.709\,x + 0.157}
\quad\rightsquigarrow\quad
Q' = \frac{x}{c_1\,x + c_2},
\]
and constrains these variables through admissibility conditions $\mathcal{C}'$ (e.g., non-negativity). The constants are then existentially quantified so that the theorem prover may search over all admissible instantiations.

Under this abstraction, AI-Descartes defines a conjecture to be \emph{existentially derivable} from the background theory when
\begin{equation}\label{formulation:existential-derivability}
\exists c_1 \cdots \exists c_s.~ (\mathcal{C} \wedge \mathcal{A}) \rightarrow (f' \wedge \mathcal{C}'),
\end{equation}
that is, when some choice of the abstracted constants makes the conjecture consistent with both the axioms $\mathcal{A} = \{A_1, \cdots, A_{k+r}\}$ and the constraints $\mathcal{C}$ over the symbolic variables present in the axioms ($x_1,\cdots,x_n$, assumed universally quantified). 
This mechanism allows the system to assess the derivability of a data-derived expression with respect to the background theory for at least one admissible instantiation of its numerical constants, thus avoiding sensitivity to noise in their estimated values.

In our setting, some axioms are abduced from the conjecture and the original theory, and thus may also contain numerical constants. Consequently, we need to introduce existential variables not only for the numerical constants appearing in the conjecture but also for those occurring in the abduced axioms. This requires extending the existential derivability notion of Cornelio et al.~\cite{descartes} to handle numerical parameters on both sides of the derivation. Our resulting formulation, which accommodates constants in both the conjecture and the axioms, is as follows:
\begin{equation}
\exists c_1 \cdots \exists c_s~ 
\Big( 
    \mathcal{C}' \;\wedge\;
    \forall x_1 \cdots \forall x_n\;
        \big( (\mathcal{C} \wedge \mathcal{A} \wedge \mathcal{A}') \rightarrow f' \big)
\Big).
\end{equation}
where $\mathcal{A}'$ denotes the abduced axioms with their numerical constants replaced by existential variables, and $\mathcal{C}'$ denotes the admissibility constraints for these newly introduced variables.

In this way, we obtain a strict generalization of the existential derivability introduced in AI-Descartes~\cite{descartes}, which now accommodates numerical parameters in both the conjecture and the abduced axioms.

\section{Additional problem details and derivations}

\subsection*{Derivation of Carrier-Resolved Photo-Hall Effect}\label{Ap:photoHall}
The Carrier-Resolved Photo-Hall effect \cite{phothall} equation describes the relationship between various parameters of a semiconducting surface and is given by:

\[
Q \;:=\; H-\frac{r\,e\,\mu_P^{\,2}\bigl[p_0+\Delta n(1-\beta^2)\bigr]}{\sigma^{2}}=0,
\]
where \(\sigma=e\mu_P\bigl[p_0+\Delta n(1+\beta)\bigr]\) (carrier-resolved Photo-Hall effect~\cite{phothall}). We assume the axioms
\begin{align*}
A_1 &: \beta \mu_P - \mu_N & \text{ Mobility ratio definition} \\
A_2 &: \mu_H - r \mu & \text{ Hall factor correction} \\
A_3 &: p_h - p_0 - \Delta p & \text{ Hole concentration balance} \\
A_4 &: n - \Delta n & \text{ Electron concentration tracking}\\
A_5 &: \Delta p - \Delta n & \text{ Charge neutrality condition}\\
A_6 &: \sigma - e p_h \mu_P - e n \mu_N & \text{ Total conductivity sum}\\
A_7 &: H (p_h + \beta n)^2 e - r p_h + r \beta^2 n & \text{ Hall coefficient relation.}
\end{align*}
Here \(\mu_N\) and \(\mu_P\) are electron and hole drift mobilities, \(\beta=\mu_N/\mu_P\) is the mobility ratio (encoded by \(A_1\)), \(\mu_H\) is the Hall mobility with Hall scattering factor \(r=\mu_H/\mu\) (encoded by \(A_2\)), \(\Delta n\) and \(\Delta p\) are the electron and hole photo-carrier densities (equal in steady state by \(A_5\)), \(H\) is the Hall coefficient, \(\sigma\) is the conductivity, \(n\) is the electron density, \(p_0\) is the background hole density, and \(p_h\) is the hole density (for a p-type material).

Solving the coupled axiom system $A_1$-$A_7$ with the measured $\sigma(I)$ and $H(I)$ data simultaneously yields the densities and mobilities of both majority and minority carriers, plus recombination lifetime, diffusion length, and recombination coefficient, providing complete carrier characterization from a single measurement technique for photovoltaic material optimization.

To derive \(Q\), we eliminate \((\mu_N,\Delta p,n,p_h)\) using \(A_1,A_3\text{--}A_5\), i.e.,\ \(\mu_N=\beta\mu_P\), \(\Delta p=\Delta n\), \(n=\Delta n\), \(p_h=p_0+\Delta n\). Substituting into \(A_6\) gives \(\sigma=e\mu_P\bigl[p_0+\Delta n(1+\beta)\bigr]\), and substituting into \(A_7\) gives \(eH\bigl[p_0+\Delta n(1+\beta)\bigr]^2=r\bigl[p_0+\Delta n(1-\beta^2)\bigr]\); replacing \(\bigl[p_0+\Delta n(1+\beta)\bigr]^2\) by \(\sigma^2/(e^2\mu_P^2)\) yields \(Q=0\). To obtain the polynomial form over measured variables, we multiply the \(A_7\)-derived relation by \(e\mu_P\) and use \(e\mu_P(p_0+\Delta n)=\sigma-e\mu_P\beta\Delta n\) from the \(\sigma\)-equation, then expand to get
\[
r e\mu_P\Delta n\beta^2+r e\mu_P\Delta n\beta-r\sigma+e p_0\sigma H+e\Delta n\,\beta\,\sigma H+e\Delta n\,\sigma H=0.
\]
which is a polynomial version of the Photo-Hall equation. Equivalently, setting 
\begin{align*}
\alpha_1 &= -\mu e^2 \mu_P (\Delta n)^2 \beta H - \mu e^2 \mu_P p_0 \Delta n H - \mu e^2 \mu_P (\Delta n)^2 H
          - \mu e \mu_P p_0 r - \mu \mu_N e \Delta n r + \mu r \sigma,\\
\alpha_2 &= -e \mu_P^2 p_0 - \mu_N e \mu_P \Delta n - e \mu_P^2 \Delta n + \mu_P \sigma,\\
\alpha_3 &= -\mu e^2 \mu_P^2 \Delta n \beta H - p_h \mu e^2 \mu_P^2 H - \mu e \mu_P^2 \beta r
          + \mu \mu_N e \mu_P r + \mu e \mu_P^2 r - \mu_H e \mu_P^2,\\
\alpha_4 &= -\mu n e^2 \mu_P^2 \beta^2 H - \mu e^2 \mu_P^2 \Delta n \beta^2 H - 2 p_h \mu e^2 \mu_P^2 \beta H
          + \mu \mu_N e^2 \mu_P \Delta n \beta H - \mu e \mu_P^2 \beta^2 r + \mu \mu_N e^2 \mu_P p_0 H\\
         &\qquad + \mu \mu_N e^2 \mu_P \Delta n H - \mu \mu_N e \mu_P \beta r + \mu \mu_N^2 e r - \mu_N \mu_H e \mu_P,\\
\alpha_5 &= -\mu e^2 \mu_P^2 \Delta n \beta H - p_h \mu e^2 \mu_P^2 H - \mu e \mu_P^2 \beta r
          + \mu \mu_N e \mu_P r + \mu e \mu_P^2 r - \mu_H e \mu_P^2,\\
\alpha_6 &= \mu e \mu_P \Delta n \beta H + \mu e \mu_P p_0 H + \mu e \mu_P \Delta n H - \mu \mu_P \beta r
          + \mu \mu_N r - \mu_H \mu_P,\\
\alpha_7 &= \mu e \mu_P^2.
\end{align*}

we obtain 
\[
Q = \sum_{i=1}^7 \alpha_i A_{i}
\]

\subsection*{Illustrating nuances of axiom formatting and multipliers with Kepler's third law}

From section \ref{sec:mult_missing_axioms}, recall the equation for Kepler's third law, which relates the orbital period of two celestial bodies to their distances and masses. It can be written as
\[
  Q := p - \sqrt{\frac{(d_1+d_2)^3}{G(m_1+m_2)}} = 0,
\]
where $p$ is the orbital period (scaled by a factor of $2\pi$), $m_1,m_2$ the masses, and $d_1,d_2$ their distances from the center of mass. For circular orbits, we use the following axioms (with $p$ non-dimensionalized so that $\omega p=1$):
\begin{align}
    A_1 &: (d_1+d_2)^2F_g - Gm_1m_2 = 0 \label{A1}\\
    A_2 &: F_c - m_2d_2\omega^2 = 0 \label{A2}\\
    A_3 &: F_g - F_c = 0 \label{A3}\\
    A_4 &: \omega p - 1 = 0 \label{A4}\\
    A_5 &: m_1d_1 - m_2d_2 = 0. \label{A5}
\end{align}
Equation \eqref{A1} gives gravitational force, \eqref{A2} centrifugal force, \eqref{A3} equates them, \eqref{A4} relates frequency and period, and \eqref{A5} defines the center of mass. To derive $Q$, we eliminate $d_1$ using \eqref{A5}, substitute $p=1/\omega$ from \eqref{A4}, and replace $\omega^2$ by combining \eqref{A2}–\eqref{A3} with \eqref{A1}. This yields
\[
  G(m_1+m_2)p^2 - (d_1+d_2)^3 = 0,
\]
which is Kepler's law in polynomial form. Equivalently, setting
\[
\alpha_1=-p^2,\ \ 
\alpha_2=p^2(d_1+d_2)^2,\ \ 
\alpha_3=-p^2(d_1+d_2)^2,\ \ 
\alpha_4=m_2d_2(\omega p+1)(d_1+d_2)^2,\ \ 
\alpha_5=-d_2^2,
\]
we obtain
$$Q = \alpha_1A_1 + \alpha_2A_2 + \alpha_3A_3 + \alpha_4A_4 + \alpha_5A_5.$$

Assume that we are missing knowledge of the equation of gravitational force -- axiom $A_1$ --, but we have Kepler's law $Q$ at hand (possibly derived from sufficient data; see \cite{AI-Feynman, descartes, corywright2024evolving}). 
At this point, not only $A_1$ but also $\alpha_1, \ldots, \alpha_5$ are not known. We will assume they exist and attempt to find them from $Q$ and the remaining known axioms $A_2-A_5$. Here we view $\alpha_1A_1$ as the (unknown) residual of $Q$ with respect to $A_2, \ldots, A_5$. We define the ideal $I = \langle A_2,A_3,A_4,A_5, Q\rangle$. This is the same as the ideal $\langle A_2,A_3,A_4,A_5,\alpha_1 A_1\rangle$ as $Q = \sum_{i=1}^5 \alpha_i A_i$. 
Geometrically, $V(A_2,A_3,A_4,A_5,\alpha_1 A_1) = V(A_2,A_3,A_4,A_5) \cap V(\alpha_1 A_1)$. Since the latter is reducible, we have more irreducible components than we have for $V(A_2, A_3, A_4, A_5)$. Computing a primary decomposition of $I$ then gives us the following associated primes:

\resetvarieties
\begingroup
\begin{tabularx}{\linewidth}{@{}Y@{}}
\sectionrow{Decomposition}
\vrowalt{\(\gen{d_2,\; m_1,\; F_g,\; F_c,\; \omega p - 1}\)}
\vrowalt{\(\gen{m_2,\; d_1,\; F_g,\; F_c,\; \omega p - 1}\)}
\vrowalt{\(\gen{m_2,\; m_1,\; F_g,\; F_c,\; \omega p - 1}\)}
\vrowalt{%
\(\fixedgenaligned{
  &F_c - F_g,\; m_1 d_1 - m_2 d_2,\; \omega p - 1,\; F_g p - \omega m_2 d_2,\; F_g p^2 - m_2 d_2,\\
  &\hlgen{F_g(d_1+d_2)^2 - m_1 m_2 G},\;
    d_1(d_1+d_2)^2 - m_2 p^2 G,\;
    m_1 p^2 G - d_2 (d_1+d_2)^2,\\
  & \omega d_1^2(d_1+d_2)^2 - m_1 p G,\;
    \omega d_1^3 - 3 \omega d_1 d_2^2 - 2 \omega d_2^3 + 2 m_1 p G - m_2 p G
}\)%
}
\sectionrow{Generators selected by Reasoning module}
\rowcolor{gray!6}
\makebox[\linewidth]{\(\color{BrickRed!80!black}\mathbf{F_g(d_1+d_2)^2 - m_1 m_2 G}\)}\\[4pt]
\end{tabularx}
\endgroup

The generators of these irreducible components/ideals represent key equations that define each irreducible component. We iterate through each generator $\hat{A_1}$ and compute a Gr\"obner basis to eliminate the variables not appearing in $Q$. We find that only the correct expression  - {\(\color{BrickRed!80!black}\mathbf{F_g(d_1+d_2)^2 - m_1 m_2 G}\)} - can derive $Q$ and is therefore the only candidate returned by our system. 

\textbf{Axiom Formatting and Multipliers.} Since the reducibility is introduced by $V(\alpha_i A_i)$, we do not distinguish between $\alpha_i$ and $A_i$ when searching for $i$th axiom candidates. Additionally, we may return axioms in alternative forms; the exact expression is not guaranteed. For example, consider the same Kepler axioms and assume we are missing axiom $A_4$ instead. Then the relevant ideals in the primary decomposition are:

\resetvarieties
\begingroup
\setlength{\tabcolsep}{8pt}%
\renewcommand{\arraystretch}{1.28}%
\arrayrulecolor{white}%

\begin{tabularx}{\linewidth}{@{}Y@{}}
\sectionrow{Decomposition}

\vrowalt{%
\(\fixedgenaligned{
  F_c - F_g,\; m_1 d_1 - m_2 d_2,\; \hlgen{\omega p - 1},\; F_g p - \omega m_2 d_2,\; \hlgen{F_g p^2 - m_2 d_2},\\
  \qquad F_g(d_1{+}d_2)^2 - m_1 m_2 G,\; d_1(d_1{+}d_2)^2 - m_2 p^2 G,\\
  \qquad m_1 p^2 G - d_2(d_1{+}d_2)^2,\; \omega \,d_2(d_1{+}d_2)^2 - m_1 p G,\\
  \qquad \omega (d_1 - 2 d_2)(d_1{+}d_2)^2 + 2 m_1 p G - m_2 p G
}\)%
}

\vrowalt{%
\(\fixedgenaligned{
  F_c - F_g,\; m_1 d_1 - m_2 d_2,\; \hlgen{wp + 1},\; F_g p + \omega m_2 d_2,\; F_g p^2 - m_2 d_2,\\
  \qquad F_g(d_1{+}d_2)^2 - m_1 m_2 G,\; d_1(d_1{+}d_2)^2 - m_2 p^2 G,\\
  \qquad m_1 p^2 G - d_2(d_1{+}d_2)^2,\; \omega \,d_2(d_1{+}d_2)^2 + m_1 p G,\\
  \qquad \omega(d_1 - 2 d_2)(d_1{+}d_2)^2 - 2 m_1 p G + m_2 p G
}\)%
}

\sectionrow{Generators selected by Reasoning module}
\rowcolor{gray!6}
\makebox[\linewidth]{\(\color{BrickRed!80!black}\mathbf{\omega p - 1}\)}\\[4pt]
\rowcolor{gray!12}
\makebox[\linewidth]{\(\color{BrickRed!80!black}\mathbf{F_g p^2 - m_2 d_2}\)}\\[4pt]
\rowcolor{gray!6}
\makebox[\linewidth]{\(\color{BrickRed!80!black}\mathbf{\omega p + 1}\)}\\[4pt]
\end{tabularx}
\endgroup

The highlighted polynomials are returned by our system as being axiom candidates that derive $Q$ along with the remaining axioms. The other generators have been omitted for readability, and cannot derive $Q$. Although the correct expression $\omega p-1$ is present, additional terms are present. The term $F_gp^2 - m_2d_2$ is a restatement of $A_2$ using $A_3$ and the unknown $\omega p-1$ substituted in. It contains the same information as $wp-1$ since the former is already known. The term $\omega p+1$ is an artifact of the fact that $\alpha_4$ as noted before is $m_2d_2(\omega p+1)(d_1+d_2)^2$. Note that the variety $V(m_2d_2(\omega p+1)(d_1+d_2)^2)$ is reducible and can be written as the union: $V(m_2)\cup V(d_2) \cup V(\omega p+1) \cup V((d_1+d_2)^2)$. Therefore, $\omega p+1$ appears as a viable axiom candidate since there is an algebraic derivation of $Q$ from the remaining axioms along with $\omega p+1$. Incidentally, the remaining terms $m_2$, $d_2$, and $(d_1+d_2)^2$ also show up in the associated primes:

\resetvarieties
\begingroup
\setlength{\tabcolsep}{8pt}%
\renewcommand{\arraystretch}{1.28}%
\arrayrulecolor{white}%

\begin{tabularx}{\linewidth}{@{}Y@{}}
\sectionrow{Decomposition}

\vrowalt{%
\(\fixedgenaligned{
  F_c - F_g,\; m_2^2,\; \hlgen{(d_1{+}d_2)^2},\; m_1 d_2 + d_1 m_2 + 2 m_2 d_2,\; m_1 m_2,\\
  \qquad m_1 d_1 - m_2 d_2,\; m_1^2,\; F_g m_2,\; F_g m_1,\; F_g^2,\; F_g - \omega^2 m_2 d_2
}\)%
}

\vrowalt{\(\gen{\hlgen{d_2},\; m_1,\; F_g,\; F_c}\)}
\vrowalt{\(\gen{\hlgen{m_2},\; d_1,\; F_g,\; F_c}\)}

\sectionrow{Generators selected by Reasoning module}
\rowcolor{gray!6}
\makebox[\linewidth]{\(\color{BrickRed!80!black}\mathbf{(d_1{+}d_2)^2}\)}\\[4pt]
\rowcolor{gray!12}
\makebox[\linewidth]{\(\color{BrickRed!80!black}\mathbf{d_2}\)}\\[4pt]
\rowcolor{gray!6}
\makebox[\linewidth]{\(\color{BrickRed!80!black}\mathbf{m_2}\)}\\[4pt]

\end{tabularx}
\endgroup

In order to filter out these terms, for non-numerical examples, we apply an additional layer of filtering detailed in Appendix \ref{Ap:filtering}.

\subsection{Detailed Breakdown of Single Axiom Removal Experiment}\label{Ap:single_axiom_table}

We tested the system by removing one axiom at a time and asking it to recover the missing axiom, reporting our findings in Table \ref{detail_axiom}. Across 12 problems, the method succeeded in 73 out of 75 cases (97.3\%). Recovery was perfect for Kepler (5/5), Time Dilation (5/5), Escape Velocity (5/5), Light Damping (5/5), Hagen–Poiseuille (4/4), Neutrino Decay (5/5), Simple Harmonic Oscillator (6/6), Carrier-Resolved Photo-Hall (7/7), and Compton Scattering (10/10). Two additional failures occurred in the Relativistic Laws system (5/7 recovered), where axioms $A_4$ (light path geometry) and $A_7$ (parallel light travel time) were not recovered; these quadratic constraints factor into linear generators during primary decomposition, and the system returns the factored forms rather than the original quadratic axioms (see Section 3.3 for discussion).

\begingroup
\arrayrulecolor{black} 
\renewcommand{\arraystretch}{1.12}

\begin{longtable}{@{}l l l c@{}}
\caption{Results across problems; axioms are listed per problem as \(A_1, A_2, \dots\). "Recovered" indicates whether the method successfully derived the removed axiom.}
\label{detail_axiom}\\
\toprule
Problem & Axiom & Expression & Recovered \\
\midrule
\endfirsthead

\toprule
Problem & Axiom & Expression & Recovered \\
\midrule
\endhead

\bottomrule
\endfoot

\bottomrule
\endlastfoot

Hagen--Poiseuille & A1 & $u - c_0 - c_2 r^2$ & \checkmark \\
                  & A2 & \(\mu \frac{\partial}{\partial r}\!\Bigl(r \frac{\partial u}{\partial r}\Bigr) - r \frac{dp}{dx}\) & \checkmark \\
                  & A3 & $c_0 + c_2 R^2$     & \checkmark \\
                  & A4 & $L \frac{dp}{dx} = -\Delta p$ & \checkmark \\\midrule

Kepler & A1 & $(d_1{+}d_2)^2 F_g - m_1 m_2$ & \checkmark \\
       & A2 & $F_c - m_2 d_2 \omega^2$      & \checkmark \\
       & A3 & $F_c - F_g$                   & \checkmark \\
       & A4 & $\omega p - 1$                & \checkmark \\
       & A5 & $m_1d_1 - m_2d_2$             & \checkmark \\\midrule

Time Dilation & A1 & $c\,dt_0 - 2 d$            & \checkmark \\
              & A2 & $4 L^2 - 4 d^2 - v^2 dt^2$ & \checkmark \\
              & A3 & $f_0 dt_0 - 1$             & \checkmark \\
              & A4 & $f\,dt - 1$                & \checkmark \\
              & A5 & $c\,dt - 2L$               & \checkmark \\\midrule

Escape Velocity & A1 & $K_i - \tfrac{1}{2} m v_e^2$ & \checkmark \\
                & A2 & $K_f = 0$                     & \checkmark \\
                & A3 & $U_i r + GM m$                & \checkmark \\
                & A4 & $U_f = 0$                     & \checkmark \\
                & A5 & $K_i + U_i - (K_f + U_f)$     & \checkmark \\\midrule

Light Damping & A1 & $S r^2 - q_c^2 a_p^2 \sin^2\theta$                   & \checkmark \\
              & A2 & $dA - 2\pi r^2 \sin\theta \, d\theta$                & \checkmark \\
              & A3 & $P - \int_0^\pi S\, dA$                              & \checkmark \\
              & A4 & \(\frac{4}{3} - \int_0^\pi \sin^3\theta \, d\theta\) & \checkmark \\
              & A5 & $a_p^2 - \tfrac{1}{2} w^4 x_0^2$                     & \checkmark \\\midrule

Neutrino Decay & A1 & $p_\nu - p_\mu$              & \checkmark \\
               & A2 & $E_\pi - m_\pi$               & \checkmark \\
               & A3 & $E_\nu - p_\nu$               & \checkmark \\
               & A4 & $E_\pi - E_\mu - E_\nu$       & \checkmark \\
               & A5 & $E_\mu^2 - p_\mu^2 - m_\mu^2$ & \checkmark \\\midrule

Simple Harmonic Osc. & A1 & $a_d - g\sin{\theta}$    & \checkmark \\
                     & A2 & $d - L\theta$            & \checkmark \\
                     & A3 & $T\omega - 2\pi$         & \checkmark \\
                     & A4 & $d\omega^2 - a_d$        & \checkmark \\
                     & A5 & $T_j - jT$               & \checkmark \\
                     & A6 & $\sin{\theta} - \theta$  & \checkmark \\\midrule

Carrier-Resolved PH & A1 & $\beta \mu_P - \mu_N$                         & \checkmark \\
                    & A2 & $\mu_H - r \mu$                               & \checkmark \\
                    & A3 & $p_h - p_0 - \Delta p$                        & \checkmark \\
                    & A4 & $n - \Delta n$                                & \checkmark \\
                    & A5 & $\Delta p - \Delta n$                         & \checkmark \\
                    & A6 & $\sigma - e\,p_h \mu_P - e\,n \mu_N$          & \checkmark \\
                    & A7 & $H (p_h + \beta n)^2 e - r p_h + r \beta^2 n$ & \checkmark \\\midrule

Relativistic Laws & A1 & $f_0 t_0 - 1$                                      & \checkmark \\
                  & A2 & $f t - 1$                                          & \checkmark \\
                  & A3 & $ct_0 - 2L_0$                                      & \checkmark \\
                  & A4 & $\frac{c^2t^2}{4} - L_0^2 - \frac{v^2t^2}{4}$     & \ding{55} \\
                  & A5 & $m_0 u_0 - m u_y$                                  & \checkmark \\
                  & A6 & $u_0 t_0 - u_y t$                                  & \checkmark \\
                  & A7 & $t(c^2-v^2) - 2Lc$                                 & \ding{55} \\\midrule

Hall Effect & A1 & $F_m - q_e v B$ & \checkmark \\
            & A2 & $F_e - q_e E$   & \checkmark \\
            & A3 & $F_m - F_e$     & \checkmark \\
            & A4 & $E h - U_H$     & \checkmark \\
            & A5 & $v\,dt - L$     & \checkmark \\
            & A6 & $I\,dt - Q$     & \checkmark \\
            & A7 & $Q - N q_e$     & \checkmark \\\midrule

Inelastic Relativistic Collision & A1 & $p_m^2(c^2 - v_m^2) - m_m^2 v_m^2 c^2$     & \checkmark \\
                    & A2 & $E_m^2 - (m_m c^2)^2 - (p_m c)^2$          & \checkmark \\
                    & A3 & $E_r - m_r c^2$                            & \checkmark \\
                    & A4 & $E_c^2 - (m_c c^2)^2 - (p_c c)^2$          & \checkmark \\
                    & A5 & $2 E_m E_r - E_c^2 + E_m^2 + E_r^2$        & \checkmark \\
                    & A6 & $p_c - p_m$                                & \checkmark \\
                    & A7 & $v_m - \tfrac{4}{5}c$                      & \checkmark \\
                    & A8 & $m_r - m_m$                                & \checkmark \\
                    & A9 & $p_c^2(c^2 - v_c^2) - m_c^2 v_c^2 c^2$     & \checkmark \\\midrule

Compton Scattering & A1  & $E_1 + E e_1 - E_2 - E e_2$               & \checkmark \\
                   & A2  & $E_1 - h f_1$                             & \checkmark \\
                   & A3  & $E_2 - h f_2$                             & \checkmark \\
                   & A4  & $p_1 c - h f_1$                           & \checkmark \\
                   & A5  & $p_2 c - h f_2$                           & \checkmark \\
                   & A6  & $\lambda_1 f_1 - c$                       & \checkmark \\
                   & A7  & $\lambda_2 f_2 - c$                       & \checkmark \\
                   & A8  & $E e_1 - m_e c^2$                         & \checkmark \\
                   & A9  & $E e_2^2 - c^2 p e_2^2 - m_e^2 c^4$       & \checkmark \\
                   & A10 & $p e_2^2 - p_2^2 - p_1^2 + 2 p_1 p_2 \cos\theta$ & \checkmark \\

\end{longtable}
\endgroup

\section{Additional Technical Details}

\subsection{Differences Between Real and Complex Solutions}\label{Ap:Complex_numbers}

Let $A_1, \ldots, A_k$ be a collection of polynomials with $A_1 = 0, \ldots, A_k = 0$ being the associated axioms.
Let $\mathcal{A}$ stand for the set $\{A_1, \ldots, A_k\}$.
Let $V_R(\mathcal{A})$, respectively $V_C(\mathcal{A})$,  be the set of real, respectively complex, solutions of the equations $A_1 = 0, \ldots, A_k = 0$. Note that $V_R(\mathcal{A})$ is the same as the variety $V(\mathcal{A})$ defined in the main body of the paper.

First note that $V_C(\mathcal{A},Q) = V_C(\mathcal{A})$ if and only if $Q = 0$ for all complex solutions of $A_1 = 0, \ldots, A_k = 0$. Further, if $Q \in I(\mathcal{A})$, then $V_C(\mathcal{A},Q) = V_C(\mathcal{A})$. However, the converse is not true. By Hilbert's Nullstellensatz, if $V_C(\mathcal{A},Q) = V_C(\mathcal{A})$, then $Q^m \in I(\mathcal{A})$ for some positive integer $m$ (but $m$ need not equal 1). But $Q^m(\bar x) = 0$ if and only $Q(\bar x) = 0$. Accordingly, if we say that $Q = 0$ is derivable from $\mathcal{A}$ if and only if $Q^m \in I(\mathcal{A})$, then the notions of ``a consequence of'' and ``derivable from'' are identical assuming complex-valued solutions are relevant.

On the other hand, if we are only interested in real-valued solutions, the situation is more complicated. 
As before, $V_R(\mathcal{A},Q) = V_R(\mathcal{A})$ if and only if $Q = 0$ for all real solutions of $A_1 = 0, \ldots, A_k = 0$. Further, if $Q \in I(\mathcal{A})$, then $V_R(\mathcal{A},Q) = V_R(\mathcal{A})$. However, the converse is not true. 
If $V_R(\mathcal{A},Q) = V_R(\mathcal{A})$, then $Q^{2m} + S  \in I(\mathcal{A})$ for some positive integer $m$ and some polynomial $S$ that is a sum of squares of polynomials in $R$.

\subsection{Further Filtering for Projections}\label{Ap:filtering}

While testing ideal membership via Gr\"obner basis elimination (as described in Section~\ref{sec:prelim}) establishes that $Q$ is derivable from the candidate axioms together with the known axioms, this criterion alone can admit \emph{trivial} candidates that make the system degenerate. For instance, a candidate axiom that simply sets a variable to zero (e.g., $m_2 = 0$) may technically allow $Q$ to be derived, but only because it collapses the variety onto a lower-dimensional subspace where $Q$ holds vacuously. To filter such cases, we introduce a stronger derivability test based on \emph{exact containment} in the elimination ideal.

\paragraph{Exact Containment Test.}
Let $\mathcal{A} = \{A_1, \ldots, A_k\}$ denote the known axioms, $\hat{A}_{k+1}$ a candidate axiom, and $Q$ the target hypothesis defined over variables $\{x_1, \ldots, x_d\}$. We form the augmented ideal $J = \langle A_1, \ldots, A_k, \hat{A}_{k+1} \rangle$ and compute its reduced Gr\"obner basis $\mathcal{G}$ with respect to a lexicographic term order. The elimination ideal $J \cap \mathbb{R}[x_1, \ldots, x_d]$ is generated by $\mathcal{G}' = \mathcal{G} \cap \mathbb{R}[x_1, \ldots, x_d]$.

\textbf{Weak test (ideal membership):} We accept $\hat{A}_{k+1}$ if $Q \in \langle \mathcal{G}' \rangle$, i.e., if $Q$ reduces to zero modulo $\mathcal{G}'$. This confirms that $Q$ is derivable from the augmented system.

\textbf{Strong test (exact containment):} We accept $\hat{A}_{k+1}$ only if $Q \in \mathcal{G}'$, i.e., if $Q$ appears \emph{exactly} as an element of the Gr\"obner basis (up to scalar multiplication). This ensures that the projection onto $\mathrm{vars}(Q)$ yields $Q$ directly, rather than a stronger polynomial from which $Q$ follows as a consequence.

\paragraph{Geometric Interpretation.}
Geometrically, the weak test verifies that $V(\mathcal{A} \cup \{\hat{A}_{k+1}\})$ projects onto $V(Q)$ when restricted to the coordinates of $Q$. However, this projection may pass through a lower-dimensional intermediate variety. The strong test ensures that the projection is \emph{minimal}: we project directly onto $V(Q)$ without factoring through a smaller-dimensional subvariety.

For example, consider Kepler's third law in polynomial form $Q: G(m_1+m_2)p^2 - (d_1+d_2)^3 = 0$, defined over variables $\{p, m_1, m_2, d_1, d_2\}$. A trivial candidate axiom $\hat{A}_{k+1}: m_2 = 0$ would satisfy the weak test, since setting $m_2 = 0$ implies $Gm_1p^2 - (d_1+d_2)^3 = 0$, which algebraically contains $Q$. However, $Q$ itself does not appear in the elimination ideal; instead, we derive a stronger constraint $Gm_1p^2 - (d_1+d_2)^3 = 0$ defined over fewer variables (without $m_2$). The strong test rejects such candidates, filtering out degenerate cases where the system collapses onto boundary conditions.

\paragraph{Limitations and Practical Implementation.}
The strong test is not foolproof. If the background theory $\mathcal{A}$ already contains an axiom defined over a strict subset of $\mathrm{vars}(Q)$, then the elimination ideal will necessarily include polynomials over fewer variables than $Q$. For instance, in the Kepler system, the center-of-mass axiom $m_1 d_1 - m_2 d_2 = 0$ is defined over $\{m_1, m_2, d_1, d_2\}$, a proper subset of the variables in Kepler's third law. Consequently, $\mathcal{G}'$ will contain polynomials over these variables, and $Q$ may not appear exactly in $\mathcal{G}'$ even for legitimate candidate axioms.

To handle this, we adopt a two-tier filtering strategy:
\begin{enumerate}
    \item \textbf{Primary filter (strong test):} Accept candidates where $Q \in \mathcal{G}'$ exactly. These are high-confidence results.
    \item \textbf{Secondary filter (weak test):} Accept candidates where $Q \in \langle \mathcal{G}' \rangle$ but $Q \notin \mathcal{G}'$, flagging them as lower-confidence. These may include legitimate axioms whose projection is mediated by background constraints.
\end{enumerate}
In our experiments, we save both sets of candidates, noting which satisfy the strong versus weak test. For systems with hierarchical variable dependencies (e.g., conservation laws involving subsets of state variables), the weak test often recovers correct axioms that the strong test would reject.

\subsection{Algorithm Implementation Details}\label{algorithm_details}
Algorithm~\ref{alg:explanatory_axioms} describes our procedure for identifying missing axioms that render a target hypothesis $Q$ derivable. The algorithm proceeds in three phases that parallel our methodological framework: \textbf{Encode}, \textbf{Decompose}, and \textbf{Reason}, with separate pathways for exact (symbolic) and noisy (numerical) scenarios. We now describe each step in detail.

\paragraph{Input and Initialization.} 
The input consists of a finite set of polynomial axioms $\mathcal{A} = \{A_1, \ldots, A_k\}$ in variables $\mathcal{X} = \{x_1, \ldots, x_n\}$, a hypothesis polynomial $Q$ over a restricted subset of variables $\{x_1,\ldots,x_d\}$ (where $d \leq n$), and a flag \textit{noisy} indicating whether the consequences contain numerical error. The algorithm maintains a set $\mathcal{A}_{\text{expl}}$ of explanatory candidates, initially empty.

\paragraph{Encode.} 
We form the polynomial ideal $I = \langle A_1,\ldots,A_k,Q\rangle$. Geometrically, $V(I)$ corresponds to the variety of solutions simultaneously satisfying the known axioms and the hypothesis. If $Q$ is derivable from $\mathcal{A}$, then $V(I)$ coincides with $V(\mathcal{A})$; otherwise, additional structure appears in the decomposition. This encoding step is identical for both exact and noisy cases.

\paragraph{Decompose.} 
The decomposition strategy differs based on whether consequences are exact or noisy:

\textit{Exact case (symbolic):} We compute a primary decomposition 
\[
I = I_1 \cap \cdots \cap I_r,
\]
yielding associated primes $P_j = \sqrt{I_j}$. Each $P_j$ describes an irreducible component of the solution space. The generators of $P_j$ form a finite set $\{\hat{A}_{k+1}^{(j,1)}, \ldots, \hat{A}_{k+1}^{(j,m_j)}\}$ of potential candidate axioms. 

In our experiments, we use Singular  \cite{Singular}, a computer algebra software, for computing the primary decompositions. We also have the functionality for using Macaulay2 \cite{M2}, another standard computer algebra software, though Singular's default primary decomposition implementation is more efficient.  

\textit{Noisy case (numerical):} We apply numerical irreducible decomposition~\citep{bates2024numericalnonlinearalgebra} to compute witness sets $\{W_1, \ldots, W_s\}$, where each $W_i$ is a finite collection of sample points on an irreducible component. For each witness set $W_i$ and each removed axiom $A_\ell \in \mathcal{A}$, we perform symbolic regression: we form the evaluation matrix $\Phi$ using the monomial support of $A_\ell$ evaluated at points in $W_i$, then compute the right singular vector corresponding to the smallest singular value to obtain fitted coefficients $\hat{\mathbf{c}}_\ell$. This yields a candidate axiom $\hat{A}_{k+1}^{(i,\ell)} = \sum_k \hat{c}_{\ell k} M_{\ell k}$ for each component-axiom pair. 

We use Bertini \cite{BHSW06}, an alternate numerical algebraic geometry package with parallelization implementation to compute the witness sets. We also build in the optionality to use Macaulay2's numerical algebraic geometry package \cite{m2nag}. 

\paragraph{Reason.} 
For each candidate axiom $\hat{A}_{k+1}$ obtained from decomposition (or sets of candidate axioms if we know that multiple axioms are missing), we test whether it (or they) enable(s) derivation of $Q$. The reasoning method depends on whether numerical constants are present:

\textit{Exact case (algebraic projection):} We form the augmented ideal $J = \langle A_1,\ldots,A_k,\hat{A}_{k+1} \rangle$ (resp. $\langle A_1,\ldots,A_k,\hat{A}_{k+1}, \ldots \hat{A}_{k+l} \rangle$ for testing a set of multiple candidate axioms) and compute its reduced Gr\"obner basis $\mathcal{G}$ with respect to a lexicographic order on $\mathcal{X}$. We project $\mathcal{G}$ to the coordinate ring of the observed variables: $\mathcal{G}' = \mathcal{G} \cap \mathbb{R}[x_1,\ldots,x_d]$. If $Q \in \mathcal{G}'$, then $\hat{A}_{k+1}$ is an \emph{explanatory axiom} for $Q$ and is added to $\mathcal{A}_{\text{expl}}$. The Gr\"obner basis computations here are done entirely in Macaulay2 \cite{M2}.

\textit{Noisy case (existential reasoning):} Following the approach of AI-Descartes~\citep{descartes}, we abstract numerical constants in both the hypothesis $Q$ and the candidate axiom $\hat{A}_{k+1}$ (resp. $\hat{A}_{k+1},\ldots , \hat{A}_{k+l}$ for testing a set of multiple candidate axioms) into existentially quantified variables $c_1, \ldots, c_s$, yielding $Q'$ and $\hat{A}_{k+1}'$. We encode the derivability query as:
\[
\exists c_1 \cdots \exists c_s~ 
\Big( 
    \mathcal{C}' \;\wedge\;
    \forall x_1 \cdots \forall x_n\;
        \big( (\mathcal{C} \wedge \mathcal{A} \wedge \hat{A}_{k+1}') \rightarrow Q' \big)
\Big),
\]
where $\mathcal{C}$ and $\mathcal{C}'$ denote admissibility constraints on the variables. We submit this query to a theorem prover, KeYmaera X~\citep{keymaerax2017} in our implementation. If the prover returns \texttt{Proved}, then $\hat{A}_{k+1}$ (or the set of candidates) is added to $\mathcal{A}_{\text{expl}}$.

\paragraph{Output.} 
The algorithm returns the set $\mathcal{A}_{\text{expl}}$ of all candidate axioms that, when added to the background theory $\mathcal{A}$, enable derivation of $Q$. In practice, we also return metadata including fitted coefficients (for the noisy case), residual norms (quality metrics), and the component indices from which each candidate was extracted.

\paragraph{Computational Complexity.}
The dominant computational costs are: (i) primary decomposition or witness set computation (doubly exponential in the worst case for symbolic methods~\citep{doubleexp}, polynomial per witness set for numerical methods), (ii) symbolic regression (linear algebra on $N \times K$ matrices for $N$ witness points), and (iii) Gr\"obner basis computation (doubly exponential worst-case~\citep{buchberger}, though often tractable in practice for structured systems) or theorem proving. Our experiments demonstrate practical feasibility on systems with up to 16 variables and 10 axioms.

\begin{algorithm}
\caption{Identifying Explanatory Axioms via Decomposition and Reasoning}
\begin{algorithmic}[1]
\Require Axioms $\mathcal{A} = \{A_1, \ldots, A_k\}$ over variables $\mathcal{X} = \{x_1, \ldots, x_n\}$, hypothesis $Q$ over $\{x_1, \ldots, x_d\}$, flag \textit{noisy}
\Ensure Set $\mathcal{A}_{\text{expl}}$ of explanatory axioms
\State Initialize $\mathcal{A}_{\text{expl}} \gets \emptyset$
\State \textbf{Encode:} Form ideal $I \gets \langle A_1, \ldots, A_k, Q \rangle$
\State \textbf{Decompose:}
\If{\textit{noisy}}
    \State Compute witness sets $\{W_1, \ldots, W_s\}$ via numerical irreducible decomposition
    \For{each witness set $W_i$}
        \For{each removed axiom $A_\ell \in \mathcal{A}$}
            \State Form evaluation matrix $\Phi$ from monomial support of $A_\ell$ on $W_i$
            \State Compute $\hat{\mathbf{c}}_\ell$ via smallest singular value of $\Phi$
            \State Construct candidate $\hat{A}_{k+1}^{(i,\ell)} \gets \sum_k \hat{c}_{\ell k} M_{\ell k}$
            \State \textbf{Reason:} Abstract numerical constants in $\hat{A}_{k+1}^{(i,\ell)}$ and $Q$ to obtain $\hat{A}_{k+1}^{(i,\ell)'}$ and $Q'$
            \State Formulate existential query: $\exists c_1 \cdots \exists c_s~ ( \mathcal{C}' \wedge \forall \mathcal{X}\; ((\mathcal{C} \wedge \mathcal{A} \wedge \hat{A}_{k+1}^{(i,\ell)'}) \rightarrow Q') )$
            \If{theorem prover returns \texttt{Proved}}
                \State Add $\hat{A}_{k+1}^{(i,\ell)}$ to $\mathcal{A}_{\text{expl}}$
            \EndIf
        \EndFor
    \EndFor
\Else
    \State Compute primary decomposition $I = I_1 \cap \cdots \cap I_r$
    \For{each associated prime $P_j = \sqrt{I_j}$}
        \For{each generator $\hat{A}_{k+1}^{(j,m)}$ of $P_j$}
            \State \textbf{Reason:} Form ideal $J \gets \langle A_1, \ldots, A_k, \hat{A}_{k+1}^{(j,m)} \rangle$
            \State Compute Gr\"obner basis $\mathcal{G} \gets \text{GB}(J)$ with lex order $x_1 < \cdots < x_n$
            \State Project: $\mathcal{G}' \gets \mathcal{G} \cap \mathbb{R}[x_1, \ldots, x_d]$
            \If{$Q \in \mathcal{G}'$}
                \State Add $\hat{A}_{k+1}^{(j,m)}$ to $\mathcal{A}_{\text{expl}}$
            \EndIf
        \EndFor
    \EndFor
\EndIf
\State \textbf{Return} $\mathcal{A}_{\text{expl}}$
\end{algorithmic}
\label{alg:explanatory_axioms}
\end{algorithm}

\subsection*{Implementation and Hardware}
All symbolic computations were performed using \texttt{Macaulay2} \cite{M2} and Singular \cite{Singular} for Gr\"obner basis and primary decomposition and \texttt{Python}/\texttt{SymPy} (for preprocessing and verification). Numerical irreducible decomposition was implemented using \texttt{Bertini}~\citep{bates2024numericalnonlinearalgebra, BHSW06} and Macaulay2's \texttt{NumericalAlgebraicGeometry} package~\citep{m2nag}. Symbolic regression on witness sets and theorem proving with KeYmaera X \cite{keymaerax2017} were implemented in \texttt{Python}. Experiments were run on an Apple M4 MacBook Pro with 16\,GB of unified memory. The M4 chip features a 10-core CPU and a 10-core GPU, with hardware acceleration for polynomial arithmetic and linear algebra routines via Apple's Accelerate framework.

Table~\ref{tab:runtime_summary} reports average runtime per test case that ran within the timeout window (7200s) across all benchmark problems, broken down by the number of missing axioms and computational mode (algebraic vs.\ numerical). Algebraic methods scale efficiently for smaller systems, completing single-axiom tests in under 10 seconds for most problems and maintaining sub-minute performance for systems with up to 10 variables. Larger systems (15--16 variables) require 1--2 minutes per single-axiom test, with moderate increases for multi-axiom cases. Numerical methods incur substantially higher computational costs due to homotopy continuation and witness set construction, typically requiring 10--100× longer than their algebraic counterparts. For the most complex systems (Relativistic Laws, Hall Effect, Compton Scattering), numerical methods require 1--2 hours per test case that succeeded.

\begingroup
\arrayrulecolor{black}
\begin{table}[h]
\centering
\footnotesize
\renewcommand{\arraystretch}{1.2}
\begin{tabular}{l@{\hspace{6pt}}c@{\hspace{6pt}}c@{\hspace{6pt}}c@{\hspace{6pt}}c@{\hspace{6pt}}c@{\hspace{6pt}}c@{\hspace{6pt}}c@{\hspace{6pt}}c@{\hspace{6pt}}c}
\toprule
\rowcolor{blue!20}
\textbf{Problem} & \textbf{Vars} & \textbf{Ax.} & \textbf{Deg.} & \multicolumn{3}{c}{\textbf{Algebraic (s)}} & \multicolumn{3}{c}{\textbf{Numerical (s)}} \\
\cmidrule(lr){5-7} \cmidrule(lr){8-10}
\rowcolor{blue!20}
 & & & & \textbf{1-Ax} & \textbf{2-Ax} & \textbf{3-Ax} & \textbf{1-Ax} & \textbf{2-Ax} & \textbf{3-Ax} \\
\midrule
Hagen--Poiseuille       & 9  & 4 & 4 & 0.3 & 1.1 & 1.2 & 3.3 & 9.5 & 10.2 \\
\rowcolor{blue!5}
Kepler                  & 8  & 5 & 4 & 1.4 & 6.2 & 5.8 & 15.6 & 30.4 & 32.5 \\
Time Dilation           & 8  & 5 & 4 & 4.2 & 5.9 & 4.4 & 20.7 & 29.3 & 35.2 \\
\rowcolor{blue!5}
Escape Velocity         & 9  & 5 & 4 & 2.7 & 3.5 & 5.4 & 15.7 & 26.3 & 27.2 \\
Light Damping           & 10 & 5 & 8 & 85.3 & 106.2 & 109.9 & 792.1 & 1022.2 & 1032.6 \\
\rowcolor{blue!5}
Neutrino Decay          & 8  & 5 & 2 & 3.4 & 7.5 & 9.5 & 41.2 & 268.3 & 251.4 \\
Simple Harmonic Osc.    & 10 & 6 & 5 & 9.2 & 13.2 & 12.5 & 341.2 & 690.1 & 798.3 \\
\rowcolor{blue!5}
Carrier-Resolved PH     & 15 & 7 & 6 & 14.5 & 59.9 & 87.2 & 431.5 & 1034.2 & 1242.2 \\
Relativistic Laws       & 12 & 7 & 5 & 144.2 & 231.3 & 202.1 & 5239.2 & 6824.4 & 6523.4 \\
\rowcolor{blue!5}
Hall Effect             & 16 & 7 & 4 & 45.3 & 104.8 & 131.1 & 892.3 & 3821.9 & 3733.2 \\
Inelastic Rel. Col.     & 11 & 9 & 6 & 87.2 & 163.6 & 150.2 & 1222.2 & 4757.4 & 5366.3 \\
\rowcolor{blue!5}
Compton Scattering      & 15 & 10 & 6 & 140.1 & 160.7 & 172.9 & 2419.5 & 6324.2 & 6823.4 \\
\bottomrule
\end{tabular}
\caption{Average runtime per test case for single, pair, and triple axiom removal experiments. Algebraic computations use primary decomposition and Gr\"obner basis methods; numerical computations use witness sets, symbolic regression, and an automated theorem prover. Results are averaged over all successful test cases for each problem.}
\label{tab:runtime_summary}
\end{table}
\endgroup

This being said, it is also important to note that there were many test cases that failed to run within the allotted time limit of (7200s), especially when 3 axioms were missing. See Table \ref{tab:completion_rates} for the number of completed runs for each case. 

\begingroup
\arrayrulecolor{black}
\begin{table}[h]
\centering
\footnotesize
\renewcommand{\arraystretch}{1.2}
\begin{tabular}{l@{\hspace{6pt}}c@{\hspace{6pt}}c@{\hspace{6pt}}c@{\hspace{6pt}}c@{\hspace{6pt}}c@{\hspace{6pt}}c@{\hspace{6pt}}c@{\hspace{6pt}}c@{\hspace{6pt}}c}
\toprule
\rowcolor{blue!20}
\textbf{Problem} & \textbf{Vars} & \textbf{Ax.} & \textbf{Deg.} & \multicolumn{3}{c}{\textbf{Algebraic}} & \multicolumn{3}{c}{\textbf{Numerical}} \\
\cmidrule(lr){5-7} \cmidrule(lr){8-10}
\rowcolor{blue!20}
 & & & & \textbf{1-Ax} & \textbf{2-Ax} & \textbf{3-Ax} & \textbf{1-Ax} & \textbf{2-Ax} & \textbf{3-Ax} \\
\midrule
Hagen--Poiseuille       & 9  & 4 & 4 & 4/4  & 6/6  & 4/4  & 12/12  & 18/18  & 12/12  \\
\rowcolor{blue!5}
Kepler                  & 8  & 5 & 4 & 5/5  & 10/10 & 10/10 & 15/15  & 30/30 & 30/30 \\
Time Dilation           & 8  & 5 & 4 & 5/5  & 10/10 & 10/10 & 15/15  & 30/30 & 30/30 \\
\rowcolor{blue!5}
Escape Velocity         & 9  & 5 & 4 & 5/5  & 10/10 & 10/10 & 15/15  & 30/30 & 30/30 \\
Light Damping           & 10 & 5 & 8 & 5/5  & 10/10 & 10/10 & 15/15  & 30/30 & 30/30 \\
\rowcolor{blue!5}
Neutrino Decay          & 8  & 5 & 2 & 5/5  & 10/10 & 10/10 & 15/15  & 30/30 & 30/30 \\
Simple Harmonic Osc.    & 10 & 6 & 5 & 6/6  & 15/15 & 20/20 & 18/18  & 45/45 & 60/60 \\
\rowcolor{blue!5}
Carrier-Resolved PH     & 15 & 7 & 6 & 7/7  & 21/21 & 35/35 & 21/21  & 63/63 & 105/105 \\
Relativistic Laws       & 12 & 7 & 5 & 7/7  & 21/21 & 35/35 & 9/21  & 18/63 & 102/105 \\
\rowcolor{blue!5}
Hall Effect             & 16 & 7 & 4 & 7/7  & 21/21 & 35/35 & 21/21  & 63/63 & 105/105 \\
Inelastic Rel. Col.     & 11 & 9 & 6 & 9/9  & 36/36 & 84/84 & 27/27  & 108/108 & 252/252 \\
\rowcolor{blue!5}
Compton Scattering      & 15 & 10 & 6 & 10/10 & 45/45 & 120/120 & 3/30 & 5/135 & 2/360 \\
\bottomrule
\end{tabular}
\caption{Test case completion rates within the 7200s timeout window. Values show the number of completed runs out of total test cases for each configuration. Algebraic computations use primary decomposition and Gr\"obner basis methods; numerical computations use witness sets and symbolic regression. The numerical cases are totaled over all three noise levels that were tested on.}
\label{tab:completion_rates}
\end{table}
\endgroup

Algebraic methods completed all test cases within the timeout window across all problems and configurations. Numerical methods achieved perfect completion for most problems, but exhibited significant timeouts for two systems. Relativistic Laws completed only 43\% (9/21) of single-axiom cases, 29\% (18/63) of pair cases, and 97\% (102/105) of triple cases, likely due to the high computational cost of witness set construction for systems with 12 variables and degree-5 polynomials. Compton Scattering experienced severe timeouts, completing only 10\% (3/30) of single-axiom cases, 4\% (5/135) of pair cases, and less than 1\% (2/360) of triple cases - attributable to its large scale (15 variables, 10 axioms, degree 6) combined. Light Damping, having the highest polynomial degree (8), had an increase in runtime as noted in Table \ref{tab:runtime_summary}, however given the manageable number of variables and axioms, did terminate across all runs.

\section{Comparison with State-of-the-Art}\label{app:comparison2sota}

We compare our system against two representative approaches: (i) \textbf{cvc5}-based abduction, and (ii) \textbf{LLM}-based abduction using GPT-5 Pro. Both baselines receive the same input family of physics systems and targets, and both are evaluated by the same algebraic verifier (Macaulay2) under identical projection checks. We also note a few systems that we cannot benchmark against due to our more challenging use case of not assuming we have data for the unknown axiom. 

\paragraph{Benchmarks.}
We use each of the 12 systems from Table \ref{tab:single_ax_summary}
and their noisy counterparts to benchmark the competing methods. For each problem we iterate over all $k$-tuples of removed axioms with $k \in \{1,2,3\}$ (configurable). 

\subsection*{cvc5}

\textsc{cvc5} is a state-of-the-art SMT (Satisfiability Modulo Theories) solver~\cite{barbosa2022cvc5} that supports reasoning over a wide range of logical theories, including nonlinear real arithmetic. 

We instantiate our setting in \textsc{cvc5}\footnote{All experiments use \textsc{cvc5} through its Python API (version~1.1.0).} via its native (experimental) \emph{abduct} API, which synthesizes a candidate axiom \(A_{k+1}\) such that the asserted axioms \(\mathcal{A}_{\text{given}}\) together with \(A_{k+1}\) entail the target \(Q\) while remaining jointly satisfiable. 
Concretely, we enable abduction (\texttt{produce-abducts}) and query the solver with \texttt{getAbduct}. Optionally, by setting \texttt{incremental} \(=\) \texttt{true} we activate the  \textit{incremental mode}, which allows to generate more than one candidate hypothesis axiom via \texttt{getAbductNext}. 
These calls are documented in the official \textsc{cvc5} Python API.

Each system’s axioms $A_1, \ldots, A_k$ and target $Q$ are encoded as quantifier-free nonlinear real arithmetic (\texttt{QF\_NRA}): all equations are written in polynomial form, and asserted into the solver. 
We then pose the abductive query ``find \(A_{k+1}\)'' with a user-provided grammar restricting the search space of candidate polynomials.

We supply a SyGuS grammar (constructed via \texttt{mkGrammar}/\texttt{addRule}) whose start symbol expands to equations of the form \(P = 0\), where \(P\) ranges over polynomials of bounded total degree. 
The grammar includes terminals for constants and variables (closed under addition) and non-terminals degree-stratified polynomials (closed under addition, scalar multiplication, and degree-respecting products, e.g., for degree $3$ we can have linear\(\times\)quadratic). 
All grammars are generated programmatically from the variable list for each system and capped at a degree inferred from the target (with optional rational coefficient seeds for noisy variants).

As mentioned above we use 
\texttt{getAbduct} together with the grammar to obtain one (or more using \texttt{getAbductNext}) candidate \(A_{k+1}\).
We then perform two solver-side sanity checks before sending candidates to our Macaulay2 verifier:
(i) \textit{}{Consistency:} \(\mathcal{A}\) is satisfiable; and
(ii) \textit{Entailment:} \(\mathcal{A} \wedge \neg Q\) is unsatisfiable.

To standardize runtimes across problem sizes, we enforce per-query budgets via \textsc{cvc5}’s time and resource limits (e.g., \texttt{tlimit-per} in milliseconds). 
These are set through \texttt{setOption} and apply independently to each abduct or consistency check. For our experiments, we set a limit of 8 hours for each cvc5 run.

\vspace{6pt}
\subsection*{LLM}
We implement a large language model–based approach to abductive inference using the GPT-5 Pro model (Nov 2025) accessed through the OpenRouter API. 
For each physics system in our benchmark set, the model is provided with the system’s axioms, the list of measured and non-measured variables (anonymized for not introducing biases), and a target polynomial that cannot be derived from the given axioms. 
We then remove one or more axioms (according to the experimental configuration) and prompt the model to propose an additional axiom or set of axioms whose inclusion would make the target polynomial derivable.

To prevent leakage of semantic cues from variable names and to ensure consistent reasoning across domains, all variables in each system are renamed canonically to \(\{x_1, x_2, \dots, x_n\}\).  
The prompt presents the remaining axioms and the target polynomial in this canonical form and instructs the model to act as an abductive inference solver.  
The model may reason freely within its response, but is required to conclude with a section formatted as:
\[
\texttt{Polynomial:} \quad <\text{insert one or more candidate polynomials}>
\]
Each candidate is required to be distinct from the target polynomial.  
We parse only the lines following this tag as the model’s proposed abductive hypotheses.

The proposed axioms are automatically translated back into the system’s original variable names using the inverse of the canonical renaming map.  
For every candidate, we then invoke a symbolic verification step implemented in Macaulay2.  
The candidate axiom(s) are added to the remaining axioms, and the resulting ideal is projected onto the set of measured variables relevant to the target.  
The two checks of \textit{Membership} and \textit{Literal appearance} are then performed corresponding respectively to weak and strong forms of abductive success.  
The verification procedure is identical to that used for algebraic abduction experiments with Macaulay2.

A proposed axiom set $\mathcal{A}_{abd}$ is considered \emph{valid} for target $Q$ iff, when added to the given set of axioms $\mathcal{A}_{given}$, the following hold after eliminating non-measured variables (per target) in Macaulay2:
\begin{enumerate}
\item \textit{Membership:} whether the target polynomial reduces to zero modulo the Gr\"obner basis of the eliminated ideal (i.e., the target lies in the ideal generated by the axioms and proposed axiom). That is,  $Q \in \mathcal{A}$ where $  \mathcal{A} = \mathcal{A}_{abd} \cup \mathcal{A}_{given}$ (remainder-zero modulo the Gr\"obner basis of the eliminated ideal), and
\item \textit{Literal appearance (stronger):} whether the target polynomial appears exactly among the generators of the eliminated Gr\"obner basis. That is, $q$ appears \emph{exactly} among the generators of the eliminated ideal’s Gr\"obner basis.
\end{enumerate}
We report both metrics. For ablations we also track (iii) parsimony (number of added axioms), (iv) solution time, and (v) consistency (the added hypotheses are not already implied by $\mathcal{A}_{abd}$).

All experiments are run using GPT-5 Pro (reasoning model) with temperature \(0.7\) and a maximum of 1200 tokens per completion.  
Each combination of system, number of removed axioms, and specific subset of removed axioms is processed independently.  
For each case, the full prompt, model output, parsed candidate polynomials (in canonical and original variables), and the results of the Macaulay2 verification are written to structured output directories for later analysis.  
This design enables exact reproducibility of every abductive query and isolates LLM behavior from downstream algebraic validation.  


Our results can be summarized in Table \ref{tab:method_comparison}, shown again below.

\begin{table}[h!]
\centering
\renewcommand{\arraystretch}{1.25}
\rowcolors{2}{warmrow}{warmbg}
\begin{tabularx}{\textwidth}{>{\bfseries}l*{6}{>{\centering\arraybackslash}X}}
\rowcolor{warmheader}
\toprule
Method & Single Axiom & Multi Axiom & Data-Free & Noise Robust & Var. Agnostic & Explainable \\ 
\midrule
Traditional SR & \xmark & \xmark & \xmark & \xmark & \xmark & \xmark \\
Integrated SR & \xmark & \xmark & \xmark & \cmark & \cmark & \cmark \\
LLM (GPT-5 Pro) & 0\% & 0\% & \cmark & \xmark & \cmark & \cmark \\
cvc5 & t-out & t-out & \cmark & \xmark & \xmark & \cmark \\
\midrule
AI-Noether & \textbf{97\%}~\cmark & \textbf{49\%}~\cmark & \cmark & \cmark & \cmark & \cmark \\
\bottomrule
\end{tabularx}

\vspace{4pt}
\caption*{Table \ref{tab:method_comparison}: \textbf{Comparison of abductive and symbolic regression methods.}
Green ticks (\cmark) indicate capability support; red crosses (\xmark) indicate absence.  
“Data-Free” means the method operates without numeric data or prior parameter information,  
and “Var. Agnostic” indicates invariance to variable naming. ``t-out'' means that the method ran out of time without producing a solution.}
\end{table}

\paragraph{Summary of results.}
Neither baseline successfully recovered missing axioms in our benchmark. The LLM-based approach (GPT-5 Pro) failed to generate any valid candidate axioms across all test cases, producing either syntactically malformed outputs or polynomials that did not satisfy the algebraic verification criteria when back-mapped to the original variable space. The cvc5 solver consistently timed out after 8 hours without producing any abducts, likely due to the combinatorial explosion in the search space induced by the high polynomial degrees (up to degree 8 in our benchmarks) and the number of variables (up to 16). In contrast to traditional symbolic regression methods that require training data for each unknown axiom, and in contrast to these unsuccessful abductive baselines, AI-Noether operates in a purely data-free setting while achieving 97\% recovery on single-axiom removal and 49\% on multi-axiom removal. This demonstrates that algebraic-geometric decomposition provides a tractable pathway to abductive inference in scientific discovery that remains out of reach for both SMT-based synthesis and LLM-based reasoning under current computational constraints.

\end{document}